\documentclass{article}
\usepackage{hyperref}

\usepackage[accepted]{icml2025}

\usepackage{amssymb}
\usepackage{amsmath}
\usepackage{nth}
\usepackage[all]{nowidow}
\usepackage{bbm}
\usepackage{booktabs,etoolbox}
\usepackage{multirow}
\usepackage{caption}
\usepackage{graphicx}
\usepackage{subcaption}
\usepackage{bbding}
\usepackage{overpic}
\usepackage{makecell}
\usepackage{enumitem}

\usepackage{tikz}
\usetikzlibrary{fit}


\usepackage{array}
\usepackage{ragged2e}
\newcolumntype{P}[1]{>{\centering\arraybackslash}p{#1}}
\newcolumntype{R}[1]{>{\RaggedLeft\arraybackslash}p{#1}}

\usepackage[export]{adjustbox}
\usepackage{arydshln}

\usepackage{pifont}
\newcommand{\vmark}{\ding{51}}%
\newcommand{\xmark}{\ding{55}}%

\usepackage{amsmath}
\DeclareMathOperator*{\softmax}{softmax}

\newcommand{\customsize}[1]{{\fontsize{9.8}{10}\selectfont #1}}
\newcommand{\labelsize}[1]{{\fontsize{9.0}{10}\selectfont #1}}

\newcommand{\COCVP}{\text{PlaySlot}}
\newcommand{\Method}{\text{PlaySlot}}

\newcommand{\NumPreds}{T}
\newcommand{\NumFrames}{T}

\newcommand{\ImageRange}[2]{\mathbf{X}_{#1:{#2}}}

\newcommand{\ImageT}[1]{\textbf{X}_{#1}}
\newcommand{\PredImageT}[1]{\hat{\textbf{X}}_{#1}}

\newcommand{\Slots}{\mathbf{S}}
\newcommand{\SlotDim}{D_{\Slots}}
\newcommand{\NumSlots}{N_\Slots}

\newcommand{\SlotsT}[1]{\textbf{S}_{#1}}
\newcommand{\PredSlotsT}[1]{\hat{\textbf{S}}_{#1}}
\newcommand{\SingleSlotsT}[2]{\textbf{s}_{#1}^{#2}}

\newcommand{\SlotsMany}[2]{\mathbf{S}_{#1:#2}}

\newcommand{\FeatureMaps}{\textbf{h}}
\newcommand{\DimFeats}{D_h}
\newcommand{\NumLocs}{L}

\newcommand{\GRU}{\text{GRU}}
\newcommand{\Attention}{\textbf{A}}

\newcommand{\SlotObject}[2]{\textbf{o}_{#1}^{#2}}
\newcommand{\SlotMask}[2]{\textbf{m}_{#1}^{#2}}
\newcommand{\ProcessedMask}[2]{\tilde{\textbf{m}}_{#1}^{#2}}

\newcommand{\NumPredLayers}{N_{\text{Pred}}}

\newcommand{\Action}{\textbf{a}}
\newcommand{\ActionT}[1]{\textbf{a}_{#1}}

\newcommand{\MeanDynT}[1]{\boldsymbol{\mu_{\textbf{d}}}_{#1}}
\newcommand{\VarDynT}[1]{\boldsymbol{\sigma_{\textbf{d}}}^2_{#1}}

\newcommand{\ActionProtoT}[1]{\textbf{p}_{#1}}
\newcommand{\ActionProtosMany}[2]{\textbf{p}_{#1:#2}}

\newcommand{\ActionVariabilityT}[1]{\textbf{v}_{#1}}
\newcommand{\ActionVariabilityMany}[2]{\textbf{v}_{#1:#2}}
\newcommand{\ActionEmbPred}{f_{\textbf{z}}}
\newcommand{\ActToken}{\texttt{[ACT]}}

\newcommand{\MeanActT}[1]{\boldsymbol{\mu_{\textbf{z}}}_{#1}}
\newcommand{\VarActT}[1]{\boldsymbol{\sigma_{\textbf{z}}}^2_{#1}}

\newcommand{\LatentBig}{\boldsymbol{Z}}

\newcommand{\DimLatent}{{D_{\textbf{z}}}}
\newcommand{\PredLatent}{\mathbf{\hat{z}}}
\newcommand{\PredLatentT}[1]{\mathbf{\hat{z}}_{#1}}

\newcommand{\PredLatentBigT}[1]{\mathbf{\hat{Z}}_{#1}}
\newcommand{\PredLatentBigSingleT}[2]{\mathbf{\hat{z}}_{#1}^{#2}}

\newcommand{\ImageEncoder}{\mathcal{E}_{\textrm{SAVi}}}
\newcommand{\SlotDecoder}{\mathcal{D}_{\text{SAVi}}}
\newcommand{\InverseDynamics}{\text{InvDyn}}
\newcommand{\InverseDynamicsSingle}{\text{InvDyn}_{\text{S}}}
\newcommand{\InverseDynamicsMany}{\text{InvDyn}_{\text{M}}}
\newcommand{\CondPred}{\text{cOCVP}}
\newcommand{\PolicyModel}{f_{\pi}}
\newcommand{\ActionDecoder}{\mathcal{D}_\textrm{a}}

\newcommand{\LinearSlots}{f_{\text{S}}}
\newcommand{\LinearProto}{f_{\text{p}}}
\newcommand{\LinearVar}{f_{\text{v}}}

\newcommand{\Loss}{\mathcal{L}}

\newcommand{\ButtonPress}{ButtonPress}
\newcommand{\RobotDB}{BlockPush}

\def\R{{\mathbb R}}

\newcommand{\Figure}[1]{Fig.~\ref{#1}}

\newcommand{\Equations}[2]{Equations~\eqref{#1} and~\eqref{#2}}
\newcommand{\Table}[1]{Tab.~\ref{#1}}

\newcommand{\Section}[1]{Sec.~\ref{#1}}

\newcommand{\Appendix}[1]{Appendix~\ref{#1}}

\icmltitlerunning{PlaySlot: Learning Inverse Latent Dynamics for Controllable Object-Centric Video Prediction and Planning}

\begin{document}

\twocolumn[

\icmltitle{{PlaySlot: Learning Inverse Latent Dynamics for Controllable \\ Object-Centric Video Prediction and Planning}}

\icmlsetsymbol{equal}{*}

\begin{icmlauthorlist}
\icmlauthor{Angel Villar-Corrales}{ais}
\icmlauthor{Sven Behnke}{ais}

\end{icmlauthorlist}

\icmlaffiliation{ais}{
	Autonomous Intelligent Systems, Computer Science Institute VI – Intelligent Systems and Robotics,
	Center for Robotics and the Lamarr Institute for Machine Learning and Artificial Intelligence,
	University of Bonn, Germany
}
\icmlcorrespondingauthor{Angel Villar-Corrales}{villar@ais.uni-bonn.de}
\icmlkeywords{Object-Centric Video Prediction, Object-Centric Learning, Video Prediction, Representation Learning, Action Planning, Transformers}

\vskip 0.3in
]

\printAffiliationsAndNotice{}

\begin{abstract}
Predicting future scene representations is a crucial task for enabling robots to understand and interact with the environment.
However, most existing methods rely on videos and simulations with precise action annotations, limiting their ability to leverage the large amount of available unlabeled video data.
To address this challenge, we propose \emph{\Method}, an object-centric video prediction model that infers object representations and latent actions from unlabeled video sequences. It then uses these representations to forecast future object states and video frames.
\Method{} allows the generation of multiple possible futures conditioned on latent actions, which can be inferred from video dynamics, provided by a user, or generated by a learned action policy, thus enabling versatile and interpretable world modeling.
Our results show that \Method{} outperforms both stochastic and object-centric baselines for video prediction across different environments.
Furthermore, we show that our inferred latent actions can be used to learn robot behaviors sample-efficiently from unlabeled video demonstrations.
Videos and code are available on \href{https://play-slot.github.io/PlaySlot/}{our project website}.
\end{abstract}

\begin{figure}[t!]
	\centering
	\includegraphics[width=\linewidth]{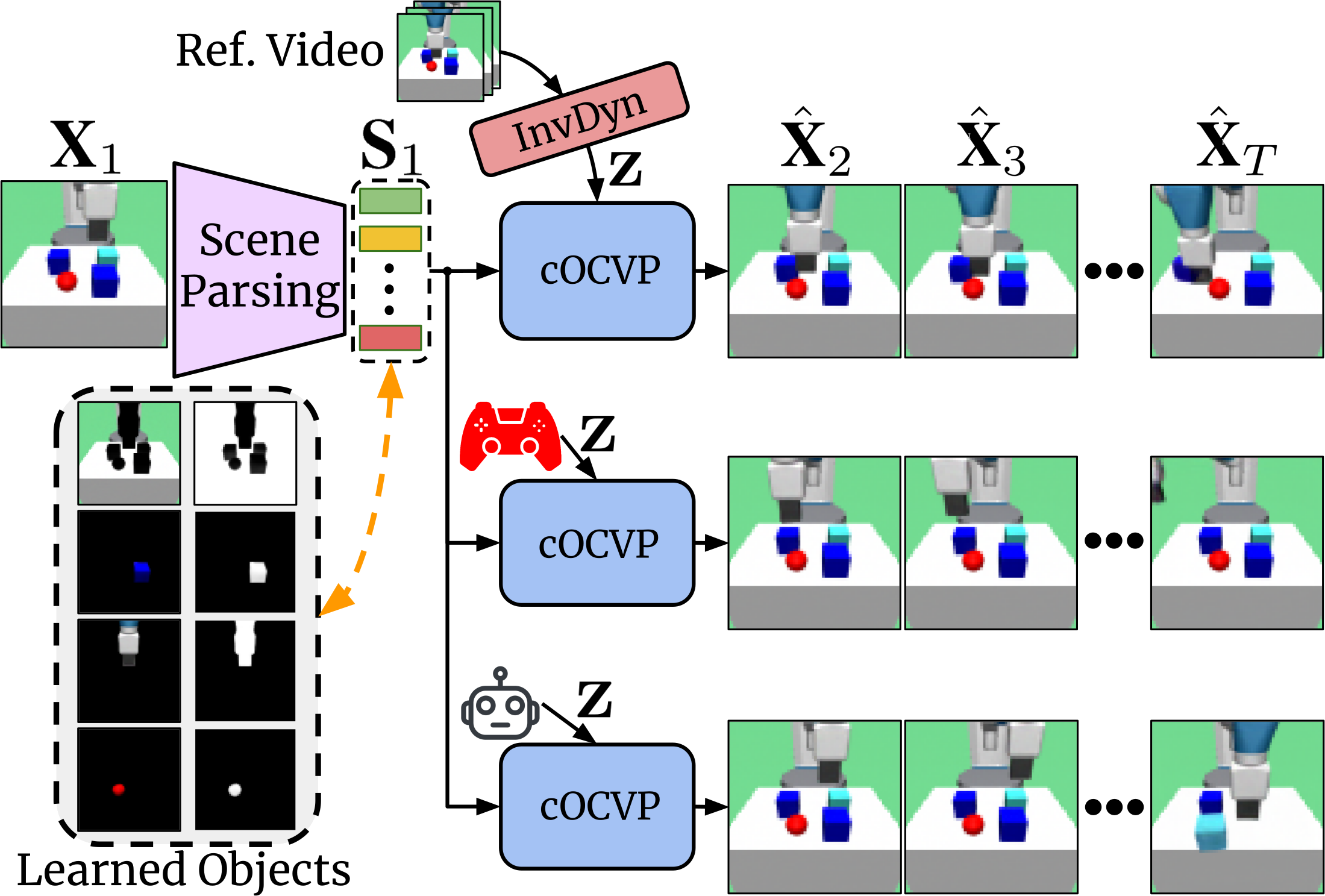}
	\vspace{-0.6cm}
	\caption{
		\Method{} parses an image $\ImageT{1}$ into its object components $\SlotsT{1}$.
		It then predicts multiple future object states and frames with an object-centric video prediction module (\CondPred) conditioned on latent actions $\LatentBig$, which can be inferred from a reference video with our $\InverseDynamics$ module, provided as input, or generated by a learned action policy.
	}
	\label{fig: teaser}
\end{figure}

\vspace{-0.5cm}
\section{Introduction}

Accurate and flexible world models are crucial for autonomous systems to reason about
their surroundings, predict possible future outcomes, and plan their actions effectively.
Such models require a structured representation of the world that supports generalization, robustness, and controllability, especially in complex and dynamic scenarios.

Humans naturally achieve such understanding by parsing their environment into a \emph{background} and multiple separate \emph{objects}, which can interact with each other and can be recombined to form more complex entities~\cite{Johnson_ObjectPerception_2018, Kahneman_ReviewingOfObjectFiles_1992}.
Neural networks equipped with such compositional inductive biases have the ability to learn structured object-centric representations with desirable properties such as robustness~\cite{Bengio_RepresentationLearningReview_2013, Dittadi_GeneralizationAndRobustnessImplicationsInObjectCentricLearning_2022}, generalization to novel compositions~\cite{Greff_OnTheBindingProblemInNeuralNetworks_2020}, transferability to novel tasks~\cite{Zhang_IsAnObject-CentricVideoRepresentationBeneficialForTransfer_2022}, and sample efficiency~\cite{Mosbach_SOLDReinforcementLearningSlotObjectCentricLatentDynamics}, among others.

Building on these foundations, the field of object-centric learning has made great advances in recent years, progressing from learning object representations in simple synthetic images~\cite{Locatello_ObjectCentricLearningWithSlotAttention_2020, Burgess_MonetUnsupervisedSceneDecompositionRepresentation_2019} and videos~\cite{Kipf_ConditionalObjectCentricLearningFromVideo_2022, Elsayed_SAVi++TowardsEndToEndObjectCentricLearningFromRealWorldVideos_2022}, towards more complex real-world scenes~\cite{Seitzer_BridgingTheGapToRealWorldObjectCentricLearning_2023, Zadaianchuk_VideoSaur_2024}.
Recently, the field of object-centric video prediction combines these learned object representations with forward-dynamics models and has shown great promise for multiple downstream applications such as modeling object dynamics~\cite{Villar_OCVP_2023, Wu_SlotFormer_2022} or action planning~\cite{Yoon_InvestigationObjectCentricRepresentationsReinforcementLearning_2023, Mosbach_SOLDReinforcementLearningSlotObjectCentricLatentDynamics}.
However, such models are currently limited to deterministic environments or rely on videos and simulations with precise action labels to forecast scene dynamics, limiting their ability to leverage unlabeled video data and serve as world models for robotic applications.

In this work, we propose \emph{\Method}, a novel method for controllable video prediction using object-centric representations.
\Method{} learns in a self-supervised manner from video to infer object representations, called slots, and latent action embeddings, which are computed using our proposed \emph{$\InverseDynamics$} module to capture the scene dynamics.
\Method{} then predicts future video frames conditioned on the inferred object slots and latent actions.
At inference time, as illustrated on \Figure{fig: teaser}, \Method{} parses the observed environment into a set of object slots, each of them representing a different object in the image.
Then, \Method{} forecasts future object states and frames conditioned on past object slots and latent actions, which can be inferred from a video sequence using our proposed $\InverseDynamics$ module, provided by a human, or generated by a learned action policy.

In our experiments, we demonstrate that $\Method${} learns a rich and semantically meaningful action space, enabling accurate video prediction while providing high levels of controllability and interpretability.
We show how $\Method${} effectively captures precise robot actions and seamlessly scales to scenes with multiple moving objects or to real-world robotics data, outperforming several controllable video prediction baselines.
Moreover, we show that the latent actions inferred by $\Method${} enable sample-efficient learning of robot behaviors from unlabeled demonstrations.

In summary, our contributions are as follows:
\begin{itemize}[itemsep=1pt, topsep=0.3pt]
	\item We propose \emph{$\COCVP$}---an object-centric video prediction model that infers object representations and latent actions from unlabeled videos, and uses them to forecast future object states and video frames.

	\item $\COCVP$ outperforms several video prediction models across diverse robotic environments, while showing superior interpretability and control capabilities.

	\item The object representations and latent actions inferred by $\COCVP$ can be used to learn robot behaviors from unlabeled video demonstrations sample efficiently.
\end{itemize}

\section{Related Work}

\paragraph{Unsupervised Slot-Based Object-Centric Learning}
Slot-based object-centric representation methods aim to parse in an unsupervised manner an image or video into a set of $\NumSlots$ latent vectors called slots, where each of them binds to a different object in the scene~\cite{Greff_OnTheBindingProblemInNeuralNetworks_2020, Locatello_ChallengingAssumptionsInLearningOfDissentangledRepresentations_2019}.
Early approaches aimed to learn object representations from synthetic images~\cite{Locatello_ObjectCentricLearningWithSlotAttention_2020, Singh_SLATE_2021, Biza_InvariantSlotAttention_2023} or videos~\cite{Kipf_ConditionalObjectCentricLearningFromVideo_2022, Creswell_UnsupervisedObjectBasedTransitionModelsFor3DPartiallyObservableEnvironments_2021, Singh_STEVE_2022} by minimizing a reconstruction objective.
To learn meaningful representations from real data, recent slot-based methods leverage weak supervision~\cite{Elsayed_SAVi++TowardsEndToEndObjectCentricLearningFromRealWorldVideos_2022, Bao_ObjectDiscoverMotion_2023}, large pretrained transformers~\cite{Seitzer_BridgingTheGapToRealWorldObjectCentricLearning_2023, Aydemir_SelfSupervisedObjectCentricLearningForVideos_2023, Zadaianchuk_VideoSaur_2024}, or diffusion models~\cite{Jiang_ObjectCentricSlotDiffusion_2023, Wu_SlotDiffusion_2023}
These object-centric representations benefit multiple downstream tasks such as reinforcement learning for robotic manipulation~\cite{Mosbach_SOLDReinforcementLearningSlotObjectCentricLatentDynamics, Ferraro_Focus_2023} or visual-question-answering~\cite{Mamaghan_ObjectCentricRepsForVQA_2025}.

\paragraph{Object-Centric Video Prediction}

Object-centric video prediction aims to model the object dynamics and interactions in a video sequence with the goal of forecasting future object states and video frames.
Several methods address this task using different architectural priors, including RNNs~\cite{Zoran_PARTS_2021, Assouel_VariationIdependentModulesVP_2022}, transformers~\cite{Villar_OCVP_2023, Wu_SlotFormer_2022, Daniel_DDLP_2024, Meo_ObjectCentricTemporalConsistency_2024} or state-space models~\cite{Jiang_SlotSSM_2024}, attaining a remarkable prediction accuracy on synthetic datasets.
Recently, some methods improve the controllability of object-centric video prediction models by conditioning the prediction process on actions~\cite{Mosbach_SOLDReinforcementLearningSlotObjectCentricLatentDynamics} or language captions~\cite{Wang_TIVDiffusion_2025, TextOCVP, Jeong_ObjectCentricWorldModelForLanguageGuidedManipulation_2025}.
However, forecasting future object states without supervision in complex environments still remains an open challenge.

\input{imgs/main_fig.tex}

\paragraph{Learning Latent Actions from Unlabeled Videos}

Videos provide abundant information about dynamics and activities, but often lack the action labels necessary for learning behaviors from video.
To address this challenge, some methods train a latent policy directly from observations by learning a discrete latent action space and sampling the actions that minimize a reconstruction error~\cite{Edwards_ImitatingLatentPoliciesFromObservation_2019, Struckmeier_ILPI_2023}.
Another group of methods, to which \Method{} belongs, learns inverse dynamics from unlabeled videos by predicting latent actions given pairs of observations, and uses them for learning behaviors for video games and robot simulations~\cite{Ye_BecomeAProficientPlayerByWatchingPureVideos_2022, Brandfonbrener_InverseDynamicsPretrainingLearnsGoodRepresentatiosForImitation_2024, Schmidt_LAPOLearningToActWithoutActions_2024}, as pretraining for Vision-Language-Action models~\cite{Ye_LatentActionPretrainingFromVideos_2025} or for learning robot policies~\cite{Cui_DynaMoDynamcisPretrainingForVisuoMotorControl_2024, Tian_PredictiveInverseDynamocs_2024}.

Latent action models have also been used for conditional video prediction.
The most similar method to ours is \emph{CADDY}~\cite{Menapace_PlayableVideoGeneration_2021, Menapace_PlayableEnvironments_2022}, which learns latent actions from a collection of unlabeled videos from a single domain and uses the latent actions as conditioning signal for predicting future frames.
At inference time, CADDY maps user inputs to the latent space for playable video generation.
Building upon this same principle, \emph{Genie}~\cite{Bruce_GenieGenerativeInteractiveEnvironments_2024} proposes a foundation world model for playable video generation on diverse environments.
However, both CADDY and Genie operate on holistic scene representations, which are limited for tasks that require relational reasoning, often struggle to model object relationships and interactions, and require human supervision to generalize to scenes with multiple moving agents.

\section{\Method}

We propose \emph{\Method}, a novel framework for controllable object-centric video prediction from unlabeled video sequences.
\Figure{fig: main}a) illustrates the training process in \Method, as well as its main four components.
Namely, given $\NumFrames$ video frames $\ImageRange{1}{\NumFrames}$, our model employs as \emph{Scene Parsing module} (\Section{section: savi}) that decomposes these images into object representations, called slots, $\SlotsMany{1}{\NumFrames} = (\SlotsT{1}, ..., \SlotsT{\NumFrames})$, where $\SlotsT{t}=(\SingleSlotsT{t}{1}, ..., \SingleSlotsT{t}{\NumSlots}) \in \R^{\NumSlots \times \SlotDim}$ is the set of $\SlotDim$-dimensional object slots parsed from frame $\ImageT{t}$.
For each consecutive pair of frames, \Method{} employs an \emph{Inverse Dynamics} (\InverseDynamics) module (\Section{section: inverse dynamics}) in order to estimate latent action embeddings
$\PredLatentT{t}$ that encode the actions taken by the agents in the scene between every consecutive pair of frames.
The \emph{Conditional Object-Centric Predictor} (\CondPred) (\Section{section: predictor}) forecasts future object states conditioned on past slots and latent actions estimated by $\InverseDynamics$.
Finally, the \emph{object rendering} module decodes the object slots to render object images and masks, which can be combined via a weighted sum to render video frames.

At inference time, as shown in \Figure{fig: main}b), \Method{} autoregressively predicts multiple possible sequence continuations conditioned on the initial object slots and latent action embeddings, which can be estimated by \InverseDynamics, provided by a user, or generated by a learned action policy.

\subsection{Object-Centric Representation Learning}
\label{section: savi}

\Method{} employs SAVi~\cite{Kipf_ConditionalObjectCentricLearningFromVideo_2022},
a recursive encoder-decoder model with a structured bottleneck of $\NumSlots$ permutation-equivariant object slots,
to parse a sequence of video frames $\ImageRange{1}{\NumFrames}$ into their object components $\SlotsMany{1}{\NumFrames}, \SlotsT{t} \in \R^{\NumSlots \times \SlotDim}$.
The slots $\SlotsT{0}$ are sampled from a learned distribution and recursively refined to bind to the objects in the video frames.
At time step $t$, SAVi encodes the corresponding input $\ImageT{t}$ into feature maps $\FeatureMaps_t \in \R^{\NumLocs \times \DimFeats}$,
which are fed to Slot Attention~\cite{Locatello_ObjectCentricLearningWithSlotAttention_2020} to
iteratively refine the previous slot representations conditioned on the current features.
Slot Attention performs cross-attention between the image features and slots with the attention weights normalized over the slot dimension, encouraging competition between slots so as to represent each feature location:
\vspace{-0.1cm}
\begin{align}
	& \Attention = \softmax_{\NumSlots} \left( \frac{q(\SlotsT{t-1})\cdot k(\FeatureMaps_t)^T}{\sqrt{\SlotDim}} \right) \in \R^{\NumSlots \times \NumLocs},
	\label{eq: slot attention}
\end{align}
\vspace{-0.1cm}
where $q$ and $k$ are linear projections. 
The slots are then independently updated via a shared Gated Recurrent Unit~\cite{Cho_GRU_2014} (GRU) followed by a residual MLP:
\begin{align}
	& \SlotsT{t} = \GRU(\Attention \cdot v(\FeatureMaps_t), \SlotsT{t-1})
	\; , \;\; \Attention_{n,l} = \frac{\Attention_{n,l}}{\sum_{i=0}^{\NumLocs-1}\Attention_{n,i}},
	\label{eq: slot update}
\end{align}
where $v$ is a linear projection. The steps described in \Equations{eq: slot attention}{eq: slot update} can be repeated multiple times with shared
weights to iteratively refine the slots and obtain an accurate object-centric
representation of the scene.

To map the object representations back to images, SAVi independently decodes each slot in $\SlotsT{t}$ with a Spatial Broadcast Decoder~\cite{Watters_SpatialBroadcastDecoder_2019} ($\SlotDecoder$) to render an object image and mask, which can be normalized and combined via a weighted sum to render the reconstructed frame:
\begin{align}
	& \SlotObject{t}{n}, \SlotMask{t}{n} = \SlotDecoder(\SingleSlotsT{t}{n}), \\
	& \PredImageT{t} = \sum_{n=1}^{\NumSlots} \SlotObject{t}{n} \cdot \ProcessedMask{t}{n} \;\; \text{with} \; \ProcessedMask{t}{n} = \softmax_{\NumSlots}(\SlotMask{t}{n}).
\end{align}

\subsection{Learning Inverse Dynamics}
\label{section: inverse dynamics}

In general, future frames depend not only on previous observations, but also on other variables, such as robot actions.
We propose an inverse dynamics module (\InverseDynamics) that estimates, given the object slots from two consecutive time steps, latent action embeddings $\PredLatent \in \R^\DimLatent$ that encode the actions taken by the agents between such time steps:
\begin{align}
	& \PredLatentT{t} = \InverseDynamics(\SlotsT{t}, \SlotsT{t+1}).
\end{align}

\subsubsection{Action Parameterization}
\label{section: action parameterization}
The parameterization of the latent actions determines the complexity of transitions that can be modeled, as well as the degree of control that we have over the predictions.
On the one hand, learning a finite set of latent actions enables controllable video prediction while limiting the complexity of the dynamics that such actions can explain.
On the other hand, continuous latent vectors offer greater flexibility to model complex transitions between frames, but at the cost of reduced interpretability and control.

As a compromise between these two approaches, inspired by~\citet{Menapace_PlayableVideoGeneration_2021}, we propose a hybrid parameterization of the latent action $\PredLatentT{t}$ as the sum of a discrete $\ActionProtoT{t}$ and a continuous $\ActionVariabilityT{t}$ component:
$\PredLatentT{t} = \ActionProtoT{t} + \ActionVariabilityT{t}$.
The discrete component $\ActionProtoT{t}$, denoted as \emph{action prototype}, determines the high-level action taking place (e.g. move left, go up), whereas the continuous \emph{action variability} $\ActionVariabilityT{t}$ captures stochastic and fine-grained variations, allowing the model to interpolate between prototypes and account for environmental uncertainty.
This hybrid design enables the model to learn expressive dynamics while maintaining control and interpretability.

\subsubsection{\InverseDynamics~Module}
\label{sec: inverse dynamics}

We propose two variants of our inverse dynamics module for inferring latent actions from object-centric representations.
{$\InverseDynamicsSingle$} processes object slots $\SlotsT{t}$ alongside a learnable token~\ActToken~using a transformer encoder $\ActionEmbPred$, producing a single latent action $\PredLatentT{t}$ that represents the agent's behavior. This formulation is particularly well-suited for single-agent environments.
In contrast, {$\InverseDynamicsMany$} processes each slot with a shared MLP, producing $\NumSlots$ latent action embeddings $\PredLatentBigT{t}=\{\PredLatentBigSingleT{t}{1}, ..., \PredLatentBigSingleT{t}{\NumSlots}\}$, where each vector represents the action of a specific object in the scene.
Below we explain the process for computing latent actions using $\InverseDynamicsSingle$, which follows a similar procedure to that of $\InverseDynamicsMany${}.

Following \citet{Menapace_PlayableVideoGeneration_2021}, we adopt a probabilistic formulation where $\InverseDynamics$ predicts the posterior distribution of scene dynamics, modeled as Gaussian:
\begin{align}
	\MeanDynT{t}, \VarDynT{t} = \ActionEmbPred(\SlotsT{t}, \ActToken).
\end{align}
The distribution of latent actions $\PredLatentT{t}$ is then defined as the difference between the distributions of scene dynamics from two consecutive time steps:
\begin{align}
	& \PredLatentT{t} \sim \mathcal{N}(\MeanActT{t}, \VarActT{t}) \;\; \text{with} \;\;
	\begin{cases}
		\MeanActT{t} = \MeanDynT{t+1} - \MeanDynT{t}, \\
		\VarActT{t} = \VarDynT{t+1} + \VarDynT{t},
	\end{cases},
\end{align}
from which latent actions $\PredLatentT{t}$ are sampled.

To prevent the model from trivially encoding future states into $\PredLatentT{t}$, we regularize the latent action space by enforcing an information bottleneck.
Specifically, we constrain the latent action space to be low dimensional, i.e., $\DimLatent << \SlotDim$.
Furthermore, as discussed in \Section{section: action parameterization}, we parameterize the latent actions as the sum of a discrete action prototype $\ActionProtoT{t}$ and a continuous action variability embedding $\ActionVariabilityT{t}$,
where $\ActionProtoT{t}$ is obtained by vector-quantizing the latent actions $\PredLatentT{t}$ ($\ActionProtoT{t} = \text{VQ}(\PredLatentT{t})$).
We empirically verify that the information bottleneck enforced by vector quantization achieves comparable performance to the one proposed by~\citep{Menapace_PlayableVideoGeneration_2021}, while requiring significantly fewer hyper-parameters.

Our hybrid latent action parameterization ensures that our $\InverseDynamics$ module encodes only the essential dynamics, effectively capturing the agent’s interaction with the scene while learning semantically meaningful action prototypes.
Moreover, this hybrid factorization improves the controllability and interpretability of the prediction process while maintaining the ability to model complex scene dynamics.
%

\subsection{Conditional Object-Centric Prediction}
\label{section: predictor}

We employ a transformer-based~\cite{Vaswani_AttentionIsAllYouNeed_2017} module to autoregressively predict future object slots
conditioned on past object states and latent actions.

Our proposed predictor, \CondPred, is a transformer encoder with $\NumPredLayers$ layers.
At each time step $t$, \CondPred~takes as input all previous slots $\SlotsMany{1}{t}$, action prototypes $\ActionProtosMany{1}{t}$ and variability embeddings $\ActionVariabilityMany{1}{t}$, all of which are first linearly projected into a shared token dimensionality.
The slots are then conditioned by adding them with the corresponding projected action prototype and variability embeddings.
Additionally, we incorporate sinusoidal positional encodings such that all slots from the same time step receive the same encoding, thus preserving the inherent permutation equivariance of the objects.

\CondPred~forecasts the future slots $\PredSlotsT{t+1}$ by jointly modeling the object dynamics and interactions from the past object slots conditioned on the inferred latent actions.
This process is summarized as:
\begin{align}
	& \PredSlotsT{t+1} = \CondPred(\LinearSlots(\SlotsMany{1}{t}) + \LinearProto(\ActionProtoT{1:t}) + \LinearVar(\ActionVariabilityT{1:t})),
\end{align}
where $\LinearSlots, \LinearProto~\text{and}~\LinearVar$ are learned linear layers.

The prediction process can be initiated from the slots of a single reference frame $\SlotsT{1}$ and the corresponding inferred latent actions $\PredLatentT{1}$.
This process is repeated autoregressively, with the predicted slots being appended to the input at each subsequent time step, allowing the generation of future object representations for a desired number of time steps $\NumPreds$.

\subsection{Learning Behaviors from Unlabeled Videos}
\label{sec: behavior learning}

Using a pretrained \InverseDynamics~module, we aim to learn a latent policy from unlabeled video expert demonstrations—without the need for action or reward information.
For this purpose, we first use $\InverseDynamics$ to annotate a dataset of unlabeled expert demonstrations with latent actions that capture the underlying scene dynamics.
Subsequently, we train a latent policy model $\PolicyModel$ via behavior cloning, regressing the inferred latent actions from the object slots observed at the corresponding time step.

At inference time, starting from a single observation $\ImageT{1}$, \Method{} computes the corresponding object slots $\SlotsT{1}$ and uses the learned policy to predict a latent action $\PredLatentT{1}$, which is decomposed into an action prototype  $\ActionProtoT{1} = \text{VQ}(\PredLatentT{1})$ and a variability embedding $\ActionVariabilityT{1} = \PredLatentT{1} - \ActionProtoT{1}$.
These representations are fed to $\CondPred$ to forecast the subsequent set of object slots $\PredSlotsT{2}$.
This process is repeated autoregressively, allowing the learned behavior to unfold within the model's latent imagination.

To enable the execution of latent actions in a simulator, we introduce an action decoder $\ActionDecoder$ that maps the predictions of the latent policy $\PolicyModel$ to real-world actions.
This module, implemented as a three-layer MLP, is trained on a small set of action-labeled data to translate latent actions inferred by $\InverseDynamics$~into the corresponding executable actions.
Importantly, the action decoder is only needed to execute latent actions in the simulator for evaluation purposes, and the core training process of \Method{} is fully unsupervised, operating entirely on unlabeled, action-free videos.

Our approach shares similarities with \citet{Schmidt_LAPOLearningToActWithoutActions_2024}, who learn policies from video using discrete latent actions in simple game environments.
In contrast, \Method{} leverages a more expressive action representation and a conditional object-centric decoder, enabling the learning of more complex robot behaviors in realistic environments.

\subsection{Training}

We differentiate three different training stages in \Method.
We first train SAVi to parse video frames into object-centric representations by minimizing a reconstruction loss:
\begin{align}
	\Loss_{\text{SAVi}} &=  
	\sum_{t=1}^{\NumFrames}|| \SlotDecoder(\ImageEncoder(\ImageT{t})) - \ImageT{t} ||_2^2, \label{eq: loss savi}
\end{align}
where $\ImageEncoder$~and $\SlotDecoder$~correspond to the scene parsing and object rendering modules, respectively.

Second, given the pretrained SAVi model, we jointly train $\InverseDynamics$ and $\CondPred$ by minimizing a combined loss:
\begin{align}
	\Loss_{{\Method}} &= 
	\sum_{t=2}^{\NumPreds+1} \lambda_{\text{Img}}
	\Loss_{\text{Img}} + \lambda_{\text{Slot}} \Loss_{\text{Slot}} + \lambda_{\text{VQ}} \Loss_{\text{VQ}},  \label{eq: full loss}  \\
	\Loss_{\text{Img}} &=  || \PredImageT{t} - \ImageT{t} ||_2^2 \label{eq: loss img}, \\
	\Loss_{\text{Slot}} &=  || \PredSlotsT{t} - \ImageEncoder(\ImageT{t}) ||_2^2,
	\label{eq: loss slot}   \\
	\Loss_{\text{VQ}} &= ||\texttt{sg}[\PredLatentT{t}] - \ActionProtoT{t}|| +
	0.25 \cdot || \PredLatentT{t}- \texttt{sg}[\ActionProtoT{t}] ||,  \label{eq: loss vq}
\end{align}
where \texttt{sg} is the stop-gradient operator. $\Loss_{\text{Img}}$ measures the future frame prediction error, $\Loss_{\text{Slot}}$ aligns the predicted object slots with the actual object-centric representations, and $\Loss_{\text{VQ}}$ encourages the learning of meaningful action prototypes while regularizing the latent actions to align with their prototypes~\cite{VanOord_NeuralDiscreteRepresentationLearning_2017}.
We do not employ teacher forcing, enabling the predictor model to learn to handle its own imperfect predictions.

Finally, the policy model $\PolicyModel$ and action decoder $\ActionDecoder$ are trained
to regress the inferred latent actions $\PredLatent$ and ground truth actions $\Action$, respectively:
\vspace{-0.1cm}
\begin{align}
	\Loss_{\PolicyModel} &=  
		\sum_{t=1}^{\NumFrames}
		|| \PolicyModel(\SlotsT{t}) - \PredLatentT{t} ||, \\
	\Loss_{\ActionDecoder} &= 
	\sum_{t=1}^{\NumFrames} ||\ActionDecoder(\PredLatentT{t}) - \ActionT{t}||.
\end{align}
\vspace{-0.0cm}


\begin{table*}[t!]
	\centering
	\caption{
		Quantitative evaluation of several object-centric (OC) and controllable (Cont.) video prediction models.
		Given six seed frames, all models predict the subsequent 15 frames.
		\Method~achieves the best results in datasets that require modeling object interactions (\RobotDB) or feature multiple moving objects (GridShapes), 
		while maintaining competitive performance on \ButtonPress{} and Sketchy.
		Best two results are highlighted in boldface and underlined, respectively.
	}
	\label{table:quantitative comparison}
	\vspace{-0.3cm}
	\footnotesize
	\addtolength{\tabcolsep}{-0.04em}
	\renewcommand{\arraystretch}{1.02}
	\begin{tabular}
		{
			p{1.25cm} 
			P{0.33cm}P{0.43cm}P{-0.02cm}
			P{0.65cm}P{0.65cm}P{0.55cm}P{0.0cm}
			P{0.65cm}P{0.65cm}P{0.55cm}P{0.0cm}
			P{0.65cm}P{0.65cm}P{0.55cm}P{0.0cm}
			P{0.65cm}P{0.65cm}P{0.75cm}
		}
		\toprule
		&&&&
		\multicolumn{4}{c}{\textbf{\RobotDB}} & \multicolumn{4}{c}{\textbf{\ButtonPress}} & 
		\multicolumn{4}{c}{\textbf{GridShapes$_{\text{2Objs}}$}} & \multicolumn{3}{c}{\textbf{Sketchy}}\\
		\cmidrule(r){5-8}\cmidrule(r){9-12}\cmidrule(r){13-15}\cmidrule(r){17-19}
		\textbf{Model} & \centering \textbf{OC} & \centering \textbf{Cont.} && 
		{ PSNR}$\uparrow$ & {SSIM}$\uparrow$ & {LPIPS}$\downarrow$ &&
		{ PSNR}$\uparrow$ & {SSIM}$\uparrow$ & {LPIPS}$\downarrow$ &&
		{ PSNR}$\uparrow$ & {SSIM}$\uparrow$ & {LPIPS}$\downarrow$ &&
		{ PSNR}$\uparrow$ & {SSIM}$\uparrow$ & {LPIPS}$\downarrow$ \\
		\midrule
		SVG & \hspace{0.01cm} \xmark & \hspace{0.1cm} \vmark && 20.96 & \underline{0.898} & 0.096 && \textbf{32.23} & \textbf{0.950} & \textbf{0.011} && 38.10 & \underline{0.988} & \underline{0.002} && \underline{24.24} & \underline{0.813} & 0.076 \\
		CADDY & \hspace{0.01cm} \xmark & \hspace{0.1cm} \vmark && \underline{21.18} & \textbf{0.901} & \underline{0.090} && 24.58 & 0.853 & 0.028 && \underline{39.16} & 0.986 & 0.006 && 23.65 & 0.809 & \textbf{0.059} \\
		SlotFormer & \hspace{0.01cm} \vmark & \hspace{0.1cm} \xmark && 17.22 & 0.729 & 0.134 && 19.52 & 0.762 & 0.111 && 19.74 & 0.795 & 0.149 && 21.17 & 0.751 & 0.091 \\
		OCVP & \hspace{0.01cm} \vmark & \hspace{0.1cm}  \xmark && 17.26 & 0.751 & 0.134 && 19.55 & 0.762 & 0.115 && 18.86 & 0.791 & 0.154 && 21.23 & 0.751 & 0.092 \\
		\Method & \hspace{0.01cm} \vmark & \hspace{0.1cm} \vmark && \textbf{21.41} & 0.890 & \textbf{0.066} && \underline{26.03} & \underline{0.878} & \underline{0.025} && \textbf{54.09} & \textbf{0.996} & \textbf{0.001} && \textbf{24.43} & \textbf{0.815} & \underline{0.063} \\
		\bottomrule
	\end{tabular}
	\vspace{-0.3cm}
\end{table*}

\section{Experiments}

\subsection{Experimental Setup}

\subsubsection{Datasets}

We evaluate our method on four environments with distinct characteristics.
Further details are provided in \Appendix{app: dataset}.

\noindent \textbf{\ButtonPress} This environment, based  on \emph{MetaWorld}~\cite{Yu_MetaworldDataset_2020}, features a Sawyer robot arm that must press a red button.
This environment depicts a non-object centric task involving complex shapes and textures.

\noindent\textbf{\RobotDB}: This environment, inspired by~\citet{Li_TowardsPracticalMultiObjectManipulationRelationalRL_2020}, features a robot arm and a table with multiple uni-colored cubes.
The robot must push the block of distinct color into the location specified by a red target.
This task evaluates relational reasoning and modeling of object collisions.

\noindent\textbf{GridShapes:} This dataset features two simple 2D shapes moving in grid-like patterns on a colored background.
The shapes randomly change direction, introducing stochasticity to their motion.
This dataset benchmarks a model's ability to jointly predict the motion of multiple moving agents.

\noindent\textbf{Sketchy:} Sketchy~\cite{Cabi_ScalingDataDrivenRoboticsRewardSketching_2020} is a real-world robotics dataset in which a robotic gripper interacts with diverse objects of different shape and color. This dataset evaluates the model performance on real-world robotics data.

\subsubsection{Implementation Details}

Our model is implemented in PyTorch~\cite{Paszke_AutomaticDifferneciationInPytorch_2017} and trained on a single NVIDIA A100 GPU.
\Method{} uses SAVi~\cite{Kipf_ConditionalObjectCentricLearningFromVideo_2022} with 128-dimensional object slots, as well as a convolutional encoder and spatial broadcast decoder as scene parsing and rendering modules, respectively.
The conditional predictor and inverse dynamics modules are transformer encoders with four layers and a token dimension of 256.
For \ButtonPress, \RobotDB{} and Sketchy we use the $\InverseDynamicsSingle$ variant with eight distinct 16-dimensional action prototypes, whereas for GridShapes we use $\InverseDynamicsMany$ with five 8-dimensional prototypes.
Further implementation details are provided in \Appendix{app: implementation details}.

\input{./imgs/qual_comb_v3.tex}

\subsection{Video Prediction}

We compare \Method{} for video prediction with the object-centric baselines SlotFormer~\cite{Wu_SlotFormer_2022} and OCVP-Seq~\cite{Villar_OCVP_2023}, the stochastic video prediction model SVG-LP~\cite{Denton_StochasticVideoGenerationWithALearnedPrior_2018} and the playable video generation model CADDY~\cite{Menapace_PlayableVideoGeneration_2021}.
For fair comparison, all models are trained with six seed frames to predict the subsequent eight, and evaluated for 15 predictions.
For CADDY, $\Method${} and SVG, we predict future frames conditioned on latent actions or vectors inferred from the ground truth sequence.
Additionally, on \RobotDB~and \ButtonPress~datasets all models are trained using sequences with random exploration policies, and evaluated on expert demonstrations.

We evaluate the quality of the predicted frames using PSNR, SSIM~\cite{Wang_SSIM_2004} and LPIPS~\cite{Zhang_TheUnreasonableEffectivenessOfDeepFeaturesLPIPS_2018}.
A quantitative comparison of the methods is presented in \Table{table:quantitative comparison}.
As expected, deterministic object-centric models (i.e. SlotFormer and OCVP) perform poorly, as they cannot infer the agent's actions and simply average over multiple possible futures.
Our proposed method outperforms all other models on both the \RobotDB~and GridShapes datasets, demonstrating \Method's superior ability to forecast future video frames in environments involving multiple object interactions and moving agents, respectively. 
On Sketchy, which features real robotics videos, we observe that \Method{}, CADDY and SVG achieve similar performance, with \Method{} slightly outperforming the baselines.
While the margin of improvement is modest, these results suggest that object-centric models like \Method{} are well-suited for controllable video prediction and can serve as world models in real robotic scenarios.

\Figure{fig: qual_01} depicts a qualitative comparison of the best performing methods on the \ButtonPress~and \RobotDB~datasets, respectively.
On the \ButtonPress~dataset, as shown in \Figure{fig: qual_01}a), all methods accurately model the motion of the robot arm.
However, on the more complex \RobotDB~task, depicted in \Figure{fig: qual_01}b), SVG and CADDY fail to model the object collisions, leading to blurriness and vanishing objects.
In contrast, \Method{} maintains sharp object representations and correctly models interactions between objects, leading to accurate frame predictions.
Further qualitative evaluations are provided in \Appendix{app: extra results}.

\input{./imgs/fig_gridshapes.tex}

\subsection{Model Analysis}

\noindent \textbf{Impact of Number of Moving Objects:}
We evaluate the performance of our method for different numbers of moving objects.
For this purpose, we train two \Method{} variants with the $\InverseDynamicsSingle$ and $\InverseDynamicsMany$ inverse dynamics modules, respectively, and compare them with the SVG and CADDY baselines on several variants of the GridShapes dataset featuring a different number of objects, ranging from one to five moving shapes.
The results are shown in \Figure{fig: gridshapes}.
CADDY and \Method{} with $\InverseDynamicsSingle$, which encode scene dynamics using a single latent action, perform strongly when forecasting one or two objects, but experience a sharp drop in performance as the number of objects increases.
SVG scales to multiple moving objects but encodes all dynamics into a single distribution, limiting its flexibility and control.
In contrast, \Method{} with $\InverseDynamicsMany$ uses a latent action per object, seamlessly scaling to a large number of moving agents by individually modeling the motion of each object, thus outperforming all baselines.


\begin{table}[t!]
	\centering
	\caption{
		Quantitative comparison of \Method~with variants using continuous and discrete latent actions, as well as an oracle model with access to the ground truth actions.
	}
	\label{table: ablation}
	\vspace{-0.3cm}
	\footnotesize
	\addtolength{\tabcolsep}{-0.2em}
	\renewcommand{\arraystretch}{1.}
	\begin{tabular}{P{1.6cm}p{2.2cm} P{0.95cm}P{0.95cm}P{0.95cm}}
		\toprule
		\multicolumn{2}{c}{} &  \multicolumn{3}{c}{\textbf{Results}} \\
		\cmidrule(r){3-5} 
		\textbf{RobotDB} &\textbf{Model Variant} & {PSNR}$\uparrow$ & {SSIM}$\uparrow$ & {LPIPS}$\downarrow$ \\
		\midrule
		\multirow{4}{*}{\makecell{\textbf{Random}\\\textbf{Exploration}}} & \Method & \underline{26.64} & \underline{0.944} & \textbf{0.016} \\
		&\hspace{0.2cm} w/ Cont. $\PredLatent$ & 26.24 & 0.924 & 0.019 \\
		&\hspace{0.2cm} w/ Discrete $\PredLatent$ & 20.60 & 0.849 & 0.040 \\
		&\hspace{0.2cm} w/ GT Actions & \textbf{27.77} & \textbf{0.955} & \textbf{0.016} \\
		\midrule
		\multirow{4}{*}{\makecell{\textbf{Expert}\\\textbf{Demos.}}} & \Method & 21.41 & 0.890 & 0.065 \\
		&\hspace{0.2cm} w/ Cont. $\PredLatent$ & \underline{22.00} & \underline{0.900} & \underline{0.061} \\
		&\hspace{0.2cm} w/ Discrete $\PredLatent$ & 18.00 & 0.791 & 0.109 \\
		&\hspace{0.2cm} w/ GT Actions & \textbf{22.30} & \textbf{0.904} & \textbf{0.058} \\
		\bottomrule
	\end{tabular}
\vspace{-0.6cm}
\end{table}

\noindent \textbf{Action Representation:}
In \Table{table: ablation} we compare the hybrid latent action representation used in \Method{} with three different variants that use continuous latent actions, a discrete set of latent actions, and an oracle variant with access to ground truth actions.
Additionally, we evaluate the models on two different \RobotDB~variants, including
a set with random exploration sequences, similar to the training distribution, and another set featuring expert demonstrations.

On the random exploration set, \Method{} with a hybrid latent action performs comparably to the oracle, highlighting the ability of our $\InverseDynamics$ module to infer latent dynamics from unlabeled sequences.
However, \Method's performance drops on the expert demonstrations, with the variant using unconstrained continuous latent actions outperforming it.
We attribute this to the large discrepancy between the action distribution of expert sequences and the training data, which challenges the generalization of the learned action prototypes.
In both cases, the variant with only discrete latent actions performs the worst, likely due to its limited flexibility in modeling the fine-grained dynamics of robot motion.
Nonetheless, our hybrid action representation remains competitive, achieving performance close to the best methods despite the distribution shift.

\input{./imgs/action_fig_00.tex}

\noindent \textbf{Learned Action Prototypes:}
\Figure{fig: action_figs robot} illustrates the effect of different action prototypes learned by \Method{} on the \RobotDB~dataset.
Given a single seed frame, we forecast 15 frames by repeatedly conditioning the predictor on the same action prototype.
We also visualize the predictions obtained using the latent actions inferred by our inverse dynamics module from the ground truth sequence.
$\Method${} effectively learns to infer precise robot actions from visual observations, as well as the physics of interacting objects.
Additionally, \Figure{fig: action_figs robot} shows that \Method{} learns consistent and semantically meaningful actions, such as moving the robot towards the right (action 2), left (action 7), or upwards (action 4).
Finally, the instance segmentation maps, obtained by assigning a different color to each slot mask, demonstrate how \Method{} distinctly represents each object into a separate slot, clearly disentangling each individual block, the robot and the background.

\vspace{0.2cm}
\noindent \textbf{Real-World Robotic Videos:}
We validate the applicability of \Method{} to real-world robotic videos using the Sketchy~\cite{Cabi_ScalingDataDrivenRoboticsRewardSketching_2020} dataset.
\Figure{fig: sketchy00} presents a qualitative assessment demonstrating \Method{}’s ability to accurately infer the scene’s inverse dynamics from a real robotic demonstration.
Starting from a single reference frame, our model reconstructs the ground truth sequence by predicting the robot’s movement, rotation, and gripper opening.
Furthermore, \Method{} learns semantically meaningful and temporally consistent action prototypes that capture diverse robot behaviors such as opening the gripper (action 6), moving downwards (action 2), or upwards (action 4).

\input{./imgs/qual_sketchy_00.tex}

\subsection{Learning Behaviors from Expert Demonstrations}
\label{sec: behavior}

We evaluate the quality of $\Method$'s object-centric representations and inferred latent actions for a downstream behavior learning task.
To this end, we train a policy model and an action decoder, as described in~\Section{sec: behavior learning}, to imitate \ButtonPress~and~\RobotDB~behaviors from a limited set of expert demonstrations.
Furthermore, we train the models using three different random seeds and show the mean success rate and its standard deviation.

\noindent \textbf{\ButtonPress~Behavior~~}
We compare our learned policy, which imitates the \ButtonPress~behavior, against oracle baselines with access to all expert demonstrations.
These models directly regress the ground-truth actions from object-centric slots (Slot Oracle), ResNet feature (ResNet Oracle), and from feature maps output by the CADDY encoder (CADDY Oracle), respectively.
In contrast, \Method{} autoregressively predicts latent actions within its latent imagination before decoding them into executable actions.
We also evaluate a modified CADDY variant with a lightweight policy and action decoder that maps observed features to real-world actions.

\begin{figure}[t!]
	\resizebox{1.0 \linewidth}{!}{
	\begin{tikzpicture} 
		%
		%
		\node(plot_0)[anchor=north, inner sep=0, outer sep=0] at (0,0.0) 
		{\includegraphics[width=\linewidth]
			{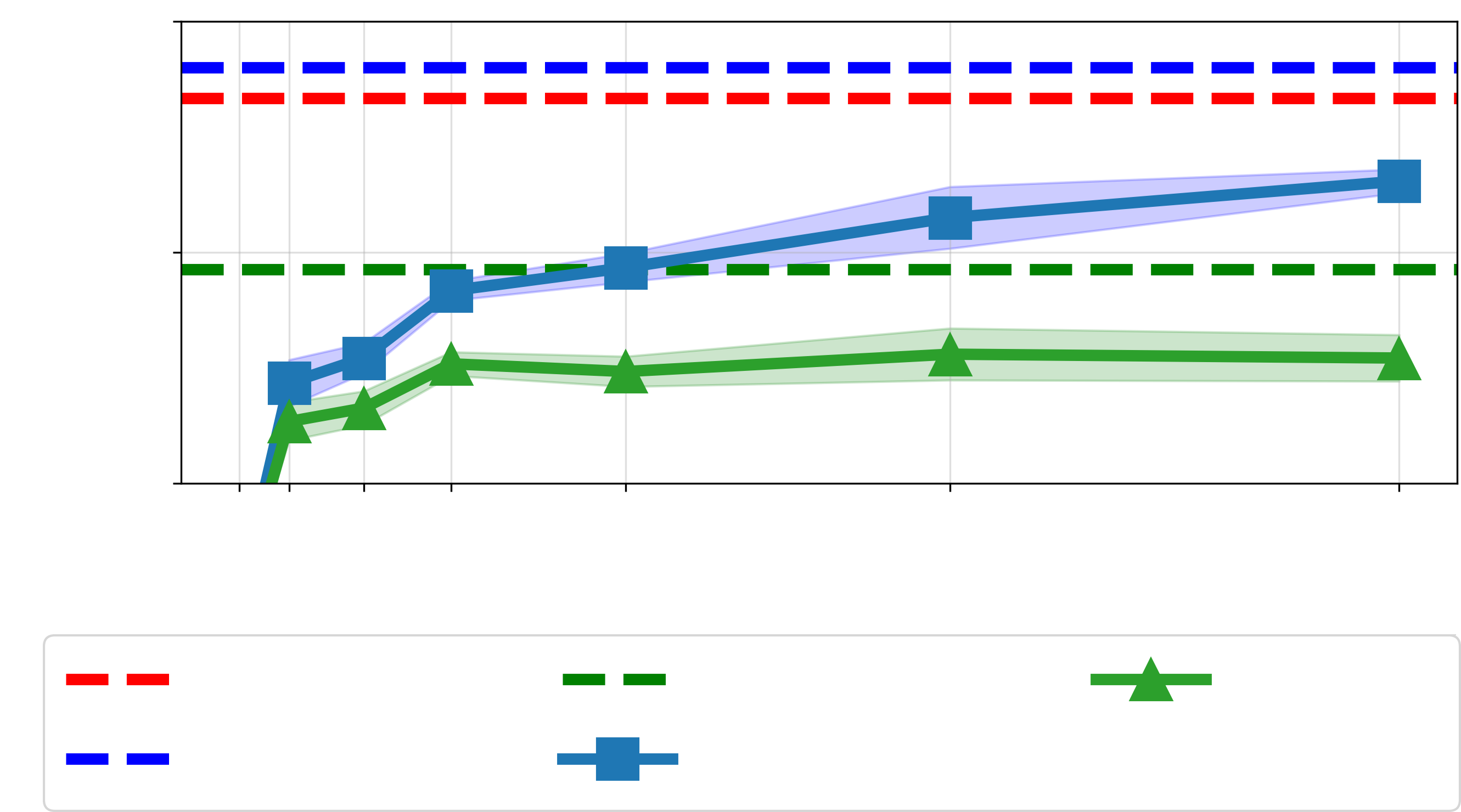}
		};
		\node(a)[anchor=north west,draw=none,ultra thick,inner sep=0, outer sep=0] at([yshift=1.5cm, xshift=0.1cm]plot_0.south west)
		{\large a) };
		\node(a)[anchor=north,draw=none,ultra thick,inner sep=0, outer sep=0] at([yshift=1.37cm, xshift=0.55cm]plot_0.south)
		{\# Expert Demonstrations };
		\node(a)[anchor=north east,draw=none,ultra thick,inner sep=0, outer sep=0] at([yshift=4.4cm, xshift=0.3cm]plot_0.south west)
		{\rotatebox{90}{Success Rate [\%]}};
		\node(title)[anchor=south,draw=none,ultra thick,inner sep=0, outer sep=0] at([yshift=-0.01cm, xshift=0.55cm]plot_0.north)
		{\textbf{\ButtonPress{} Behavior}};
		\node(legend1)[anchor=north,draw=none,ultra thick,inner sep=0, outer sep=0] at([yshift=0.86cm, xshift=-2.10cm]plot_0.south){\customsize{ResNet Orac.}};
		\node(legend2)[anchor=north,draw=none,ultra thick,inner sep=0, outer sep=0] at([yshift=0.86cm, xshift=0.80cm]plot_0.south){\customsize{CADDY Orac.}};
		\node(legend3)[anchor=north,draw=none,ultra thick,inner sep=0, outer sep=0] at([yshift=0.86cm, xshift=3.40cm]plot_0.south){\customsize{CADDY}};
		\node(legend4)[anchor=north west,draw=none,ultra thick,inner sep=0, outer sep=0] at([yshift=-0.24cm, xshift=0.0cm]legend1.south west){\customsize{Slot Orac.}};
		\node(legend5)[anchor=north west,draw=none,ultra thick,inner sep=0, outer sep=0] at([yshift=-0.24cm, xshift=0.01cm]legend2.south west){\customsize{\Method}};
		\node(lab0)[anchor=east,draw=none,ultra thick,inner sep=0, outer sep=0] at([yshift=-0.1cm, xshift=0.92cm]plot_0.north west){\labelsize{100}};
		\node(lab0)[anchor=east,draw=none,ultra thick,inner sep=0, outer sep=0] at([yshift=0.88cm, xshift=0.92cm]plot_0.west){\labelsize{60}};
		\node(lab0)[anchor=east,draw=none,ultra thick,inner sep=0, outer sep=0] at([yshift=1.9cm, xshift=0.92cm]plot_0.south west){\labelsize{20}};
		\node(lab0)[anchor=north,draw=none,ultra thick,inner sep=0, outer sep=0]at([yshift=1.77cm, xshift=-2.8cm]plot_0.south){\labelsize{0}};
		\node(lab0)[anchor=north,draw=none,ultra thick,inner sep=0, outer sep=0]at([yshift=1.77cm, xshift=-2.5cm]plot_0.south){\labelsize{10}};
		\node(lab0)[anchor=north,draw=none,ultra thick,inner sep=0, outer sep=0]at([yshift=1.77cm, xshift=-2.1cm]plot_0.south){\labelsize{50}};
		\node(lab0)[anchor=north,draw=none,ultra thick,inner sep=0, outer sep=0]at([yshift=1.77cm, xshift=-1.58cm]plot_0.south){\labelsize{100}};
		\node(lab0)[anchor=north,draw=none,ultra thick,inner sep=0, outer sep=0]at([yshift=1.77cm, xshift=-0.6cm]plot_0.south){\labelsize{200}};
		\node(lab0)[anchor=north,draw=none,ultra thick,inner sep=0, outer sep=0]at([yshift=1.77cm, xshift=1.22cm]plot_0.south){\labelsize{500}};
		\node(lab0)[anchor=north,draw=none,ultra thick,inner sep=0, outer sep=0]at([yshift=1.77cm, xshift=3.78cm]plot_0.south){\labelsize{900}};
		%
		%
		%
		%
		%
		%
		\node(plot_1)[anchor=north, inner sep=0, outer sep=0] at ([yshift=-0.65cm, xshift=0cm]plot_0.south) 
		{\includegraphics[width=\linewidth]
			{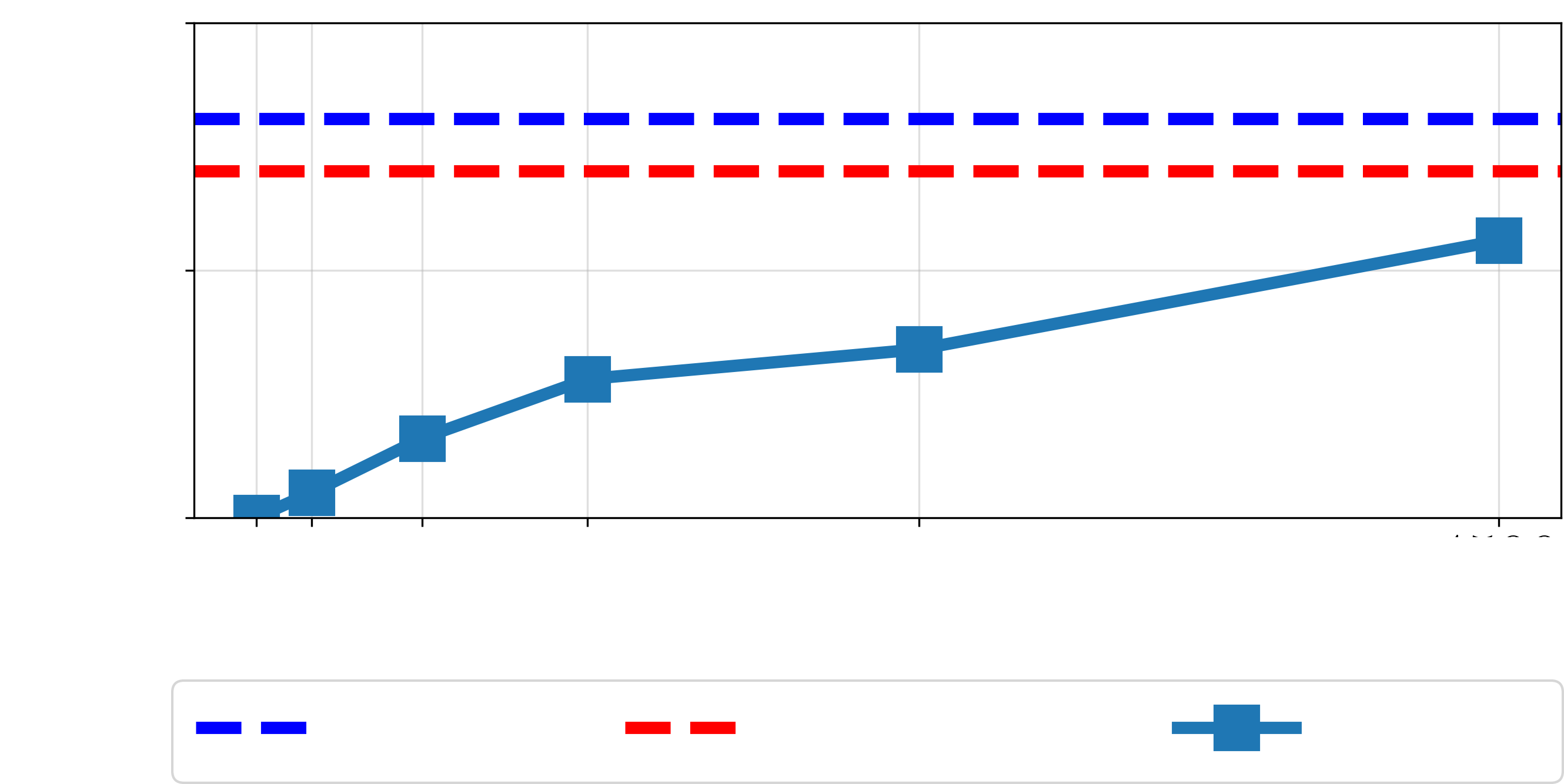}
		};
		\node(b)[anchor=north west,draw=none,ultra thick,inner sep=0, outer sep=0] at([yshift=0.8cm, xshift=0.1cm]plot_1.south west)
		{\large b) };
		\node(title)[anchor=south,draw=none,ultra thick,inner sep=0, outer sep=0] at([yshift=-0.01cm, xshift=0.55cm]plot_1.north)
		{\textbf{\RobotDB{} Behavior}};
		\node(a)[anchor=north,draw=none,ultra thick,inner sep=0, outer sep=0] at([yshift=0.9cm, xshift=0.55cm]plot_1.south)
		{\# Expert Demonstrations };
		\node(a)[anchor=north east,draw=none,ultra thick,inner sep=0, outer sep=0] at([yshift=3.9cm, xshift=0.35cm]plot_1.south west)
		{\rotatebox{90}{Success Rate [\%]}};
		\node(legend1)[anchor=north,draw=none,ultra thick,inner sep=0, outer sep=0] at([yshift=0.41cm, xshift=-1.96cm]plot_1.south){\customsize{SOLD}};
		\node(legend2)[anchor=north,draw=none,ultra thick,inner sep=0, outer sep=0] at([yshift=0.41cm, xshift=0.75cm]plot_1.south){\customsize{DreamerV3}};
		\node(legend3)[anchor=north,draw=none,ultra thick,inner sep=0, outer sep=0] at([yshift=0.41cm, xshift=3.42cm]plot_1.south){\customsize{\Method}};
		\node(lab0)[anchor=east,draw=none,ultra thick,inner sep=0, outer sep=0] at([yshift=-0.12cm, xshift=0.92cm]plot_1.north west){\labelsize{100}};
		\node(lab0)[anchor=east,draw=none,ultra thick,inner sep=0, outer sep=0] at([yshift=0.63cm, xshift=0.92cm]plot_1.west){\labelsize{50}};
		\node(lab0)[anchor=east,draw=none,ultra thick,inner sep=0, outer sep=0] at([yshift=1.4cm, xshift=0.92cm]plot_1.south west){\labelsize{0}};
		\node(lab0)[anchor=north,draw=none,ultra thick,inner sep=0, outer sep=0]at([yshift=1.25cm, xshift=-2.8cm]plot_1.south){\labelsize{0}};
		\node(lab0)[anchor=north,draw=none,ultra thick,inner sep=0, outer sep=0]at([yshift=1.25cm, xshift=-2.45cm]plot_1.south){\labelsize{200}};
		\node(lab0)[anchor=north,draw=none,ultra thick,inner sep=0, outer sep=0]at([yshift=1.25cm, xshift=-1.9cm]plot_1.south){\labelsize{500}};
		\node(lab0)[anchor=north,draw=none,ultra thick,inner sep=0, outer sep=0]at([yshift=1.25cm, xshift=-1.05cm]plot_1.south){\labelsize{1000}};
		\node(lab0)[anchor=north,draw=none,ultra thick,inner sep=0, outer sep=0]at([yshift=1.25cm, xshift=0.7cm]plot_1.south){\labelsize{2000}};
		\node(lab0)[anchor=north,draw=none,ultra thick,inner sep=0, outer sep=0]at([yshift=1.25cm, xshift=3.78cm]plot_1.south){\labelsize{4500}};
	\end{tikzpicture}
	}
	\vspace{-0.5cm}
	\caption{
		 Success rate (\%) as a function of available expert demonstrations on the a) \ButtonPress{} and b) \RobotDB{} environments.
		 Object-centric models (Slot Oracle, \Method{} and SOLD) outperform their counterparts. \Method{} consistently improves with more demonstrations.		 
	}
	\label{fig: quant_beh_paper}
	\vspace{-0.3cm}
\end{figure}

As shown in \Figure{fig: quant_beh_paper}a), \Method{} consistently outperforms CADDY across all training regimes.
With as few as 200 demonstrations, \Method{} matches the performance of the CADDY Oracle, which has direct
access to more expert demonstrations and ground-truth actions.
As the number of demonstrations increases, \Method{} continues to improve, approaching the performance of the oracle models, while CADDY struggles to generalize.
These findings highlight the effectiveness of \Method's object-centric representations in learning robot behaviors from limited expert data.
Additionally, we observe that slot-based models outperform their holistic counterparts, further emphasizing the advantage of structured object-centric representations for learning robot behaviors.

\noindent \textbf{\RobotDB~Behavior~~}
We evaluate a policy learned using \Method{} on the challenging \RobotDB~task, which requires reasoning over object properties.
Our learned policy imitates the behavior from a limited number of expert demonstrations, with only approximately 80\% being successful.
The imperfect nature of the expert data adds an extra layer of difficulty to the learning process.

We compare \Method{} to two model-based reinforcement learning baselines with holistic (DreamerV3~\cite{hafner2023mastering}) and object-centric (SOLD~\cite{Mosbach_SOLDReinforcementLearningSlotObjectCentricLatentDynamics}) latent spaces, both of which learn the robot behavior by interacting with the environment.
As shown in \Figure{fig: quant_beh_paper}b), \Method{} closes the gap with the baseline models as the number of available demonstrations increases, demonstrating its ability to generalize from limited data.

\noindent \textbf{Visualization of Learned Behaviors~~}
\Figure{fig: behavior} illustrates the learned latent policies in the (a) \ButtonPress~and (b) \RobotDB~environments, respectively.
The top row in each sequence (labeled \emph{Latent Pred.}) shows predicted trajectories within \Method's latent imagination, where the model, starting from a single reference frame, autoregressively generates latent actions using the policy model and predicts future scene states in the latent space.
The bottom row (labeled \emph{Sim.~Actions}) visualizes the simulated execution of the decoded latent actions in the respective environments.
\Method{} learns to solve both tasks within its latent imagination, successfully reasoning over object properties and producing coherent sequences of latent actions that can be decoded into executable motions.

\input{./imgs/beh_button_00.tex}

\section{Conclusion}

We introduced $\Method$, a novel framework for controllable object-centric video prediction.
$\Method$~parses video frames into object slots, infers the scene's inverse dynamics, and predicts future object states and video frames by modeling the object dynamics and interactions, conditioned on inferred latent actions.
Through extensive experiments, we demonstrated that $\Method$ learns a semantically rich and meaningful action space, allowing for accurate video frame predictions.
Our method outperforms several baseline models, offering superior controllability and interpretability.
We further demonstrated that the learned object-centric representations and inferred latent actions can be utilized to predict future frames with precise robot control, while also enabling learning complex robot behaviors from limited, unlabeled demonstrations---outperforming methods that rely on holistic scene encoding.
These results highlight \Method{} as a versatile and interpretable world model, well-suited for a wide range of tasks in autonomous systems.

\clearpage

\section*{Acknowledgment}

This work was funded by grant BE 2556/16-2 (Research Unit FOR 2535 Anticipating Human Behavior) of the German Research Foundation (DFG).
Computational resources were provided by the German AI Service Center WestAI.

\section*{Impact Statement}

This paper presents work whose goal is to advance the field of machine learning by introducing \Method{}---an object-centric video prediction model that forecasts future video frames conditioned on inferred object-centric representations and latent actions, enhancing its interpretability and controllability, as well as learning representations that can be used for action planning.
This advancement is particularly relevant for domains such as robotics and autonomous systems, where understanding and predicting complex environments are essential for correct and safe operation.
However, the deployment of such models in real systems requires careful consideration for ethical and safety challenges, such as biases in the training data or lack of transparency in the decision-making process.

\bibliography{referencesAngel}

\begin{thebibliography}{60}
\providecommand{\natexlab}[1]{#1}
\providecommand{\url}[1]{\texttt{#1}}
\expandafter\ifx\csname urlstyle\endcsname\relax
  \providecommand{\doi}[1]{doi: #1}\else
  \providecommand{\doi}{doi: \begingroup \urlstyle{rm}\Url}\fi

\bibitem[Assouel et~al.(2022)Assouel, Castrejon, Courville, Ballas, and
  Bengio]{Assouel_VariationIdependentModulesVP_2022}
Assouel, R., Castrejon, L., Courville, A., Ballas, N., and Bengio, Y.
\newblock Vim: Variational independent modules for video prediction.
\newblock In \emph{Conference on Causal Learning and Reasoning}, pp.\  70--89.
  PMLR, 2022.

\bibitem[Aydemir et~al.(2023)Aydemir, Xie, and
  Guney]{Aydemir_SelfSupervisedObjectCentricLearningForVideos_2023}
Aydemir, G., Xie, W., and Guney, F.
\newblock Self-supervised object-centric learning for videos.
\newblock In \emph{Advances in Neural Information Processing Systems
  (NeurIPS)}, 2023.

\bibitem[Bao et~al.(2023)Bao, Tokmakov, Wang, Gaidon, and
  Hebert]{Bao_ObjectDiscoverMotion_2023}
Bao, Z., Tokmakov, P., Wang, Y.-X., Gaidon, A., and Hebert, M.
\newblock Object discovery from motion-guided tokens.
\newblock In \emph{IEEE/CVF Conference on Computer Vision and Pattern
  Recognition (2023)}, 2023.

\bibitem[Bengio et~al.(2013)Bengio, Courville, and
  Vincent]{Bengio_RepresentationLearningReview_2013}
Bengio, Y., Courville, A., and Vincent, P.
\newblock Representation learning: A review and new perspectives.
\newblock \emph{IEEE {T}ransactions on {P}attern {A}nalysis and {M}achine
  {I}ntelligence ({TPAMI})}, 35\penalty0 (8):\penalty0 1798--1828, 2013.

\bibitem[Biza et~al.(2023)Biza, Van~Steenkiste, Sajjadi, Elsayed, Mahendran,
  and Kipf]{Biza_InvariantSlotAttention_2023}
Biza, O., Van~Steenkiste, S., Sajjadi, M.~S., Elsayed, G.~F., Mahendran, A.,
  and Kipf, T.
\newblock Invariant slot attention: Object discovery with slot-centric
  reference frames.
\newblock In \emph{International Conference on Machine Learning (ICML)}, 2023.

\bibitem[Brandfonbrener et~al.(2024)Brandfonbrener, Nachum, and
  Bruna]{Brandfonbrener_InverseDynamicsPretrainingLearnsGoodRepresentatiosForImitation_2024}
Brandfonbrener, D., Nachum, O., and Bruna, J.
\newblock Inverse dynamics pretraining learns good representations for
  multitask imitation.
\newblock In \emph{Advances in Neural Information Processing Systems
  (NeurIPS)}, volume~36, 2024.

\bibitem[Bruce et~al.(2024)Bruce, Dennis, Edwards, Parker-Holder, Shi, Hughes,
  Lai, Mavalankar, Steigerwald, Apps,
  et~al.]{Bruce_GenieGenerativeInteractiveEnvironments_2024}
Bruce, J., Dennis, M.~D., Edwards, A., Parker-Holder, J., Shi, Y., Hughes, E.,
  Lai, M., Mavalankar, A., Steigerwald, R., Apps, C., et~al.
\newblock Genie: Generative interactive environments.
\newblock In \emph{International Conference on Machine Learning (ICML)}, 2024.

\bibitem[Burgess et~al.(2019)Burgess, Matthey, Watters, Kabra, Higgins,
  Botvinick, and
  Lerchner]{Burgess_MonetUnsupervisedSceneDecompositionRepresentation_2019}
Burgess, C.~P., Matthey, L., Watters, N., Kabra, R., Higgins, I., Botvinick,
  M., and Lerchner, A.
\newblock Monet: Unsupervised scene decomposition and representation.
\newblock \emph{arXiv:1901.11390}, 2019.

\bibitem[Cabi et~al.(2020)Cabi, Colmenarejo, Novikov, Konyushkova, Reed, Jeong,
  Zolna, Aytar, Budden, Vecerik,
  et~al.]{Cabi_ScalingDataDrivenRoboticsRewardSketching_2020}
Cabi, S., Colmenarejo, S.~G., Novikov, A., Konyushkova, K., Reed, S., Jeong,
  R., Zolna, K., Aytar, Y., Budden, D., Vecerik, M., et~al.
\newblock Scaling data-driven robotics with reward sketching and batch
  reinforcement learning.
\newblock In \emph{Robotics: Science and Systems (RSS)}, 2020.

\bibitem[Cho et~al.(2014)Cho, van Merri{\"e}nboer, Gulcehre, Bahdanau,
  Bougares, Schwenk, and Bengio]{Cho_GRU_2014}
Cho, K., van Merri{\"e}nboer, B., Gulcehre, C., Bahdanau, D., Bougares, F.,
  Schwenk, H., and Bengio, Y.
\newblock Learning phrase representations using {RNN} encoder{--}decoder for
  statistical machine translation.
\newblock In \emph{Conference on Empirical Methods in Natural Language
  Processing ({EMNLP})}, 2014.

\bibitem[Creswell et~al.(2021)Creswell, Kabra, Burgess, and
  Shanahan]{Creswell_UnsupervisedObjectBasedTransitionModelsFor3DPartiallyObservableEnvironments_2021}
Creswell, A., Kabra, R., Burgess, C., and Shanahan, M.
\newblock Unsupervised object-based transition models for 3{D} partially
  observable environments.
\newblock \emph{Advances in Neural Information Processing Systems (NeurIPS)},
  2021.

\bibitem[Cui et~al.(2024)Cui, Pan, Iyer, Haldar, and
  Pinto]{Cui_DynaMoDynamcisPretrainingForVisuoMotorControl_2024}
Cui, Z.~J., Pan, H., Iyer, A., Haldar, S., and Pinto, L.
\newblock {DynaMo: In-Domain Dynamics Pretraining for Visuo-Motor Control}.
\newblock In \emph{Advances in Neural Information Processing Systems
  (NeurIPS)}, 2024.

\bibitem[Daniel \& Tamar(2024)Daniel and Tamar]{Daniel_DDLP_2024}
Daniel, T. and Tamar, A.
\newblock {DDLP: Unsupervised Object-centric Video Prediction with Deep Dynamic
  Latent Particles}.
\newblock \emph{Transactions on Machine Learning Research (TMLR)}, 2024.

\bibitem[Denton \& Fergus(2018)Denton and
  Fergus]{Denton_StochasticVideoGenerationWithALearnedPrior_2018}
Denton, E. and Fergus, R.
\newblock Stochastic video generation with a learned prior.
\newblock In \emph{{I}nternational {C}onference on {M}achine {L}earning
  (ICML)}, 2018.

\bibitem[Dittadi et~al.(2022)Dittadi, Papa, De~Vita, Sch{\"o}lkopf, Winther,
  and
  Locatello]{Dittadi_GeneralizationAndRobustnessImplicationsInObjectCentricLearning_2022}
Dittadi, A., Papa, S., De~Vita, M., Sch{\"o}lkopf, B., Winther, O., and
  Locatello, F.
\newblock Generalization and robustness implications in object-centric
  learning.
\newblock In \emph{International Conference on Machine Learning (ICML)}, 2022.

\bibitem[Edwards et~al.(2019)Edwards, Sahni, Schroecker, and
  Isbell]{Edwards_ImitatingLatentPoliciesFromObservation_2019}
Edwards, A., Sahni, H., Schroecker, Y., and Isbell, C.
\newblock Imitating latent policies from observation.
\newblock In \emph{International Conference on Machine Learning (ICML)}, 2019.

\bibitem[Elsayed et~al.(2022)Elsayed, Mahendran, van Steenkiste, Greff, Mozer,
  and
  Kipf]{Elsayed_SAVi++TowardsEndToEndObjectCentricLearningFromRealWorldVideos_2022}
Elsayed, G.~F., Mahendran, A., van Steenkiste, S., Greff, K., Mozer, M.~C., and
  Kipf, T.
\newblock {SAV}i++: Towards end-to-end object-centric learning from real-world
  videos.
\newblock In \emph{Advances in Neural Information Processing Systems
  (NeurIPS)}, 2022.

\bibitem[Ferraro et~al.(2023)Ferraro, Mazzaglia, Verbelen, and
  Dhoedt]{Ferraro_Focus_2023}
Ferraro, S., Mazzaglia, P., Verbelen, T., and Dhoedt, B.
\newblock Focus: Object-centric world models for robotics manipulation.
\newblock In \emph{Advances in Neural Information Processing Systems Workshops
  (NeurIPSw)}, 2023.

\bibitem[Greff et~al.(2020)Greff, Van~Steenkiste, and
  Schmidhuber]{Greff_OnTheBindingProblemInNeuralNetworks_2020}
Greff, K., Van~Steenkiste, S., and Schmidhuber, J.
\newblock On the binding problem in artificial neural networks.
\newblock \emph{arXiv preprint arXiv:2012.05208}, 2020.

\bibitem[Hafner et~al.(2023)Hafner, Pasukonis, Ba, and
  Lillicrap]{hafner2023mastering}
Hafner, D., Pasukonis, J., Ba, J., and Lillicrap, T.
\newblock Mastering diverse domains through world models.
\newblock \emph{arXiv preprint arXiv:2301.04104}, 2023.

\bibitem[Jeong et~al.(2025)Jeong, Chun, Cha, and
  Kim]{Jeong_ObjectCentricWorldModelForLanguageGuidedManipulation_2025}
Jeong, Y., Chun, J., Cha, S., and Kim, T.
\newblock Object-centric world model for language-guided manipulation.
\newblock \emph{arXiv preprint arXiv:2503.06170}, 2025.

\bibitem[Jiang et~al.(2023)Jiang, Deng, Singh, and
  Ahn]{Jiang_ObjectCentricSlotDiffusion_2023}
Jiang, J., Deng, F., Singh, G., and Ahn, S.
\newblock Object-centric slot diffusion.
\newblock In \emph{Advances in Neural Information Processing Systems
  (NeurIPS)}, 2023.

\bibitem[Jiang et~al.(2024)Jiang, Deng, Singh, Lee, and
  Ahn]{Jiang_SlotSSM_2024}
Jiang, J., Deng, F., Singh, G., Lee, M., and Ahn, S.
\newblock {SlotSSMs: Slot State Space Models}.
\newblock In \emph{Advances in Neural Information Processing Systems
  (NeurIPS)}, 2024.

\bibitem[Johnson(2018)]{Johnson_ObjectPerception_2018}
Johnson, S.~P.
\newblock Object perception.
\newblock In \emph{Oxford Research Encyclopedia of Psychology}. Oxford
  University Press, 2018.

\bibitem[Kahneman et~al.(1992)Kahneman, Treisman, and
  Gibbs]{Kahneman_ReviewingOfObjectFiles_1992}
Kahneman, D., Treisman, A., and Gibbs, B.~J.
\newblock The reviewing of object files: Object-specific integration of
  information.
\newblock \emph{Cognitive psychology}, 24\penalty0 (2):\penalty0 175--219,
  1992.

\bibitem[Kingma \& Ba(2015)Kingma and Ba]{Kingma_Adam_2014}
Kingma, D.~P. and Ba, J.
\newblock A method for stochastic optimization.
\newblock In \emph{International Conference on Learning Representations
  ({ICLR})}, 2015.

\bibitem[Kipf et~al.(2022)Kipf, Elsayed, Mahendran, Stone, Sabour, Heigold,
  Jonschkowski, Dosovitskiy, and
  Greff]{Kipf_ConditionalObjectCentricLearningFromVideo_2022}
Kipf, T., Elsayed, G.~F., Mahendran, A., Stone, A., Sabour, S., Heigold, G.,
  Jonschkowski, R., Dosovitskiy, A., and Greff, K.
\newblock {Conditional Object-Centric Learning from Video}.
\newblock In \emph{International Conference on Learning Representations
  (ICLR)}, 2022.

\bibitem[Li et~al.(2020)Li, Jabri, Darrell, and
  Agrawal]{Li_TowardsPracticalMultiObjectManipulationRelationalRL_2020}
Li, R., Jabri, A., Darrell, T., and Agrawal, P.
\newblock Towards practical multi-object manipulation using relational
  reinforcement learning.
\newblock In \emph{IEEE International Conference on Robotics and Automation
  (ICRA)}, 2020.

\bibitem[Locatello et~al.(2019)Locatello, Bauer, Lucic, Raetsch, Gelly,
  Sch{\"o}lkopf, and
  Bachem]{Locatello_ChallengingAssumptionsInLearningOfDissentangledRepresentations_2019}
Locatello, F., Bauer, S., Lucic, M., Raetsch, G., Gelly, S., Sch{\"o}lkopf, B.,
  and Bachem, O.
\newblock Challenging common assumptions in the unsupervised learning of
  disentangled representations.
\newblock In \emph{{I}nternational {C}onference on {M}achine {L}earning
  ({ICML})}, 2019.

\bibitem[Locatello et~al.(2020)Locatello, Weissenborn, Unterthiner, Mahendran,
  Heigold, Uszkoreit, Dosovitskiy, and
  Kipf]{Locatello_ObjectCentricLearningWithSlotAttention_2020}
Locatello, F., Weissenborn, D., Unterthiner, T., Mahendran, A., Heigold, G.,
  Uszkoreit, J., Dosovitskiy, A., and Kipf, T.
\newblock Object-centric learning with slot attention.
\newblock In \emph{International Conference on Neural Information Processing
  Systems ({NeurIPS})}, 2020.

\bibitem[Mamaghan et~al.(2025)Mamaghan, Papa, Johansson, Bauer, and
  Dittadi]{Mamaghan_ObjectCentricRepsForVQA_2025}
Mamaghan, A. M.~K., Papa, S., Johansson, K.~H., Bauer, S., and Dittadi, A.
\newblock Exploring the effectiveness of object-centric representations in
  visual question answering: Comparative insights with foundation models.
\newblock In \emph{International Conference on Learning Representations
  (ICLR)}, 2025.

\bibitem[Menapace et~al.(2021)Menapace, Lathuiliere, Tulyakov, Siarohin, and
  Ricci]{Menapace_PlayableVideoGeneration_2021}
Menapace, W., Lathuiliere, S., Tulyakov, S., Siarohin, A., and Ricci, E.
\newblock Playable video generation.
\newblock In \emph{IEEE/CVF Conference on Computer Vision and Pattern
  Recognition (CVPR)}, 2021.

\bibitem[Menapace et~al.(2022)Menapace, Lathuili{\`e}re, Siarohin, Theobalt,
  Tulyakov, Golyanik, and Ricci]{Menapace_PlayableEnvironments_2022}
Menapace, W., Lathuili{\`e}re, S., Siarohin, A., Theobalt, C., Tulyakov, S.,
  Golyanik, V., and Ricci, E.
\newblock Playable environments: Video manipulation in space and time.
\newblock In \emph{IEEE/CVF Conference on Computer Vision and Pattern
  Recognition (CVPR)}, 2022.

\bibitem[Meo et~al.(2024)Meo, Nakano, Lic{\u{a}}, Didolkar, Suzuki, Goyal,
  Zhang, Dauwels, Matsuo, and
  Bengio]{Meo_ObjectCentricTemporalConsistency_2024}
Meo, C., Nakano, A., Lic{\u{a}}, M., Didolkar, A., Suzuki, M., Goyal, A.,
  Zhang, M., Dauwels, J., Matsuo, Y., and Bengio, Y.
\newblock Object-centric temporal consistency via conditional autoregressive
  inductive biases.
\newblock \emph{arXiv preprint arXiv:2410.15728}, 2024.

\bibitem[Mosbach et~al.(2025)Mosbach, Niklas~Ewertz, Villar-Corrales, and
  Behnke]{Mosbach_SOLDReinforcementLearningSlotObjectCentricLatentDynamics}
Mosbach, M., Niklas~Ewertz, J., Villar-Corrales, A., and Behnke, S.
\newblock {SOLD: Slot Object-Centric Latent Dynamics Models for Relational
  Manipulation Learning from Pixels}.
\newblock In \emph{International Conference on Machine Learning (ICML)}, 2025.

\bibitem[Paszke et~al.(2017)Paszke, Gross, Chintala, Chanan, Yang, DeVito, Lin,
  Desmaison, Antiga, and Lerer]{Paszke_AutomaticDifferneciationInPytorch_2017}
Paszke, A., Gross, S., Chintala, S., Chanan, G., Yang, E., DeVito, Z., Lin, Z.,
  Desmaison, A., Antiga, L., and Lerer, A.
\newblock Automatic differentiation in {P}y{T}orch.
\newblock In \emph{International Conference on Neural Information Processing
  Systems Workshops ({NeurIPSw})}, 2017.

\bibitem[Schmidt \& Jiang(2024)Schmidt and
  Jiang]{Schmidt_LAPOLearningToActWithoutActions_2024}
Schmidt, D. and Jiang, M.
\newblock Learning to act without actions.
\newblock In \emph{International Conference on Learning Representations
  (ICLR)}, 2024.

\bibitem[Seitzer et~al.(2023)Seitzer, Horn, Zadaianchuk, Zietlow, Xiao,
  Simon-Gabriel, He, Zhang, Sch{\"o}lkopf, Brox,
  et~al.]{Seitzer_BridgingTheGapToRealWorldObjectCentricLearning_2023}
Seitzer, M., Horn, M., Zadaianchuk, A., Zietlow, D., Xiao, T., Simon-Gabriel,
  C.-J., He, T., Zhang, Z., Sch{\"o}lkopf, B., Brox, T., et~al.
\newblock Bridging the gap to real-world object-centric learning.
\newblock In \emph{International Conference on Learning Representations
  (ICLR)}, 2023.

\bibitem[Singh et~al.(2021)Singh, Deng, and Ahn]{Singh_SLATE_2021}
Singh, G., Deng, F., and Ahn, S.
\newblock Illiterate dall-e learns to compose.
\newblock \emph{arXiv preprint arXiv:2110.11405}, 2021.

\bibitem[Singh et~al.(2022)Singh, Wu, and Ahn]{Singh_STEVE_2022}
Singh, G., Wu, Y.-F., and Ahn, S.
\newblock Simple unsupervised object-centric learning for complex and
  naturalistic videos.
\newblock In \emph{Advances in Neural Information Processing Systems
  (NeurIPS)}, 2022.

\bibitem[Struckmeier \& Kyrki(2023)Struckmeier and
  Kyrki]{Struckmeier_ILPI_2023}
Struckmeier, O. and Kyrki, V.
\newblock {ILPO-MP: Mode Priors Prevent Mode Collapse when Imitating Latent
  Policies from Observations}.
\newblock \emph{Transactions on Machine Learning Research (TMLR)}, 2023.

\bibitem[Tian et~al.(2025)Tian, Yang, Zeng, Wang, Lin, Dong, and
  Pang]{Tian_PredictiveInverseDynamocs_2024}
Tian, Y., Yang, S., Zeng, J., Wang, P., Lin, D., Dong, H., and Pang, J.
\newblock Predictive inverse dynamics models are scalable learners for robotic
  manipulation.
\newblock In \emph{International Conference on Learning Representations
  (ICLR)}, 2025.

\bibitem[Todorov et~al.(2012)Todorov, Erez, and Tassa]{todorov2012mujoco}
Todorov, E., Erez, T., and Tassa, Y.
\newblock Mujoco: A physics engine for model-based control.
\newblock In \emph{IEEE/RSJ International Conference on Intelligent Robots and
  Systems (IROS)}, 2012.

\bibitem[Van Den~Oord \& Vinyals(2017)Van Den~Oord and
  Vinyals]{VanOord_NeuralDiscreteRepresentationLearning_2017}
Van Den~Oord, A. and Vinyals, O.
\newblock Neural discrete representation learning.
\newblock In \emph{Advances in Neural Information Processing Systems
  (NeurIPS)}, 2017.

\bibitem[Vaswani et~al.(2017)Vaswani, Shazeer, Parmar, Uszkoreit, Jones, Gomez,
  Kaiser, and Polosukhin]{Vaswani_AttentionIsAllYouNeed_2017}
Vaswani, A., Shazeer, N., Parmar, N., Uszkoreit, J., Jones, L., Gomez, A.~N.,
  Kaiser, {\L}., and Polosukhin, I.
\newblock Attention is all you need.
\newblock In \emph{Advances in {N}eural {I}nformation {P}rocessing {S}ystems
  (NeurIPS)}, 2017.

\bibitem[Villar-Corrales et~al.(2023)Villar-Corrales, Wahdan, and
  Behnke]{Villar_OCVP_2023}
Villar-Corrales, A., Wahdan, I., and Behnke, S.
\newblock Object-centric video prediction via decoupling of object dynamics and
  interactions.
\newblock In \emph{IEEE International Conference on Image Processing (ICIP)},
  2023.

\bibitem[Villar-Corrales et~al.(2025)Villar-Corrales, Plepi, and
  Behnke]{TextOCVP}
Villar-Corrales, A., Plepi, G., and Behnke, S.
\newblock Object-centric image to video generation with language guidance.
\newblock \emph{arXiv preprint arXiv:2502.11655}, 2025.

\bibitem[Wang et~al.(2024)Wang, Li, Hu, Zhu, Hou, Lan, and
  Chen]{Wang_TIVDiffusion_2025}
Wang, X., Li, X., Hu, Y., Zhu, H., Hou, C., Lan, C., and Chen, Z.
\newblock {TIV-Diffusion}: Towards object-centric movement for text-driven
  image to video generation.
\newblock In \emph{AAAI Conference on Artificial Intelligence}, 2024.

\bibitem[Wang et~al.(2004)Wang, Bovik, Sheikh, and Simoncelli]{Wang_SSIM_2004}
Wang, Z., Bovik, A.~C., Sheikh, H.~R., and Simoncelli, E.~P.
\newblock Image quality assessment: from error visibility to structural
  similarity.
\newblock \emph{IEEE Transactions on Image Processing (TIP)}, 13\penalty0
  (4):\penalty0 600--612, 2004.

\bibitem[Watters et~al.(2019)Watters, Matthey, Burgess, and
  Lerchner]{Watters_SpatialBroadcastDecoder_2019}
Watters, N., Matthey, L., Burgess, C.~P., and Lerchner, A.
\newblock Spatial broadcast decoder: A simple architecture for disentangled
  representations in {VAE}s, 2019.
\newblock URL \url{https://openreview.net/forum?id=S1x7WjnzdV}.

\bibitem[Wu et~al.(2023{\natexlab{a}})Wu, Dvornik, Greff, Kipf, and
  Garg]{Wu_SlotFormer_2022}
Wu, Z., Dvornik, N., Greff, K., Kipf, T., and Garg, A.
\newblock Slot{F}ormer: Unsupervised visual dynamics simulation with
  object-centric models.
\newblock In \emph{International Conference on Learning Representations
  (ICLR)}, 2023{\natexlab{a}}.

\bibitem[Wu et~al.(2023{\natexlab{b}})Wu, Hu, Lu, Gilitschenski, and
  Garg]{Wu_SlotDiffusion_2023}
Wu, Z., Hu, J., Lu, W., Gilitschenski, I., and Garg, A.
\newblock Slotdiffusion: Object-centric generative modeling with diffusion
  models.
\newblock In \emph{Advances in Neural Information Processing Systems
  (NeurIPS)}, 2023{\natexlab{b}}.

\bibitem[Ye et~al.(2025)Ye, Jang, Jeon, Joo, Yang, Peng, Mandlekar, Tan, Chao,
  Lin, et~al.]{Ye_LatentActionPretrainingFromVideos_2025}
Ye, S., Jang, J., Jeon, B., Joo, S., Yang, J., Peng, B., Mandlekar, A., Tan,
  R., Chao, Y.-W., Lin, B.~Y., et~al.
\newblock Latent action pretraining from videos.
\newblock In \emph{International Conference on Learning Representations
  (ICLR)}, 2025.

\bibitem[Ye et~al.(2022)Ye, Zhang, Abbeel, and
  Gao]{Ye_BecomeAProficientPlayerByWatchingPureVideos_2022}
Ye, W., Zhang, Y., Abbeel, P., and Gao, Y.
\newblock Become a proficient player with limited data through watching pure
  videos.
\newblock In \emph{International Conference on Learning Representations
  (ICLR)}, 2022.

\bibitem[Yoon et~al.(2023)Yoon, Wu, Bae, and
  Ahn]{Yoon_InvestigationObjectCentricRepresentationsReinforcementLearning_2023}
Yoon, J., Wu, Y.-F., Bae, H., and Ahn, S.
\newblock An investigation into pre-training object-centric representations for
  reinforcement learning.
\newblock In \emph{International Conference on Machine Learning (ICML)}, 2023.

\bibitem[Yu et~al.(2020)Yu, Quillen, He, Julian, Hausman, Finn, and
  Levine]{Yu_MetaworldDataset_2020}
Yu, T., Quillen, D., He, Z., Julian, R., Hausman, K., Finn, C., and Levine, S.
\newblock Meta-world: A benchmark and evaluation for multi-task and meta
  reinforcement learning.
\newblock In \emph{Conference on Robot Learning (CoRL)}, 2020.

\bibitem[Zadaianchuk et~al.(2024)Zadaianchuk, Seitzer, and
  Martius]{Zadaianchuk_VideoSaur_2024}
Zadaianchuk, A., Seitzer, M., and Martius, G.
\newblock Object-centric learning for real-world videos by predicting temporal
  feature similarities.
\newblock In \emph{Advances in Neural Information Processing Systems
  (NeurIPS)}, 2024.

\bibitem[Zhang et~al.(2022)Zhang, Gupta, and
  Zisserman]{Zhang_IsAnObject-CentricVideoRepresentationBeneficialForTransfer_2022}
Zhang, C., Gupta, A., and Zisserman, A.
\newblock Is an object-centric video representation beneficial for transfer?
\newblock In \emph{Asian Conference on Computer Vision (ACCV)}, 2022.

\bibitem[Zhang et~al.(2018)Zhang, Isola, Efros, Shechtman, and
  Wang]{Zhang_TheUnreasonableEffectivenessOfDeepFeaturesLPIPS_2018}
Zhang, R., Isola, P., Efros, A.~A., Shechtman, E., and Wang, O.
\newblock The unreasonable effectiveness of deep features as a perceptual
  metric.
\newblock In \emph{IEEE Conference on Computer Vision and Pattern Recognition
  (CVPR)}, 2018.

\bibitem[Zoran et~al.(2021)Zoran, Kabra, Lerchner, and
  Rezende]{Zoran_PARTS_2021}
Zoran, D., Kabra, R., Lerchner, A., and Rezende, D.~J.
\newblock Parts: Unsupervised segmentation with slots, attention and
  independence maximization.
\newblock In \emph{IEEE/CVF International Conference on Computer Vision
  (ICCV)}, 2021.

\end{thebibliography}
\bibliographystyle{icml2025}

\newpage
\appendix

\onecolumn  

\section{Limitations \& Future Work}

We recognize two main limitations that currently limit the scope of our \Method~framework to simple tabletop robotic scenarios, preventing it from generalizing to more complex domains.

\paragraph{Limited Decomposition Model} The first limitation arises from the SAVi object-centric decomposition model used at the core of our framework.
SAVi achieves a great decomposition performance on datasets with objects of simple shapes and textures. However, it fails to generalize to complex real-world robotic scenarios, hence limiting the current scope of \Method{} to robotic tabletop simulations, or simple real-world environments as in Sketchy.

\paragraph{Single Latent Action} The second limitation lies at the representational capability of our latent actions and action prototypes.
In robotic scenarios, several actions often happen simultaneously, such as moving and rotating the robot, as well as opening or closing the gripper.
Our current latent action representation jointly models the scene's inverse dynamics, encoding together all actions into a single latent space.
This entangled representation limits our ability to control the agents in the scene with greater precision.

\paragraph{Future Work} In future work, we plan to extend our proposed \Method{} framework with more capable decomposition models, such as DINOSAUR~\cite{Seitzer_BridgingTheGapToRealWorldObjectCentricLearning_2023} or SOLV~\cite{Aydemir_SelfSupervisedObjectCentricLearningForVideos_2023}, as well as scale our inverse dynamics and predictor models.
Furthermore, we can employ factorized latent action vectors, which represent in a disentangled manner different actions that happen simultaneously, such as moving the robot arm and opening the gripper.
We believe that this architectural modifications will enable us to use \Method{} on more complex robotic simulations and perform real-world robotic experiments.

\section{Implementation Details}
\label{app: implementation details}

In this section, we describe the network architecture and training details for each of the components in our \Method{} framework.
Our models are implemented in PyTorch~\cite{Paszke_AutomaticDifferneciationInPytorch_2017} and are trained on a single NVIDIA A100 GPU.

\subsection{Object-Centric Learning}

We closely follow~\citet{Kipf_ConditionalObjectCentricLearningFromVideo_2022} for the implementation of the SAVi object-centric decomposition model, which we employ as scene parsing and object rendering modules.
We strictly adhere to the architecture of their proposed CNN-based image encoder $\ImageEncoder$, slot decoder $\SlotDecoder$, transformer-based dynamics transition module, and Slot Attention corrector.
We use a variable number of 128-dimensional object slots, depending of the dataset.
Namely, we employ eight object slots on \RobotDB, four slots on \ButtonPress, three slots on GridShapes, and eight slots on Sketchy.
On all datasets, we sample the initial object slots $\SlotsT{0}$ from a Gaussian distribution with learned mean and covariance.
Furthermore, we use three Slot Attention iterations for the initial video
frame to obtain a good initial object-centric decomposition, and a single iteration for subsequent frames, which suffices to recursively update the slot state given the newly observed image features.

\subsection{Inverse Dynamics Module}

We propose two variants of our \InverseDynamics~module.

\noindent \textbf{$\InverseDynamicsSingle$:} $\InverseDynamicsSingle$~jointly processes the object slots from a single time step $\SlotsT{t}$ along with an additional
token \ActToken~using a transformer encoder.
We use a four-layer transformer encoder with a 256-dimensional tokens, four 64-dimensional heads and hidden dimension of 1024.
This module aggregates information from the object slots into the \ActToken~token, and outputs a single latent action $\PredLatentT{t}$  that captures the agent’s action, making it well-suited for single-agent environments.

\noindent \textbf{$\InverseDynamicsMany$:} $\InverseDynamicsMany$~independently processes each object slot with a shared MLP, thus generating $\NumSlots$ latent action embeddings, each representing the action of a specific object in the scene.
This design makes $\InverseDynamicsMany$~well-suited for environments with multiple moving agents.
We employ a two-layer MLP, featuring a ReLU nonlinear activation and layer normalization.

As described in~\Section{sec: inverse dynamics}, the generated latent actions $\PredLatentT{t}$ are vector-quantized to assign them to their corresponding action prototype $\ActionProtoT{t}$.
On the \ButtonPress, \RobotDB, and Sketchy datasets, we use the $\InverseDynamicsSingle$ variant with eight different 16-dimensional action prototypes,
whereas for GridShapes we use $\InverseDynamicsMany$ with five distinct eight-dimensional action prototypes.
Following common practice, we update the action prototypes using the exponential moving average updates of cluster assignment counts~\cite{VanOord_NeuralDiscreteRepresentationLearning_2017}.

\subsection{Conditional Object-Centric Predictor}

Our conditional object-centric predictor (\CondPred) is inspired by the transformer-based SlotFormer~\cite{Wu_SlotFormer_2022} architecture.
Our \CondPred~module features four layers, 256-dimensional tokens, eight 64-dimensional attention heads, and hidden dimension of 1024.

To enable predictions conditioned on the inferred latent actions, \CondPred~maps the action prototypes $\ActionProtoT{1:t}$, variability embeddings $\ActionVariabilityT{1:t}$ and object slots $\SlotsT{1:t}$ into the token dimensionality.
The projected object slots are then conditioned by adding them with the projected action prototype and variability embedding from the corresponding time step.
Furthermore, following~\citet{Wu_SlotFormer_2022}, we augment these representations with a temporal sinusoidal positional encoding, such that all tokens from the same time step receive the same
encoding, thus preserving the inherent permutation equivariance of the objects.

\subsection{Policy Model and Action Decoder}

The policy model $\PolicyModel$ follows a similar architecture to $\InverseDynamicsSingle$.
$\PolicyModel$ is a four layer transformer that jointly processes the objects slots from a single time step $\SlotsT{t}$ and an additional token \ActToken in order to regress a latent action.

The action decoder is a three-layer MLP with a hidden dimension of 128 that maps latent action vectors into real-world actions.

\subsection{Training Details}

\noindent \textbf{SAVi Training:}  SAVi is trained for object-centric decomposition using the Adam optimizer~\cite{Kingma_Adam_2014}, a batch size of 64, sequences of length eight frames, and a base learning rate of $10^{-4}$, which is linearly warmed-up for the first 4000 steps, followed by cosine annealing for the remaining of the training process.
Moreover, we clip the gradients to a maximum norm of 0.05.

\noindent \textbf{$\InverseDynamics$~and \CondPred~Training:} We jointly train our $\InverseDynamics$~and \CondPred{} modules given a pretrained SAVi decomposition model.
These modules are trained with the Adam optimizer~\cite{Kingma_Adam_2014}, batch size of 64, and a base learning rate of $2 \times 10^{-4}$, which decreases during training with a cosine annealing schedule.
To stabilize the training, we clip the gradients to a maximum norm of 0.05.
We set the loss weights to $\lambda_{\text{Img}} = 1$, $\lambda_{\text{Slot}} = 1$, and $\lambda_{\text{VQ}} = 0.25$.

\noindent \textbf{$\PolicyModel$~and $\ActionDecoder$~Training:} We train the $\PolicyModel${} and $\ActionDecoder${} modules given pretrained and frozen SAVi, $\InverseDynamics${} and \CondPred{} modules.
These modules are trained with the Adam optimizer~\cite{Kingma_Adam_2014}, batch size of 64, and a fixed learning rate of $2 \times 10^{-4}$.

\section{Dataset Details}
\label{app: dataset}

\subsection{\RobotDB} This environment, inspired by~\citet{Li_TowardsPracticalMultiObjectManipulationRelationalRL_2020} and simulated using MuJoCo~\cite{todorov2012mujoco}, features a robot arm on a tabletop interacting with multiple uni-colored blocks.
We use two different dataset variants.

The first variant consists of a robot controlled by a random exploration policy, moving in the environment with minimal meaningful object interactions.
This easily simulated dataset includes 20,000 training sequences and 2,000 validation sequences.
We use this dataset variant for training SAVi, as well as for jointly training our inverse dynamics and conditional predictor modules.

The second variant contains a smaller subset of expert demonstrations where the robot is tasked to push the block of distinct color to a target location marked with a red sphere.
This task evaluates the capabilities of an agent to reason about object relations and model object collisions.
We collect 5,000 expert demonstrations using a pretrained policy~\cite{Mosbach_SOLDReinforcementLearningSlotObjectCentricLatentDynamics} with a success rate of $\approx 80\%$ for this pushing task.
We split this dataset into 4,500 training sequences, which are used for training the policy model and action decoder; and 500 evaluation sequences that are used for benchmarking the prediction models.

\subsection{\ButtonPress}

This environment, based on MetaWorld~\cite{Yu_MetaworldDataset_2020} features a Sawyer robot arm tasked with pressing a red button.
Unlike \RobotDB, it involves a non-object-centric task with complex shapes and textures.
As before, we use two different dataset variants.

The first variant contains 10,000 sequences, split into 9,000 training and 1,000 validation videos, with the robot controlled by a random exploration policy.
We use this dataset variant to train SAVi, as well as our inverse dynamics and conditional predictor modules.

The second variant includes a small subset of expert demonstrations where the robot successfully presses the red button.
We collect 1,000 expert demonstrations using an expert policy, from which 900 are used for training the policy model and action decoder, and 100 demonstrations are used for benchmarking the prediction models.

\subsection{GridShapes}

This dataset features one or more simple 2D shapes moving in a grid-like pattern on top of a colored background.
The shapes can be either a ball, triangle or square, and have a random color. These shapes can move up, down, left, right or remain still, and revert their motion when an image boundary is reached, thus emulating a bouncing effect.
To introduce some stochasticity into the motion, the shapes randomly change direction with a predefined probability of 0.25.
We train and evaluate the models on several variants of the GridShapes dataset featuring different number of objects, ranging from one single moving shape, to five objects moving independently in the same sequence.
This simple dataset serves as benchmark to evaluate a model’s ability to jointly predict the motion of multiple moving agents in the scene.

\subsection{Sketchy}

Sketchy~\cite{Cabi_ScalingDataDrivenRoboticsRewardSketching_2020} is a real-world robotics dataset featuring videos in which a robotic gripper interacts with diverse objects of different shape and color on a tabletop scenario.
We use the human demonstration sequences provided in the dataset, resulting in a total of 394 sequences with an average length of 80 frames. 
For each sequence, we use the observations taken with the \texttt{front right} and \texttt{front left} cameras, and discard the cropped images.
We resize the video frames to an image size of $64 \times 64$ and use 316 and 78 sequences for training and evaluation, respectively.

\section{Baselines}

In our experiments, we compare our approach with different baseline models, including the stochastic and playable video prediction models SVG~\cite{Denton_StochasticVideoGenerationWithALearnedPrior_2018} and CADDY~\cite{Menapace_PlayableVideoGeneration_2021}, as well as the object-centric video prediction models SlotFormer~\cite{Wu_SlotFormer_2022} and OCVP~\cite{Villar_OCVP_2023}.
To ensure a fair comparison, we balance the number of learnable parameters and compute requirements for all methods. \Table{table:params} lists the  number of learnable parameters in our proposed \Method{} as well as for all baselines.

\begin{table}[h]
	\centering
	\caption{Number of learnable parameters for \Method{} and baselines.}
	\vspace{-0.3cm}
	\footnotesize
	\begin{tabular}{l l}
		\toprule
		\textbf{Model} & \textbf{Trainable Parameters (M)} \\
		\midrule
		SVG & 18.84 \\
		CADDY & 9.34 \\
		SlotFormer & 3.97 (2.96 Predictor + 1.01 SAVi) \\
		OCVP & 4.86 (3.85 Predictor + 1.01 SAVi) \\
		\Method & 8.49 (3.19 InvDyn + 4.27 cOCVP + 1.01 SAVi) \\
		\bottomrule
	\end{tabular}
	\label{table:params}
\end{table}

\subsection{SVG}

SVG~\cite{Denton_StochasticVideoGenerationWithALearnedPrior_2018} is a generative model for video prediction that captures both deterministic dynamics and stochastic variations in video sequences.
It combines a variational autoencoder (VAE) with recurrent neural networks (RNNs) to model stochastic temporal dynamics.
SVG represents a probabilistic framework, where the next frame $\PredImageT{t+1}$ is generated based on the previous frame $\ImageT{t}$ and a latent sample $\PredLatentT{t}$ drawn from a latent distribution.
For a fair comparison with \Method{}, we use SVG's posterior module to infer latent vectors using future video frames, and use those latent vectors to guide the video prediction process.
In our experiments, we adapt the original implementation\footnote{\url{https://github.com/edenton/svg}} and use an SVG variant with a learned prior, VGG-like encoder and decoders, and two recurrent predictor layers.

\textbf{Main Difference with \Method:}
SVG operates with holistic scene representations, whereas our proposed method employs a structured object-centric representation.
Furthermore, SVG encodes the stochastic scene dynamics into a single continuous latent vector. In contrast, \Method{} follows a hybrid approach in which the latent action vectors are composed of an action prototype and action variability embeddings, making the prediction process more controllable and interpretable.

\subsection{CADDY}
CADDY~\cite{Menapace_PlayableEnvironments_2022} is a recurrent encoder-decoder  model designed for playable video generation, enabling user-controllable future video prediction.
CADDY infers latent actions that encode the agent's actions between consecutive pairs of frames.
These latent actions are parameterized with a discrete \emph{one-hot action label}, which determines the high-level action taking place; and a high-dimensional \emph{action variability embedding}, which describes the variability of each action and captures the possible non-determinism in the environment.
We adapt the original implementation\footnote{\url{https://github.com/willi-menapace/PlayableVideoGeneration}} for our experiments.

\textbf{Main Difference with \Method:}
CADDY operates with holistic scene representations, i.e. CNN features, whereas our proposed method employs a structured object-centric representation.
Moreover, despite both methods using a hybrid parameterization of the latent actions, they differ in their implementation.
CADDY learns a discrete one-hot action label by minimizing multiple regularization objectives, including an action matching loss and several Kullback–Leibler  (KL) divergences losses.
In contrast, \Method{} employs a discrete set of high-dimensional action prototypes, which are learned via vector-quantization of the latent space.
We empirically verify that both approaches achieve comparable performance.
However, our vector quantization approach requires significantly fewer hyper-parameters and is easier to tune.

\subsection{SlotFormer}
SlotFormer~\cite{Wu_SlotFormer_2022} is an object-centric video prediction model that builds upon slot-based representations.
First, it uses Slot Attention to parse videos frames into object-centric latent representations, called slots.
SlotFormer then employs a transformer-based autoregressive predictor module, which jointly processes all input slots in order to forecast future object representations.
Finally, the predicted slots are decoded into object images and video frames.
In our experiments, we adapt the original implementation\footnote{\url{https://github.com/pairlab/SlotFormer}}.

\textbf{Main Difference with \Method:} SlotFormer forecasts future slots in an unconditional manner, thus not being able to model stochastic environments or agents such as robots.
\Method{} addresses this challenge by inferring the scene inverse dynamics and using them to condition the prediction process.

\subsection{OCVP}
OCVP~\cite{Villar_OCVP_2023}, similar to SlotFormer, is a slot-based object-centric video prediction model.
Differing from SlotFormer, OCVP leverages two specialized decoupled attention mechanisms, \emph{relational} and \emph{temporal} attention, which model the object interactions and dynamics, respectively.
In our experiments, we use the OCVP-Seq variant with default settings adapted from original implementation\footnote{\url{https://github.com/AIS-Bonn/OCVP-object-centric-video-prediction}}.

\textbf{Main Difference with \Method:} OCVP forecasts future slots in an unconditional manner, thus not being able to model stochastic environments or agents such as robots.
\Method{} addresses this challenge by inferring the scene inverse dynamics and using them to condition the prediction process.

\section{Additional Results}
\label{app: extra results}

\subsection{Effect of the Number of Actions}

In \Table{table: num_actions} we evaluate on the \RobotDB~dataset multiple \Method{} variants using a different number of learned action prototypes.
We show that using eight learned action prototypes achieves the best video prediction performance, while learning a concise semantically meaningful set of action prototypes.

\begin{table}[h!]
	\centering
	\caption{
		 Evaluation of \Method~variants on the \RobotDB~dataset using a different number
		 of learned action prototypes.
	}
	\label{table: num_actions}
	\vspace{-0.3cm}
	\footnotesize
	\renewcommand{\arraystretch}{1.}
	\begin{tabular}{P{2.2cm} P{1.05cm}P{1.05cm}P{1.05cm}}
		\toprule
		\multicolumn{1}{c}{} &  \multicolumn{3}{c}{\textbf{\RobotDB}} \\
		\cmidrule(r){2-4} 
		\textbf{\# Actions} & {PSNR}$\uparrow$ & {SSIM}$\uparrow$ & {LPIPS}$\downarrow$ \\
		\midrule
		5  & 21.26 & 0.886  & 0.071 \\
		8  & \textbf{21.41} & \textbf{0.890} & \textbf{0.066} \\
		10 & 21.36 & \underline{0.889} & \underline{0.067} \\
		15 & 21.38 & \underline{0.889} & \underline{0.067} \\
		\bottomrule
	\end{tabular}
\end{table}

\subsection{Learned Behaviors from Expert Demonstrations}
\label{app: learned behaviors}

In this section, we present additional experiments and results demonstrating \Method{}'s ability to learn robot behaviors from unlabeled expert demonstrations.
In \Section{app: rollout vs simulation}, we compare policies learned with \Method{} under two different evaluation protocols.
In \Section{app: comparison with lapo}, we compare \Method{} with LAPO~\cite{Schmidt_LAPOLearningToActWithoutActions_2024}, a recent model for behavior learning from unlabeled demonstrations.
Finally, \Section{app: vis learned behaviors} provides additional rollout visualizations from \Method's policy, along with simulated executions of the predicted action sequences.

\subsubsection{Imagined Rollout vs. Simulation}
\label{app: rollout vs simulation}

\begin{figure*}[b!]
	\begin{tikzpicture} 
		%
		%
		\node(plot_0)[anchor=north, inner sep=0, outer sep=0] at (0,0.0) 
		{\includegraphics[width=0.48\linewidth]
			{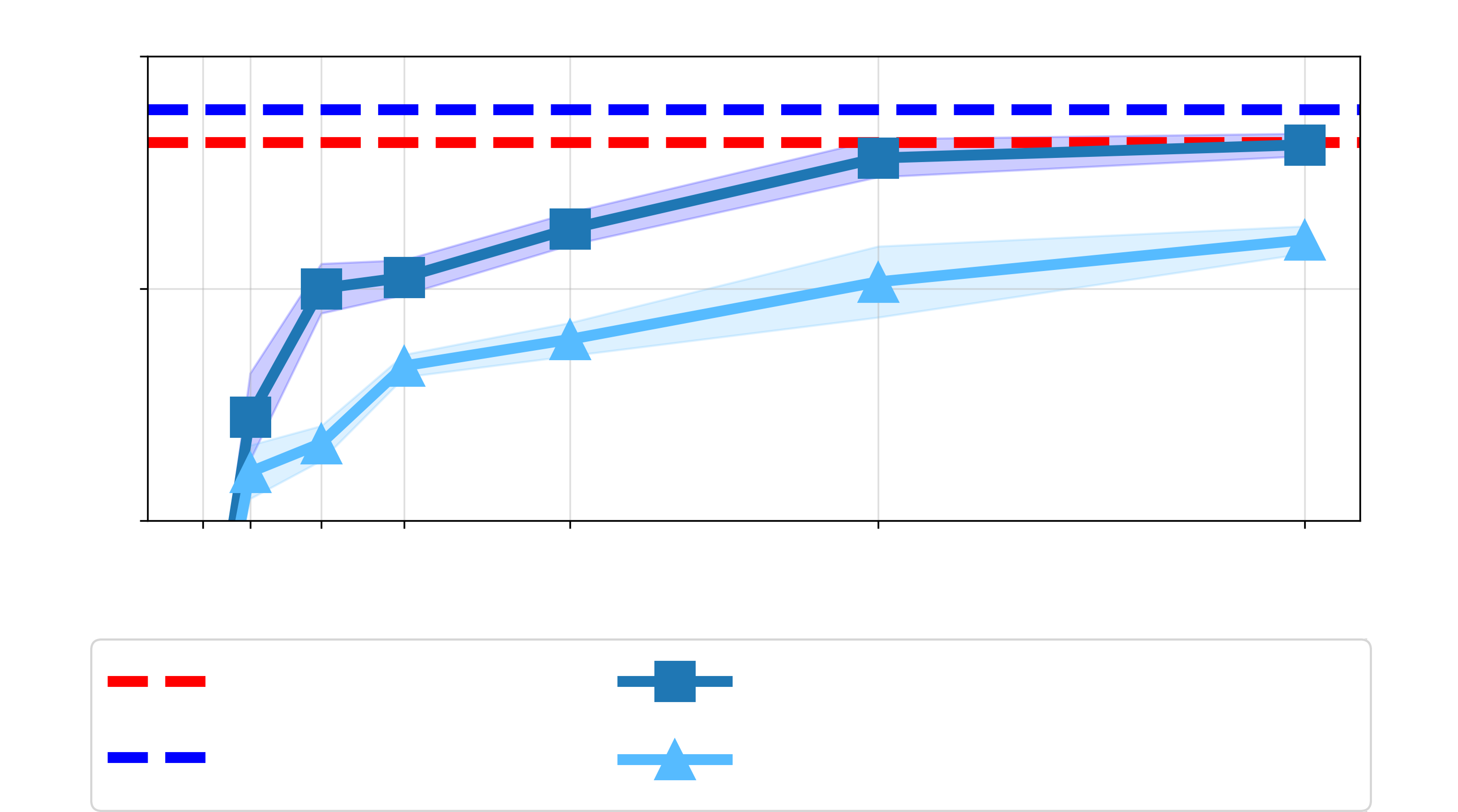}
		};
		\node(a)[anchor=north west,draw=none,ultra thick,inner sep=0, outer sep=0] at([yshift=0.8cm, xshift=0.0cm]plot_0.south west)
		{\large a) };
		\node(a)[anchor=north,draw=none,ultra thick,inner sep=0, outer sep=0] at([yshift=1.28cm, xshift=0.35cm]plot_0.south)
		{\# Expert Demonstrations };
		\node(a)[anchor=north east,draw=none,ultra thick,inner sep=0, outer sep=0] at([yshift=4.2cm, xshift=0.25cm]plot_0.south west)
		{\rotatebox{90}{Success Rate [\%]}};
		\node(title)[anchor=south,draw=none,ultra thick,inner sep=0, outer sep=0] at([yshift=-0.2cm, xshift=0.2cm]plot_0.north)
		{\textbf{\ButtonPress{} Behavior}};
		\node(legend1)[anchor=north,draw=none,ultra thick,inner sep=0, outer sep=0] at([yshift=0.84cm, xshift=-1.8cm]plot_0.south){\customsize{ResNet Oracle}};
		\node(legend2)[anchor=north west,draw=none,ultra thick,inner sep=0, outer sep=0] at([yshift=-0.2cm, xshift=0.0cm]legend1.south west){\customsize{Slot Oracle}};
		\node(legend3)[anchor=west,draw=none,ultra thick,inner sep=0, outer sep=0] at([yshift=-0.01cm, xshift=0.92cm]legend1.east){\customsize{\Method{} w/ Simulator}};
		\node(legend4)[anchor=north west,draw=none,ultra thick,inner sep=0, outer sep=0] at([yshift=-0.15cm, xshift=0.0cm]legend3.south west){\customsize{\Method{} w/ Latent Imag.}};
		\node(lab0)[anchor=east,draw=none,ultra thick,inner sep=0, outer sep=0] at([yshift=-0.35cm, xshift=0.78cm]plot_0.north west){\labelsize{100}};
		\node(lab0)[anchor=east,draw=none,ultra thick,inner sep=0, outer sep=0] at([yshift=0.65cm, xshift=0.78cm]plot_0.west){\labelsize{65}};
		\node(lab0)[anchor=east,draw=none,ultra thick,inner sep=0, outer sep=0] at([yshift=1.68cm, xshift=0.78cm]plot_0.south west){\labelsize{30}};
		\node(lab0)[anchor=north,draw=none,ultra thick,inner sep=0, outer sep=0]at([yshift=1.55cm, xshift=-2.98cm]plot_0.south){\labelsize{0}};
		\node(lab0)[anchor=north,draw=none,ultra thick,inner sep=0, outer sep=0]at([yshift=1.55cm, xshift=-2.68cm]plot_0.south){\labelsize{10}};
		\node(lab0)[anchor=north,draw=none,ultra thick,inner sep=0, outer sep=0]at([yshift=1.55cm, xshift=-2.25cm]plot_0.south){\labelsize{50}};
		\node(lab0)[anchor=north,draw=none,ultra thick,inner sep=0, outer sep=0]at([yshift=1.55cm, xshift=-1.8cm]plot_0.south){\labelsize{100}};
		\node(lab0)[anchor=north,draw=none,ultra thick,inner sep=0, outer sep=0]at([yshift=1.55cm, xshift=-0.9cm]plot_0.south){\labelsize{200}};
		\node(lab0)[anchor=north,draw=none,ultra thick,inner sep=0, outer sep=0]at([yshift=1.55cm, xshift=0.86cm]plot_0.south){\labelsize{500}};
		\node(lab0)[anchor=north,draw=none,ultra thick,inner sep=0, outer sep=0]at([yshift=1.55cm, xshift=3.25cm]plot_0.south){\labelsize{900}};
		%
		%
		%
		%
		%
		%
		\node(plot_1)[anchor=west, inner sep=0, outer sep=0] at ([yshift=0.0cm, xshift=0.5cm]plot_0.east) 
		{\includegraphics[width=0.48\linewidth]
			{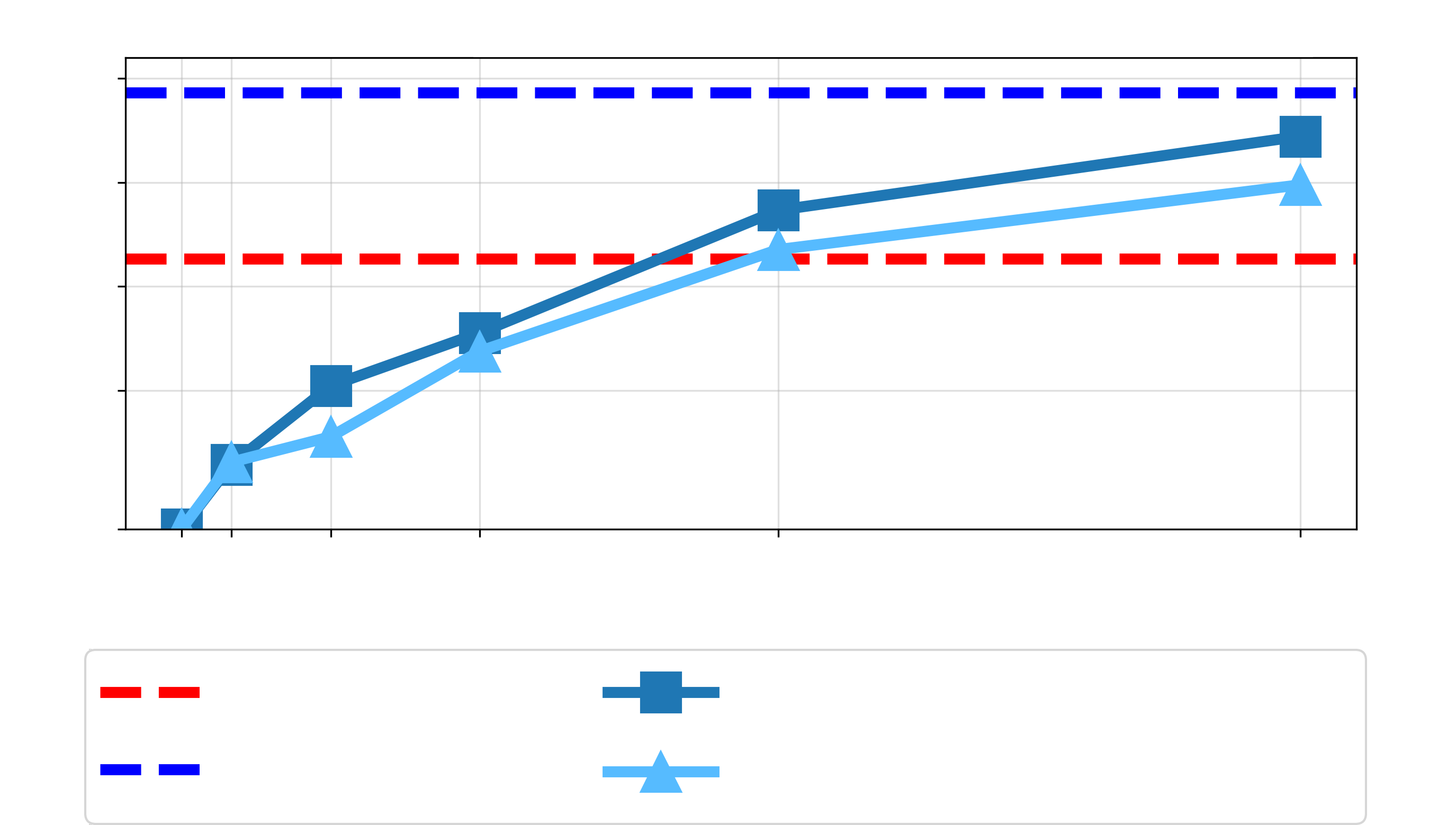}
		};
		\node(b)[anchor=north west,draw=none,ultra thick,inner sep=0, outer sep=0] at([yshift=0.8cm, xshift=0.0cm]plot_1.south west)
		{\large b) };
		\node(a)[anchor=north,draw=none,ultra thick,inner sep=0, outer sep=0] at([yshift=1.28cm, xshift=0.35cm]plot_1.south)
		{\# Expert Demonstrations };
		\node(a)[anchor=north east,draw=none,ultra thick,inner sep=0, outer sep=0] at([yshift=4.2cm, xshift=0.25cm]plot_1.south west)
		{\rotatebox{90}{Success Rate [\%]}};
		\node(title)[anchor=south,draw=none,ultra thick,inner sep=0, outer sep=0] at([yshift=-0.2cm, xshift=0.2cm]plot_1.north)
		{\textbf{\RobotDB{} Behavior}};
		\node(legend1)[anchor=north,draw=none,ultra thick,inner sep=0, outer sep=0] at([yshift=0.84cm, xshift=-1.9cm]plot_1.south){\customsize{ResNet Oracle}};
		\node(legend2)[anchor=north west,draw=none,ultra thick,inner sep=0, outer sep=0] at([yshift=-0.2cm, xshift=0.0cm]legend1.south west){\customsize{Slot Oracle}};
		\node(legend3)[anchor=west,draw=none,ultra thick,inner sep=0, outer sep=0] at([yshift=-0.01cm, xshift=0.92cm]legend1.east){\customsize{\Method{} w/ Simulator}};
		\node(legend4)[anchor=north west,draw=none,ultra thick,inner sep=0, outer sep=0] at([yshift=-0.15cm, xshift=0.0cm]legend3.south west){\customsize{\Method{} w/ Latent Imag.}};
		\node(lab0)[anchor=east,draw=none,ultra thick,inner sep=0, outer sep=0] at([yshift=-0.45cm, xshift=0.65cm]plot_1.north west){\labelsize{65}};
		\node(lab0)[anchor=east,draw=none,ultra thick,inner sep=0, outer sep=0] at([yshift=1.3cm, xshift=0.65cm]plot_1.west){\labelsize{50}};
		\node(lab0)[anchor=east,draw=none,ultra thick,inner sep=0, outer sep=0] at([yshift=0.70cm, xshift=0.65cm]plot_1.west){\labelsize{35}};
		\node(lab0)[anchor=east,draw=none,ultra thick,inner sep=0, outer sep=0] at([yshift=0.12cm, xshift=0.65cm]plot_1.west){\labelsize{20}};
		\node(lab0)[anchor=east,draw=none,ultra thick,inner sep=0, outer sep=0] at([yshift=1.7cm, xshift=0.65cm]plot_1.south west){\labelsize{0}};
		\node(lab0)[anchor=north,draw=none,ultra thick,inner sep=0, outer sep=0]at([yshift=1.55cm, xshift=-3.13cm]plot_1.south){\labelsize{0}};
		\node(lab0)[anchor=north,draw=none,ultra thick,inner sep=0, outer sep=0]at([yshift=1.55cm, xshift=-2.75cm]plot_1.south){\labelsize{200}};
		\node(lab0)[anchor=north,draw=none,ultra thick,inner sep=0, outer sep=0]at([yshift=1.55cm, xshift=-2.2cm]plot_1.south){\labelsize{500}};
		\node(lab0)[anchor=north,draw=none,ultra thick,inner sep=0, outer sep=0]at([yshift=1.55cm, xshift=-1.4cm]plot_1.south){\labelsize{1000}};
		\node(lab0)[anchor=north,draw=none,ultra thick,inner sep=0, outer sep=0]at([yshift=1.55cm, xshift=0.295cm]plot_1.south){\labelsize{2000}};
		\node(lab0)[anchor=north,draw=none,ultra thick,inner sep=0, outer sep=0]at([yshift=1.55cm, xshift=3.24cm]plot_1.south){\labelsize{4500}};
	\end{tikzpicture}
	\vspace{-0.25cm}
	\caption{
		 Success rates (\%) of policies learned with \Method{} under two evaluation protocols on the a) \ButtonPress{} and b) \RobotDB{} environments.
		 Simulation-based rollouts outperform latent imagination, but both approaches improve with more available expert demonstrations and surpass the ResNet Oracle.
	}
	\label{fig: comparison protocols}
	\vspace{-0.4cm}
\end{figure*}

In \Figure{fig: comparison protocols}, we compare two different protocols for evaluating the behaviors learned with \Method{} on the \ButtonPress{} and \RobotDB{} environments. Namely:

\textbf{w/ Latent Imagination~~} The learned behavior unfolds within the model’s latent imagination, where future states are predicted with \Method{}.
To assess the performance of the policy, the predicted sequence of latent actions is decoded and executed in the simulator, with the final outcome---success or failure---used as the performance measure. 
This protocol assesses how well the learned representations and policies generalize within the internal model's constraints.

\textbf{w/ Simulation~~}  The latent actions output by the policy are directly decoded and executed in the simulator, which generates the subsequent environment state.
This process is repeated iteratively for a fixed number of steps or until the task is successfully completed.
This protocol evaluates the quality of both the learned policy and action representation independently of the world model’s accuracy, providing a more direct measure of the policy’s effectiveness in an interactive setting.

As shown in \Figure{fig: comparison protocols}, success rates improve for both evaluation protocols as the number of available expert demonstrations increases.
Notably, with as few as 500 and 2000 expert demonstrations, \Method{} w/ Simulator matches---and even outperforms---the ResNet Oracle baseline on the \ButtonPress{} and \RobotDB{} environments, respectively.
This suggests that \Method{} can effectively learn robust and performant robot behaviors from limited expert data when evaluated in an interactive setting.

Unsurprisingly, the latent imagination-based rollouts perform slightly worse than those executed in the simulator.
This discrepancy arises because errors in the learned world model can compound over time, causing deviations from real-world dynamics.
Nonetheless, the performance gap remains moderate, with the imagination-based rollouts succeeding approximately 70\% and 50\% of the scenes on the \ButtonPress{} and \RobotDB{} environments, respectively.

\subsubsection{Comparison with LAPO}
\label{app: comparison with lapo}

\begin{figure*}[t!]
	\begin{tikzpicture} 
		%
		%
		\node(plot_0)[anchor=north, inner sep=0, outer sep=0] at (0,0.0) 
		{\includegraphics[width=0.48\linewidth]
			{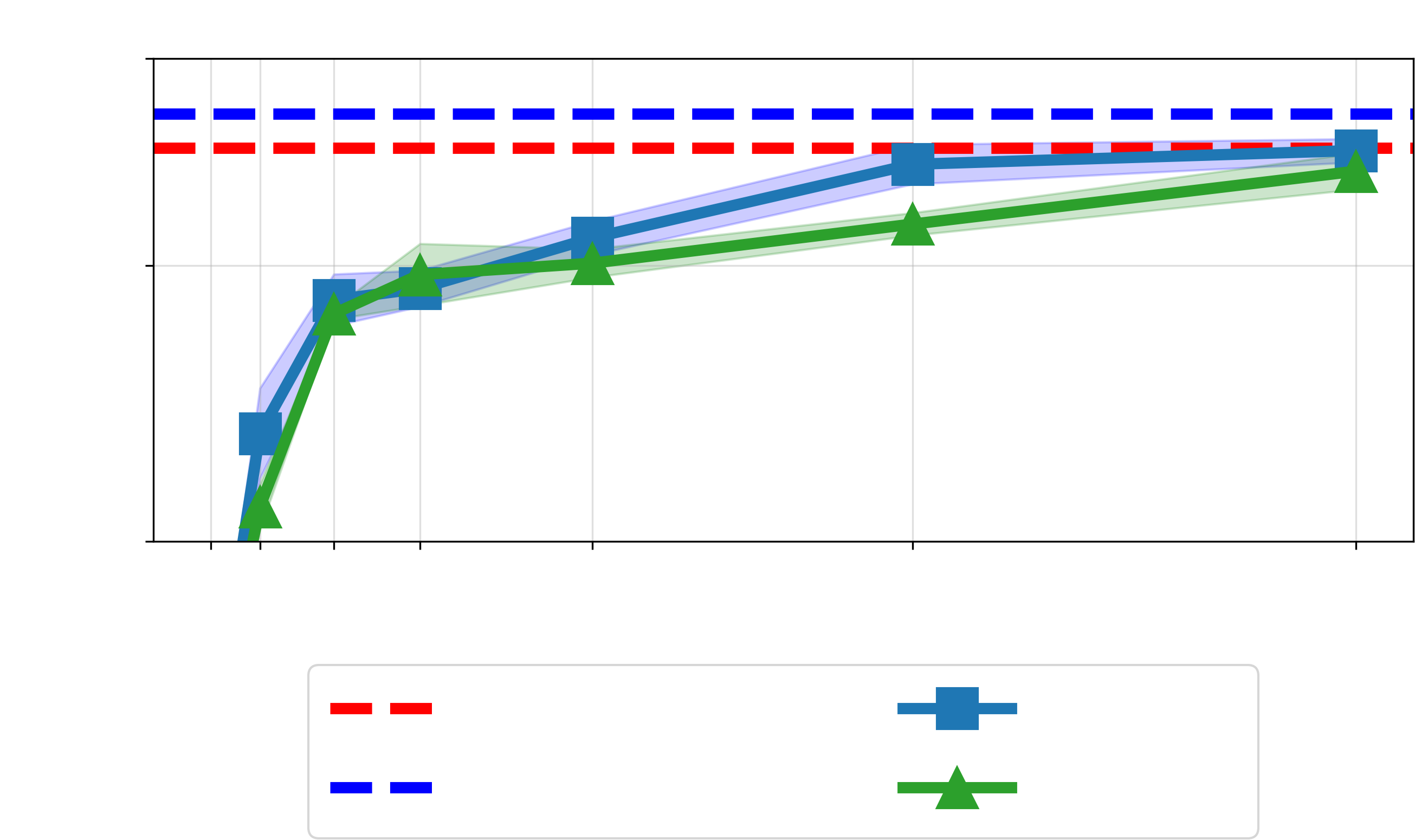}
		};
		\node(a)[anchor=north west,draw=none,ultra thick,inner sep=0, outer sep=0] at([yshift=0.8cm, xshift=0.1cm]plot_0.south west)
		{\large a) };
		\node(a)[anchor=north,draw=none,ultra thick,inner sep=0, outer sep=0] at([yshift=1.34cm, xshift=0.55cm]plot_0.south)
		{\# Expert Demonstrations };
		\node(a)[anchor=north east,draw=none,ultra thick,inner sep=0, outer sep=0] at([yshift=4.4cm, xshift=0.3cm]plot_0.south west)
		{\rotatebox{90}{Success Rate [\%]}};
		\node(title)[anchor=south,draw=none,ultra thick,inner sep=0, outer sep=0] at([yshift=-0.2cm, xshift=0.4cm]plot_0.north)
		{\textbf{\ButtonPress{} Behavior}};
		\node(legend1)[anchor=north,draw=none,ultra thick,inner sep=0, outer sep=0] at([yshift=0.86cm, xshift=-0.4cm]plot_0.south){\customsize{ResNet Oracle}};
		\node(legend2)[anchor=north west,draw=none,ultra thick,inner sep=0, outer sep=0] at([yshift=-0.2cm, xshift=0.0cm]legend1.south west){\customsize{Slot Oracle}};
		\node(legend3)[anchor=west,draw=none,ultra thick,inner sep=0, outer sep=0] at([yshift=-0.01cm, xshift=1.3cm]legend1.east){\customsize{PlaySlot}};
		\node(legend4)[anchor=north west,draw=none,ultra thick,inner sep=0, outer sep=0] at([yshift=-0.2cm, xshift=0.0cm]legend3.south west){\customsize{LAPO}};
		\node(lab0)[anchor=east,draw=none,ultra thick,inner sep=0, outer sep=0] at([yshift=-0.35cm, xshift=0.8cm]plot_0.north west){\labelsize{100}};
		\node(lab0)[anchor=east,draw=none,ultra thick,inner sep=0, outer sep=0] at([yshift=0.88cm, xshift=0.8cm]plot_0.west){\labelsize{70}};
		\node(lab0)[anchor=east,draw=none,ultra thick,inner sep=0, outer sep=0] at([yshift=1.75cm, xshift=0.8cm]plot_0.south west){\labelsize{30}};
		\node(lab0)[anchor=north,draw=none,ultra thick,inner sep=0, outer sep=0]at([yshift=1.62cm, xshift=-2.9cm]plot_0.south){\labelsize{0}};
		\node(lab0)[anchor=north,draw=none,ultra thick,inner sep=0, outer sep=0]at([yshift=1.62cm, xshift=-2.58cm]plot_0.south){\labelsize{10}};
		\node(lab0)[anchor=north,draw=none,ultra thick,inner sep=0, outer sep=0]at([yshift=1.62cm, xshift=-2.15cm]plot_0.south){\labelsize{50}};
		\node(lab0)[anchor=north,draw=none,ultra thick,inner sep=0, outer sep=0]at([yshift=1.62cm, xshift=-1.65cm]plot_0.south){\labelsize{100}};
		\node(lab0)[anchor=north,draw=none,ultra thick,inner sep=0, outer sep=0]at([yshift=1.62cm, xshift=-0.66cm]plot_0.south){\labelsize{200}};
		\node(lab0)[anchor=north,draw=none,ultra thick,inner sep=0, outer sep=0]at([yshift=1.62cm, xshift=1.2cm]plot_0.south){\labelsize{500}};
		\node(lab0)[anchor=north,draw=none,ultra thick,inner sep=0, outer sep=0]at([yshift=1.62cm, xshift=3.78cm]plot_0.south){\labelsize{900}};
		%
		%
		%
		%
		%
		%
		\node(plot_1)[anchor=west, inner sep=0, outer sep=0] at ([yshift=0.0cm, xshift=0.5cm]plot_0.east) 
		{\includegraphics[width=0.48\linewidth]
			{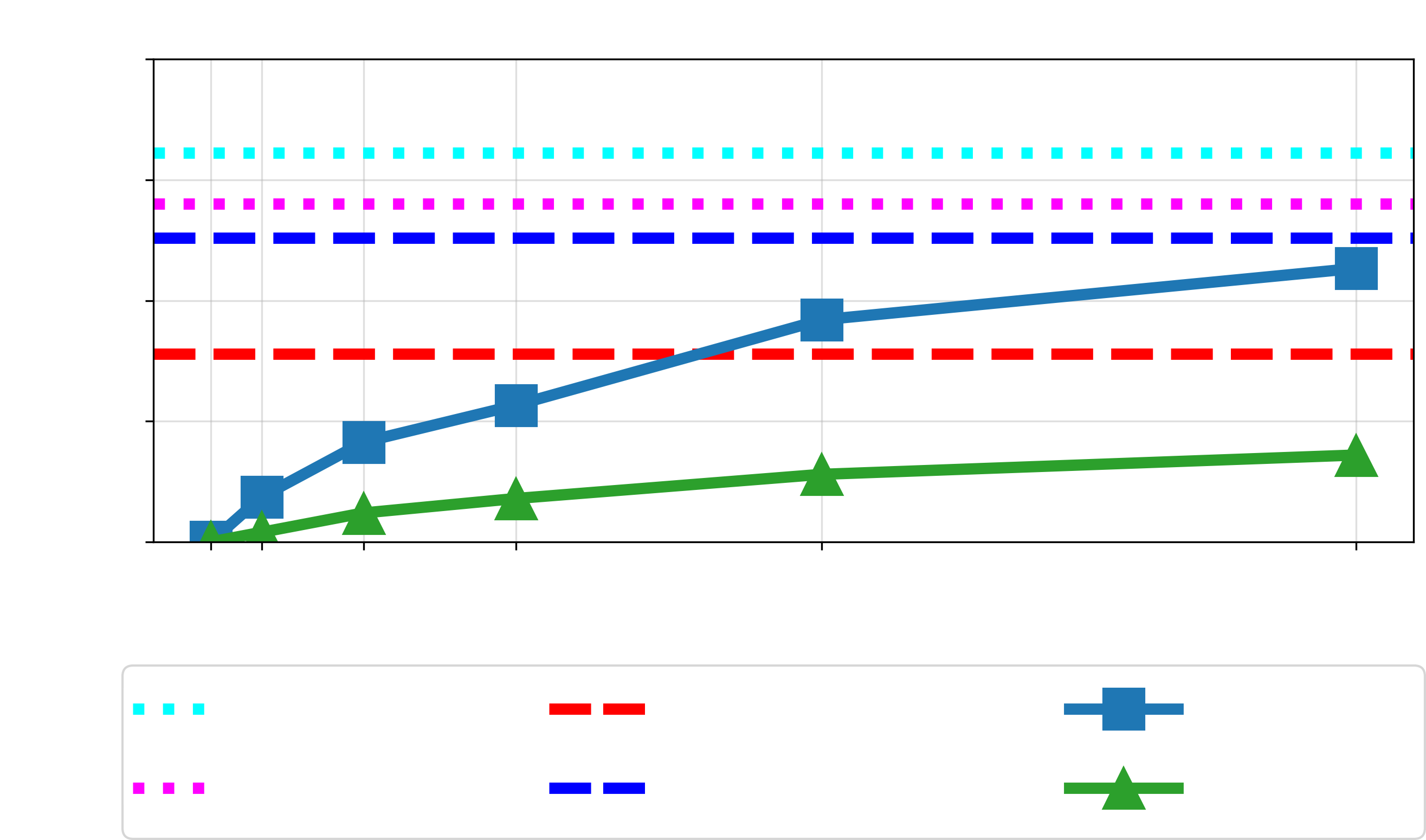}
		};
		\node(b)[anchor=north west,draw=none,ultra thick,inner sep=0, outer sep=0] at([yshift=0.8cm, xshift=-0.01cm]plot_1.south west)
		{\large b) };
		\node(title)[anchor=south,draw=none,ultra thick,inner sep=0, outer sep=0] at([yshift=-0.2cm, xshift=0.4cm]plot_1.north)
		{\textbf{\RobotDB{} Behavior}};
		\node(a)[anchor=north,draw=none,ultra thick,inner sep=0, outer sep=0] at ([yshift=1.34cm, xshift=0.55cm]plot_1.south)
		{\# Expert Demonstrations };
		\node(a)[anchor=north east,draw=none,ultra thick,inner sep=0, outer sep=0] at([yshift=4.4cm, xshift=0.28cm]plot_1.south west)
		{\rotatebox{90}{Success Rate [\%]}};
		\node(legend1)[anchor=north,draw=none,ultra thick,inner sep=0, outer sep=0] at([yshift=0.87cm, xshift=-2.3cm]plot_1.south){\customsize{SOLD}};
		\node(legend2)[anchor=north,draw=none,ultra thick,inner sep=0, outer sep=0] at([yshift=0.87cm, xshift=0.75cm]plot_1.south){\customsize{ResNet Oracle}};
		\node(legend3)[anchor=north,draw=none,ultra thick,inner sep=0, outer sep=0] at([yshift=0.87cm, xshift=3.42cm]plot_1.south){\customsize{\Method}};
		\node(legend4)[anchor=north west,draw=none,ultra thick,inner sep=0, outer sep=0] at([yshift=-0.21cm, xshift=0.0cm]legend1.south west){\customsize{DreamerV3}};
		\node(legend5)[anchor=north west,draw=none,ultra thick,inner sep=0, outer sep=0] at([yshift=-0.21cm, xshift=0.0cm]legend2.south west){\customsize{Slot Oracle}};
		\node(legend6)[anchor=north west,draw=none,ultra thick,inner sep=0, outer sep=0] at([yshift=-0.18cm, xshift=0.0cm]legend3.south west){\customsize{LAPO}};
		\node(lab0)[anchor=east,draw=none,ultra thick,inner sep=0, outer sep=0] at([yshift=-0.35cm, xshift=0.8cm]plot_1.north west){\labelsize{100}};
		\node(lab0)[anchor=east,draw=none,ultra thick,inner sep=0, outer sep=0] at([yshift=1.38cm, xshift=0.8cm]plot_1.west){\labelsize{75}};
		\node(lab0)[anchor=east,draw=none,ultra thick,inner sep=0, outer sep=0] at([yshift=0.68cm, xshift=0.8cm]plot_1.west){\labelsize{50}};
		\node(lab0)[anchor=east,draw=none,ultra thick,inner sep=0, outer sep=0] at([yshift=-0.02cm, xshift=0.8cm]plot_1.west){\labelsize{25}};
		\node(lab0)[anchor=east,draw=none,ultra thick,inner sep=0, outer sep=0] at([yshift=1.75cm, xshift=0.8cm]plot_1.south west){\labelsize{0}};
		\node(lab0)[anchor=north,draw=none,ultra thick,inner sep=0, outer sep=0]at([yshift=1.62cm, xshift=-2.93cm]plot_1.south){\labelsize{0}};
		\node(lab0)[anchor=north,draw=none,ultra thick,inner sep=0, outer sep=0]at([yshift=1.62cm, xshift=-2.55cm]plot_1.south){\labelsize{200}};
		\node(lab0)[anchor=north,draw=none,ultra thick,inner sep=0, outer sep=0]at([yshift=1.62cm, xshift=-1.95cm]plot_1.south){\labelsize{500}};
		\node(lab0)[anchor=north,draw=none,ultra thick,inner sep=0, outer sep=0]at([yshift=1.62cm, xshift=-1.15cm]plot_1.south){\labelsize{1000}};
		\node(lab0)[anchor=north,draw=none,ultra thick,inner sep=0, outer sep=0]at([yshift=1.62cm, xshift=0.65cm]plot_1.south){\labelsize{2000}};
		\node(lab0)[anchor=north,draw=none,ultra thick,inner sep=0, outer sep=0]at([yshift=1.62cm, xshift=3.7cm]plot_1.south){\labelsize{4500}};
	\end{tikzpicture}
	\vspace{-0.25cm}
	\caption{
		 Success rate (\%) as a function of available expert demonstrations on the a) \ButtonPress{} and b) \RobotDB{} environments.
		 \Method{} outperforms LAPO, especially on the challenging \RobotDB{} environment, which requires reasoning about object properties.
	}
	\label{fig: comparison lapo}
	\vspace{-0.4cm}
\end{figure*}

LAPO~\cite{Schmidt_LAPOLearningToActWithoutActions_2024} is a recent model proposed for learning a world model, an inverse dynamics model, and a latent action policy from unlabeled videos.
In contrast to the object-centric representations employed by \Method{}, LAPO relies on feature maps output by a convolutional encoder.

\noindent \textbf{\ButtonPress~Behavior~~}
In \Figure{fig: comparison lapo}a), we compare policies learned by \Method{} and LAPO on the \ButtonPress{} task, using a varying number of expert demonstrations.
As reference baselines, we include two oracle models trained with access to all available expert demonstrations and to ground-truth actions: the Slot Oracle, which uses object-centric representations, and the ResNet Oracle, which uses ResNet feature maps.

Both \Method{} and LAPO perform comparably on the \ButtonPress{} task, which requires limited object-centric reasoning. Nonetheless, \Method{} achieves slightly better sample-efficiency and consistently higher performance than LAPO across most data regimes.
Notably, both methods approach the performance of the Oracle models (which have access to ground-truth actions during training) with a limited number of expert demonstrations. This is especially notable for \Method, which matches the performance of the ResNet oracle with as few as 500 demonstrations.

\noindent \textbf{\RobotDB~Behavior~~}
On the challenging \RobotDB{} task, which requires reasoning over object properties and relations, both models learn policies from a varying number of expert demonstrations, with only ~80\% depicting successful task completions---adding further difficulty due to noisy supervision.

As reference baselines, we include two model-based reinforcement learning methods, using holistic (DreamerV3~\cite{hafner2023mastering}) and object-centric (SOLD~\cite{Mosbach_SOLDReinforcementLearningSlotObjectCentricLatentDynamics}) latent spaces, both of which learn the robot behavior by directly interacting with the environment and using action and reward information.
Additionally, we include two oracle behavior cloning baselines trained on all expert demonstrations, with access to ground-truth actions, and leveraging object-centric representations (Slot Oracle) or ResNet feature maps (ResNet Oracle).

\Figure{fig: comparison lapo}b) shows that \Method{} consistently outperforms LAPO by a large margin across all data regimes, demonstrating much stronger sample-efficiency and substantially higher final performance.
While LAPO struggles to scale with the number of demonstrations, \Method{} shows steady improvement, narrowing the gap between latent action models and baseline methods trained with access to ground-truth actions.
Notably, \Method{} even outperforms the ResNet Oracle model with as few as 2000 noisy expert demonstrations, and closely approaches the performance of the strong Slot Oracle baseline.

Additionally, we observe that slot-based object-centric models outperform their holistic counterparts, i.e. SOLD outperforms DreamerV3, Slot Oracle outperforms ResNet Oracle, and \Method{} outperforms LAPO; further emphasizing the advantage of structured slot-based latent spaces.

These results highlight the benefits of object-centric representations for sample-efficient behavior learning, especially in tasks that require understanding object properties and their relations.
By parsing the scene into individual objects, \Method{} is able to generalize more effectively from limited and noisy demonstrations and infer complex behaviors, which are often challenging for models relying on monolithic, holistic representations.

\subsubsection{Visualizations of Learned Behaviors}
\label{app: vis learned behaviors}

\begin{figure}[b]
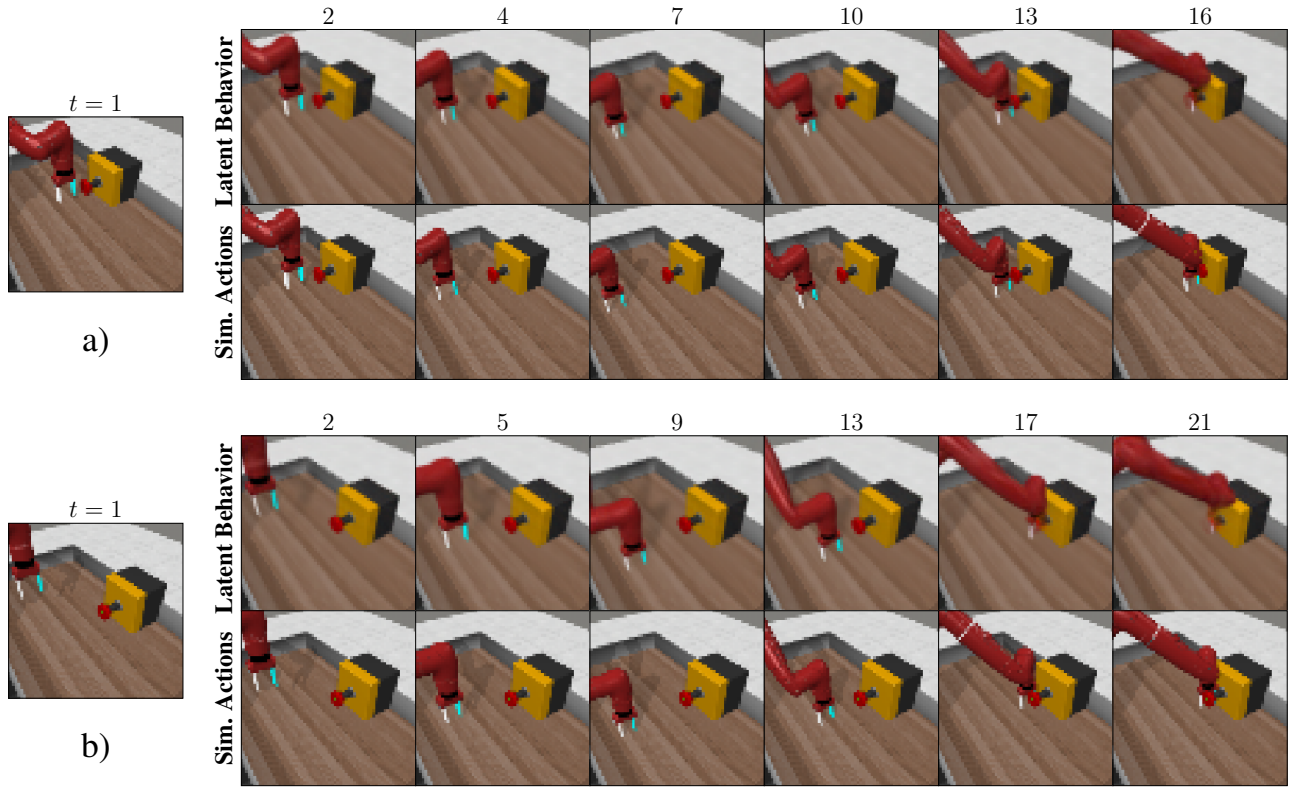

	
	\input{./imgs/supplementary/behaviors_button/button_00.tex}
	
	\input{./imgs/supplementary/behaviors_button/button_01.tex}

	\vspace{-0.4cm}
	\caption{
		Predicted frames using latent actions from the learned policy, and sequences simulated by executing the decoded latent actions for two \ButtonPress{} scenarios.
		\textbf{a)} \Method{} successfully plans the trajectory within its latent imagination and translates latent actions into executable motions.
		\textbf{b)} \Method{} fails to decode its latent actions into correct robot motion, missing the button.
	}
	\label{fig: button beh 00}
\end{figure}

\noindent \textbf{\ButtonPress~Behavior:}
%
\Figure{fig: button beh 00} illustrates \Method's learned behavior in the \ButtonPress~environment.
The top row in each sequence (labeled as \emph{Latent Behavior}) shows predicted trajectories within \Method's latent imagination, where the model autoregressively generates actions using the policy model and predicts future scene states in the latent space.
The bottom row (labeled \emph{Sim.~Actions}) depicts the simulated execution of the decoded latent actions in the environment.
\Method{} learns to solve the task within its latent imagination, successfully reasoning about the required action sequences before translating its latent actions into executable motions.
\Figure{fig: button beh 00}b) shows a failure case where \Method{} predicts within its latent imagination a trajectory that leads to successfully pressing the button. However, accumulated errors in the world model and during action decoding cause the simulated execution to miss the button.

\begin{figure}[t]
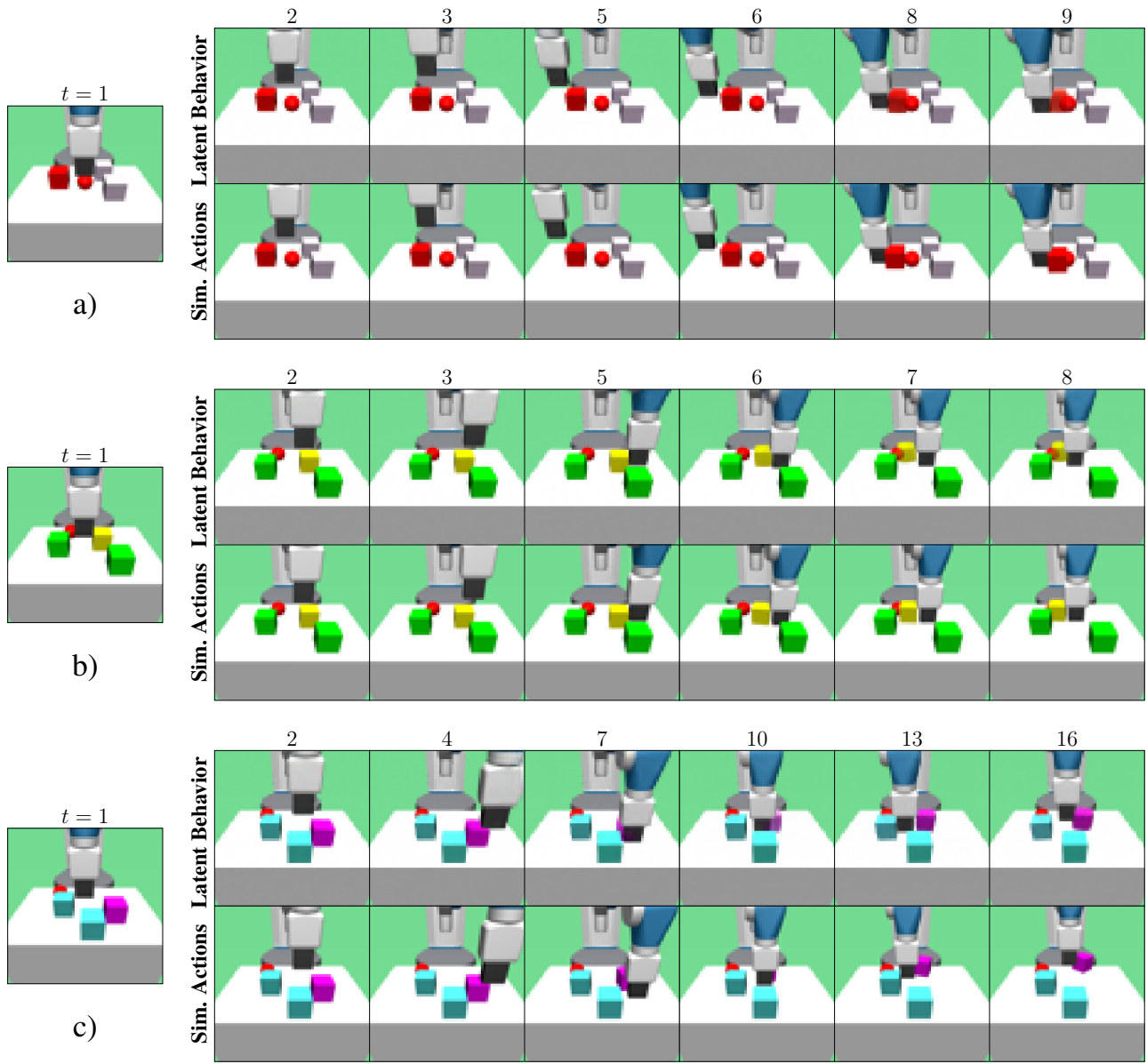

	
	\input{./imgs/supplementary/behaviors_robotdb/robotdb_00.tex}
	
	\input{./imgs/supplementary/behaviors_robotdb/robotdb_01.tex}
	
	\input{./imgs/supplementary/behaviors_robotdb/robotdb_02.tex}
	
	\vspace{-0.4cm}
	\caption{
		Predicted frames using latent actions from the learned policy, and simulation computed by executing the decoded latent actions for three \RobotDB~scenarios.
		\textbf{a) \& b)} \Method~identifies the block of distinct color and generates executable latent actions to push it to the target location.
		\textbf{c)} \Method~identifies the block but fails to push it into the target location.
	}
	\label{fig: robot beh 00}
\end{figure}

\noindent \textbf{\RobotDB~Behavior:}
%
\Figure{fig: robot beh 00} illustrates \Method's learned behavior in the \RobotDB~environment.
The top row in each sequence (labeled as \emph{Latent Behavior}) shows predicted trajectories within \Method's latent imagination, where the model autoregressively generates actions using the policy model and predicts future scene states in the latent space.
The bottom row (labeled \emph{Sim.~Actions}) depicts the simulated execution of the decoded latent actions in the environment.
\Method{} learns to solve the task within its latent imagination, successfully reasoning about object properties and generating a precise sequence of latent actions, which can be decoded into executable motions.
\Figure{fig: robot beh 00}c) shows a failure case where \Method{} controls the robot to interact with the correct block, but fails to place it in the target location.

\subsection{Qualitative Results}

In the following sections, we provide additional qualitative results on all datasets, as well as qualitative comparisons with baselines.
Further qualitative results and animations are provided \href{https://play-slot.github.io/PlaySlot/}{in the project website}.

\subsubsection{Comparison with Baselines}

\input{./imgs/qual_comb_v2.tex}

\Figure{fig: qual_00} depicts a qualitative comparison between SVG, CADDY and $\Method${} 
on the \ButtonPress{} and \RobotDB{} datasets, respectively.
On the  \ButtonPress{} dataset, as shown in \Figure{fig: qual_00}a), all methods accurately model the trajectory of the robot arm.
However, on the complex \RobotDB{} task, depicted in \Figure{fig: qual_00}b), SVG and CADDY fail to model the object collisions, failing to move the block of distinct color to the target location, and leading to blurriness and disappearing objects.
In contrast, $\Method${} maintains sharp object representations and correctly models interactions between objects, leading to accurate frame predictions.

\subsubsection{\RobotDB~Dataset}

\begin{figure}[t]
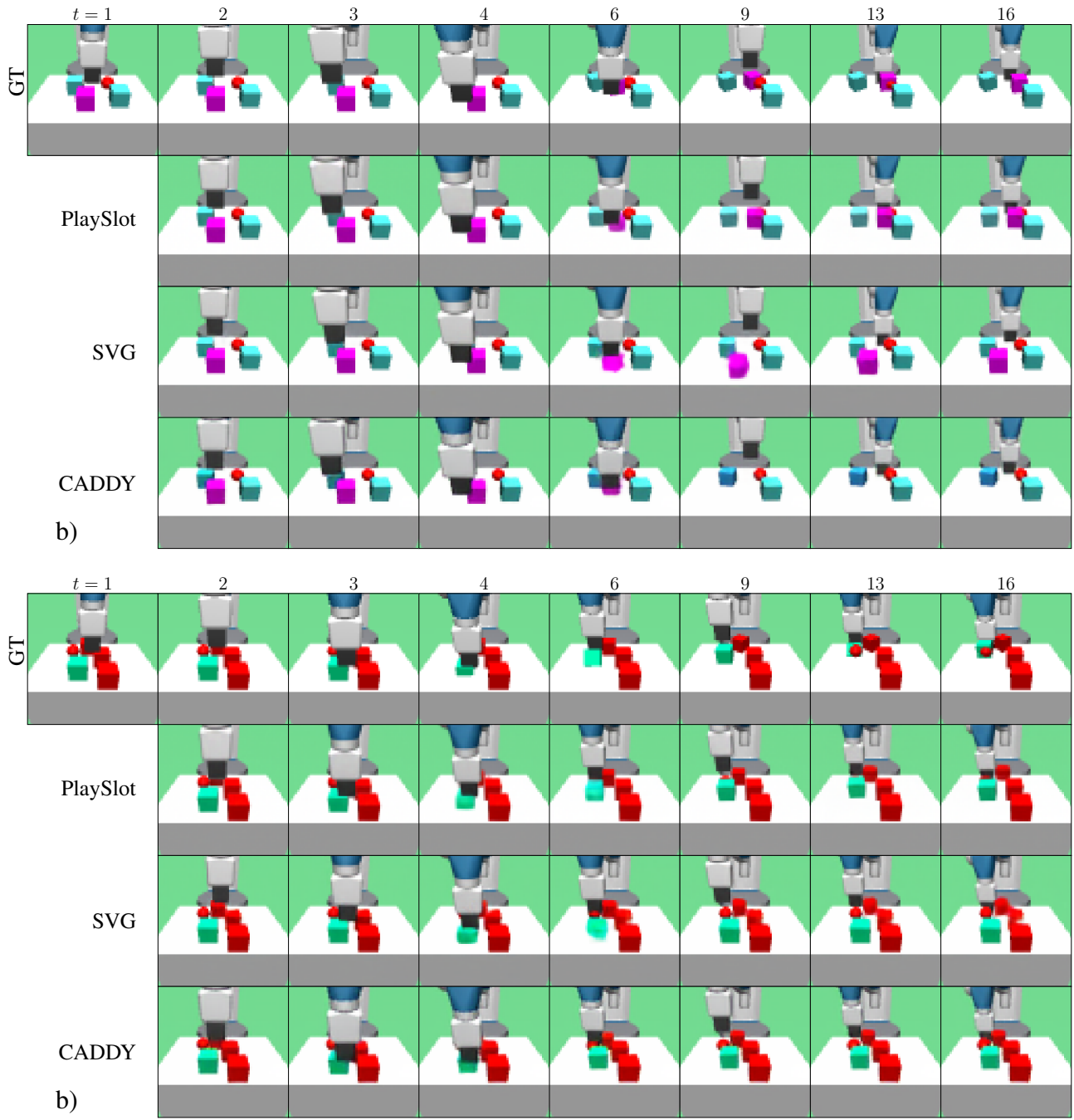

	
	\input{./imgs/supplementary/qual_robot_02/qual_02.tex}
	
	\input{./imgs/supplementary/qual_robot_01/qual_01.tex}
	
	\caption{
		Qualitative comparison on \RobotDB.
		Our method accurately predicts the scene dynamics and object interactions, whereas the baselines fail to model object collisions, leading to blurry or disappearing objects.
	}
	\label{fig: robot qual}
\end{figure}

\Figure{fig: robot qual} shows two qualitative comparisons between \Method, CADDY and SVG on the \RobotDB~dataset.
Our proposed method, which explicitly models object interactions, preserves sharp object representations and accurately predicts future frames.
In contrast, SVG and CADDY, which rely on holistic scene features for forecasting, struggle to model object collisions, resulting in blurry predictions and disappearing objects.

\begin{figure}[t]
	\resizebox{1.0 \linewidth}{!}{
		\begin{tikzpicture} 
			\node(P0)[fill=none] {};
			%
			%
			\node(gt_00)[anchor=north west, inner sep=0, outer sep=0] at (P0) 
			{\includegraphics[height=3cm]
				{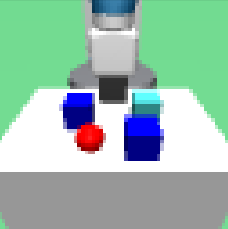}
			};
			\draw[black](gt_00.south west) rectangle (gt_00.north east);
			\node(0)[anchor=south,draw=none,ultra thick,inner sep=0, outer sep=0] at
			([xshift=-0.25cm, yshift=-1.65cm]gt_00.north west)
			{{\rotatebox{90}{\Large GT}}};
			\node(gt_01)[anchor=west, inner sep=0, outer sep=0] at
			([xshift=0.00cm]gt_00.east) 
			{\includegraphics[height=3cm]
				{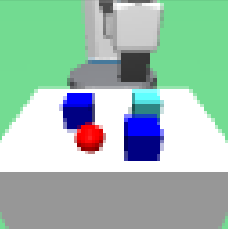}
			};
			\draw[black](gt_01.south west) rectangle (gt_01.north east);
			\node(gt_02)[anchor=west, inner sep=0, outer sep=0] at
			([xshift=0.00cm]gt_01.east) 
			{\includegraphics[height=3cm]
				{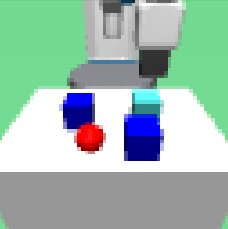}
			};
			\draw[black](gt_02.south west) rectangle (gt_02.north east);
			\node(gt_03)[anchor=west, inner sep=0, outer sep=0] at
			([xshift=0.00cm]gt_02.east) 
			{\includegraphics[height=3cm]
				{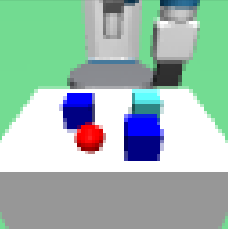}
			};
			\draw[black](gt_03.south west) rectangle (gt_03.north east);
			\node(gt_05)[anchor=west, inner sep=0, outer sep=0] at
			([xshift=0.00cm]gt_03.east)
			{\includegraphics[height=3cm]
				{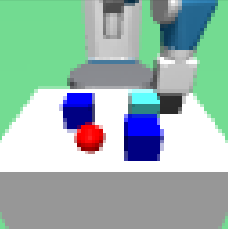}
			};
			\draw[black](gt_05.south west) rectangle (gt_05.north east);
			\node(gt_08)[anchor=west, inner sep=0, outer sep=0] at
			([xshift=0.00cm]gt_05.east)
			{\includegraphics[height=3cm]
				{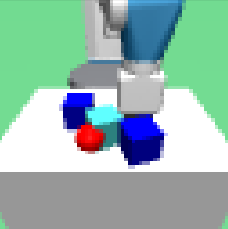}
			};
			\draw[black](gt_08.south west) rectangle (gt_08.north east);
			\node(gt_10)[anchor=west, inner sep=0, outer sep=0] at
			([xshift=0.00cm]gt_08.east)
			{\includegraphics[height=3cm]
				{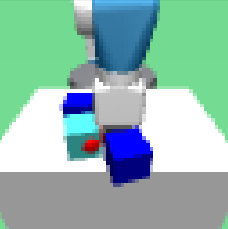}
			};
			\draw[black](gt_10.south west) rectangle (gt_10.north east);
			\node(gt_12)[anchor=west, inner sep=0, outer sep=0] at
			([xshift=0.00cm]gt_10.east)
			{\includegraphics[height=3cm]
				{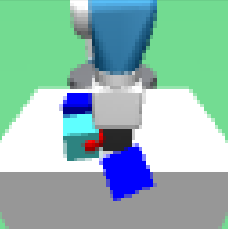}
			};
			\draw[black](gt_12.south west) rectangle (gt_12.north east);
			\node(gt_15)[anchor=west, inner sep=0, outer sep=0] at
			([xshift=0.00cm]gt_12.east)
			{\includegraphics[height=3cm]
				{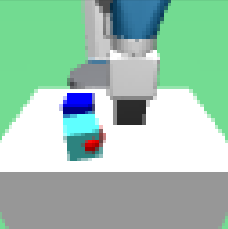}
			};
			\draw[black](gt_15.south west) rectangle (gt_15.north east);
			%
			%
			%
			%
			%
			\node(playslot_01)[anchor=north, inner sep=0, outer sep=0] at
			([xshift=0.00cm]gt_01.south) 
			{\includegraphics[height=3cm]
				{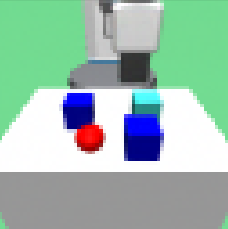}
			};
			\draw[black](playslot_01.south west) rectangle (playslot_01.north east);
			\node(playslot_02)[anchor=north, inner sep=0, outer sep=0] at
			([xshift=0.00cm]gt_02.south) 
			{\includegraphics[height=3cm]
				{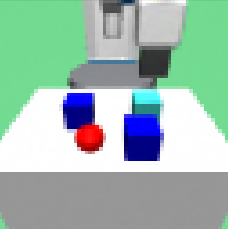}
			};
			\draw[black](playslot_02.south west) rectangle (playslot_02.north east);
			\node(playslot_03)[anchor=north, inner sep=0, outer sep=0] at
			([xshift=0.00cm]gt_03.south) 
			{\includegraphics[height=3cm]
				{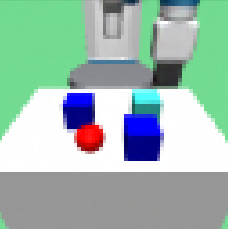}
			};
			\draw[black](playslot_03.south west) rectangle (playslot_03.north east);
			\node(playslot_05)[anchor=north, inner sep=0, outer sep=0] at
			([xshift=0.00cm]gt_05.south) 
			{\includegraphics[height=3cm]
				{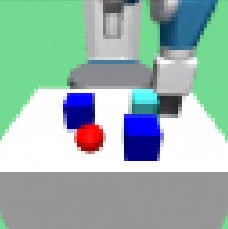}
			};
			\draw[black](playslot_05.south west) rectangle (playslot_05.north east);
			\node(playslot_08)[anchor=north, inner sep=0, outer sep=0] at
			([xshift=0.00cm]gt_08.south) 
			{\includegraphics[height=3cm]
				{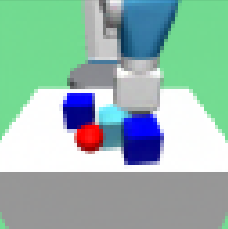}
			};
			\draw[black](playslot_08.south west) rectangle (playslot_08.north east);
			\node(playslot_10)[anchor=north, inner sep=0, outer sep=0] at
			([xshift=0.00cm]gt_10.south) 
			{\includegraphics[height=3cm]
				{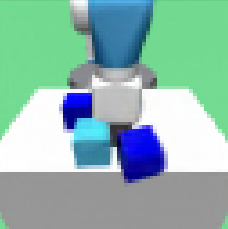}
			};
			\draw[black](playslot_10.south west) rectangle (playslot_10.north east);
			\node(playslot_12)[anchor=north, inner sep=0, outer sep=0] at
			([xshift=0.00cm]gt_12.south) 
			{\includegraphics[height=3cm]
				{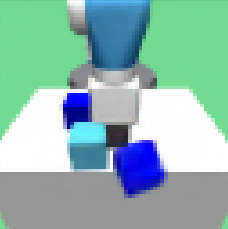}
			};
			\draw[black](playslot_12.south west) rectangle (playslot_12.north east);
			\node(playslot_15)[anchor=north, inner sep=0, outer sep=0] at
			([xshift=0.00cm]gt_15.south) 
			{\includegraphics[height=3cm]
				{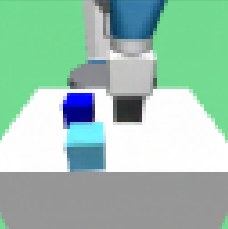}
			};
			\draw[black](playslot_15.south west) rectangle (playslot_15.north east);
			\node(PlaySlotLabel)[anchor=east,draw=none,ultra thick,inner sep=0, outer sep=0] at 	([xshift=-0.5cm, yshift=0.0cm]playslot_00.west)
			{\Large{Predictions}};
			%
			%
			%
			%
			\foreach \i [count=\j] in {01,02,03,05,08,10,12,15} {
				\node(segm_\i)[anchor=north, inner sep=0, outer sep=0] at 
				([xshift=0.00cm]playslot_\i.south) 
				{\includegraphics[height=3cm]
					{./imgs/supplementary/qual_robot_00/PlaySlot_ind_figs/segm_\i.png}
				};
				\draw[black](segm_\i.south west) rectangle (segm_\i.north east);
			}
			\node(PlaySlotLabel)[anchor=east,draw=none,ultra thick,inner sep=0, outer sep=0] at 
			([xshift=-0.5cm, yshift=0.0cm]segm_01.west)
			{\makecell{\Large{Slot}\\ \Large{Masks}}};
			%
			%
			%
			\foreach \i [count=\j] in {01,02,03,05,08,10,12,15} {
				\node(obj_0_\j)[anchor=north, inner sep=0, outer sep=0] at 
				([xshift=0.00cm]segm_\i.south) 
				{\includegraphics[height=3cm]
					{./imgs/supplementary/qual_robot_00/PlaySlot_ind_figs/MaskedObj_00_\i.png}
				};
				\draw[black](obj_0_\j.south west) rectangle (obj_0_\j.north east);
			}
			\node(PlaySlotLabel)[anchor=east,draw=none,ultra thick,inner sep=0, outer sep=0] at
			([xshift=-0.5cm, yshift=0.0cm]obj_0_1.west)
			{\makecell{\Large{Object 1}}};
			%
			%
			%
			\foreach \i [count=\j] in {01,02,03,05,08,10,12,15} {
				\node(obj_1_\j)[anchor=north, inner sep=0, outer sep=0] at 
				([xshift=0.00cm]obj_0_\j.south) 
				{\includegraphics[height=3cm]
					{./imgs/supplementary/qual_robot_00/PlaySlot_ind_figs/MaskedObj_01_\i.png}
				};
				\draw[black](obj_1_\j.south west) rectangle (obj_1_\j.north east);
			}
			\node(PlaySlotLabel)[anchor=east,draw=none,ultra thick,inner sep=0, outer sep=0] at
			([xshift=-0.5cm, yshift=0.0cm]obj_1_1.west)
			{\makecell{\Large{Object 2}}};
			%
			%
			\foreach \i [count=\j] in {01,02,03,05,08,10,12,15} {
				\node(obj_2_\j)[anchor=north, inner sep=0, outer sep=0] at 
				([xshift=0.00cm]obj_1_\j.south) 
				{\includegraphics[height=3cm]
					{./imgs/supplementary/qual_robot_00/PlaySlot_ind_figs/MaskedObj_02_\i.png}
				};
				\draw[black](obj_2_\j.south west) rectangle (obj_2_\j.north east);
			}
			\node(PlaySlotLabel)[anchor=east,draw=none,ultra thick,inner sep=0, outer sep=0] at
			([xshift=-0.5cm, yshift=0.0cm]obj_2_1.west)
			{\makecell{\Large{Object 3}}};
			%
			%
			%
			\foreach \i [count=\j] in {01,02,03,05,08,10,12,15} {
				\node(obj_4_\j)[anchor=north, inner sep=0, outer sep=0] at 
				([xshift=0.00cm]obj_2_\j.south) 
				{\includegraphics[height=3cm]
					{./imgs/supplementary/qual_robot_00/PlaySlot_ind_figs/MaskedObj_04_\i.png}
				};
				\draw[black](obj_4_\j.south west) rectangle (obj_4_\j.north east);
			}
			\node(PlaySlotLabel)[anchor=east,draw=none,ultra thick,inner sep=0, outer sep=0] at
			([xshift=-0.5cm, yshift=0.0cm]obj_4_1.west)
			{\makecell{\Large{Object 4}}};
			%
			%
			%
			\foreach \i [count=\j] in {01,02,03,05,08,10,12,15} {
				\node(obj_6_\j)[anchor=north, inner sep=0, outer sep=0] at 
				([xshift=0.00cm]obj_4_\j.south) 
				{\includegraphics[height=3cm]
					{./imgs/supplementary/qual_robot_00/PlaySlot_ind_figs/MaskedObj_06_\i.png}
				};
				\draw[black](obj_6_\j.south west) rectangle (obj_6_\j.north east);
			}
			\node(PlaySlotLabel)[anchor=east,draw=none,ultra thick,inner sep=0, outer sep=0] at
			([xshift=-0.5cm, yshift=0.0cm]obj_6_1.west)
			{\makecell{\Large{Object 5}}};
			%
			%
			%
			%
			\foreach \i [count=\j] in {01,02,03,05,08,10,12,15} {
				\node(obj_7_\j)[anchor=north, inner sep=0, outer sep=0] at 
				([xshift=0.00cm]obj_6_\j.south) 
				{\includegraphics[height=3cm]
					{./imgs/supplementary/qual_robot_00/PlaySlot_ind_figs/MaskedObj_07_\i.png}
				};
				\draw[black](obj_7_\j.south west) rectangle (obj_7_\j.north east);
			}
			\node(PlaySlotLabel)[anchor=east,draw=none,ultra thick,inner sep=0, outer sep=0] at
			([xshift=-0.5cm, yshift=0.0cm]obj_7_1.west)
			{\makecell{\Large{Object 6}}};
			%
			%
			%
			%
			\node(0)[anchor=south,draw=none,ultra thick,inner sep=0, outer sep=0] at 	([yshift=0.1cm]gt_00.north)
			{\large{$t=1$}};
			\node(0)[anchor=south,draw=none,ultra thick,inner sep=0, outer sep=0] at 	([yshift=0.1cm]gt_01.north)
			{\large{$2$}};
			\node(0)[anchor=south,draw=none,ultra thick,inner sep=0, outer sep=0] at 	([yshift=0.1cm]gt_02.north)
			{\large{$3$}};
			\node(0)[anchor=south,draw=none,ultra thick,inner sep=0, outer sep=0] at 	([yshift=0.1cm]gt_03.north)
			{\large{$4$}};
			\node(0)[anchor=south,draw=none,ultra thick,inner sep=0, outer sep=0] at 	([yshift=0.1cm]gt_05.north)
			{\large{$6$}};
			\node(0)[anchor=south,draw=none,ultra thick,inner sep=0, outer sep=0] at 	([yshift=0.1cm]gt_08.north)
			{\large{$9$}};
			\node(0)[anchor=south,draw=none,ultra thick,inner sep=0, outer sep=0] at 	([yshift=0.1cm]gt_10.north)
			{\large{$11$}};
			\node(0)[anchor=south,draw=none,ultra thick,inner sep=0, outer sep=0] at 	([yshift=0.1cm]gt_12.north)
			{\large{$13$}};
			\node(0)[anchor=south,draw=none,ultra thick,inner sep=0, outer sep=0] at 	([yshift=0.1cm]gt_15.north)
			{\large{$16$}};
		\end{tikzpicture}
	}
\vspace{-0.7cm}
\caption{
		\Method~predictions and object representations on a \RobotDB~sequence.
		We visualize the ground truth sequence, the predicted frames, segmentation obtained by assigning a different color to each slot mask, as well as the object reconstructions for five slots.
		\Method~assigns one slot to the background, one slot for the robot, one slot for the red target, and the remaining slots to the blocks.
	}
\label{fig: sup robot objs 00}
\end{figure}

\Figure{fig: sup robot objs 00} shows the predicted video frames, slot masks, and objects representations on a \RobotDB~sequence.
\Method{} parses the scene into precise object images and masks, which can be assigned a unique color to obtain a segmentation of the scene.
Our approach decomposes the \RobotDB~environment using eight object slots, where one slot represents the background, five slots to different blocks, one slot to the red target, and one slot to the robot arm.
The sharp object images and masks demonstrate that \Method{} encodes into each slot features from the corresponding object.
This allows our method to directly reason about object properties, dynamics and interactions, allowing for accurate future frame predictions.

\begin{figure}[t]
	\resizebox{1.0 \linewidth}{!}{
		\begin{tikzpicture} 
			\node(P0)[fill=none] {};
			%
			%
			\node(seed)[anchor=north west, inner sep=0, outer sep=0] at (P0) 
			{\includegraphics[height=3cm]
				{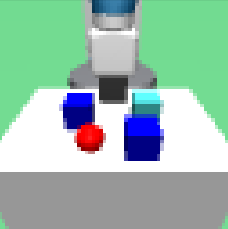}
			};
			\draw[black](gt_00.south west) rectangle (gt_00.north east);
			\node(0)[anchor=south,draw=none,ultra thick,inner sep=0, outer sep=0] at
			([xshift=-0.25cm, yshift=-1.65cm]seed.north west)
			{{\rotatebox{90}{\Large GT}}};
			\node(gt_1)[anchor=west, inner sep=0, outer sep=0] at
			([xshift=0.00cm]seed.east) 
			{\includegraphics[height=3cm]
				{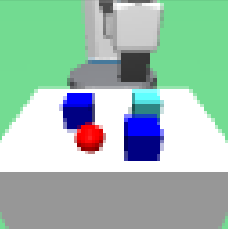}
			};
			\draw[black](gt_1.south west) rectangle (gt_1.north east);
			\node(gt_2)[anchor=west, inner sep=0, outer sep=0] at
			([xshift=0.00cm]gt_1.east) 
			{\includegraphics[height=3cm]
				{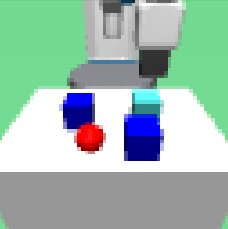}
			};
			\draw[black](gt_2.south west) rectangle (gt_2.north east);
			\node(gt_3)[anchor=west, inner sep=0, outer sep=0] at
			([xshift=0.00cm]gt_2.east) 
			{\includegraphics[height=3cm]
				{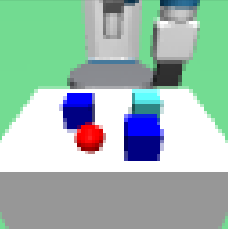}
			};
			\draw[black](gt_3.south west) rectangle (gt_3.north east);
			\node(gt_4)[anchor=west, inner sep=0, outer sep=0] at
			([xshift=0.00cm]gt_3.east)
			{\includegraphics[height=3cm]
				{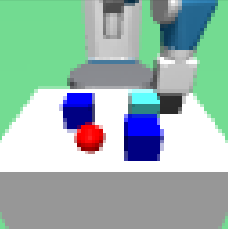}
			};
			\draw[black](gt_4.south west) rectangle (gt_4.north east);
			\node(gt_5)[anchor=west, inner sep=0, outer sep=0] at
			([xshift=0.00cm]gt_4.east)
			{\includegraphics[height=3cm]
				{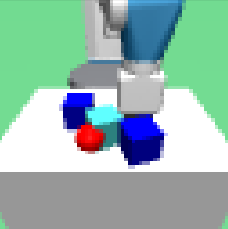}
			};
			\draw[black](gt_5.south west) rectangle (gt_5.north east);
			\node(gt_6)[anchor=west, inner sep=0, outer sep=0] at
			([xshift=0.00cm]gt_5.east)
			{\includegraphics[height=3cm]
				{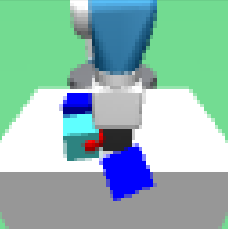}
			};
			\draw[black](gt_6.south west) rectangle (gt_6.north east);
			\node(gt_7)[anchor=west, inner sep=0, outer sep=0] at
			([xshift=0.00cm]gt_6.east)
			{\includegraphics[height=3cm]
				{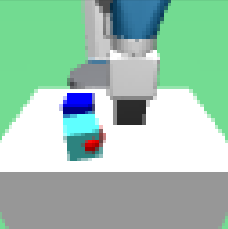}
			};
			\draw[black](gt_7.south west) rectangle (gt_7.north east);
			%
			%
			%
			\foreach \i [count=\j] in {00,01,02,04,07,11,14} {
				\node(post_\i)[anchor=north, inner sep=0, outer sep=0] at 
				([xshift=0.00cm]gt_\j.south) 
				{\includegraphics[height=3cm]
					{./imgs/supplementary/qual_robot_00/PlaySlot_act_figs/post_\i.png}
				};
				\draw[black](post_\i.south west) rectangle (post_\i.north east);
			}
			\node(PlaySlotLabel)[anchor=east,draw=none,ultra thick,inner sep=0, outer sep=0] at 
			([xshift=-0.5cm, yshift=0.0cm]post_00.west)
			{\makecell{ {\Large Inferred} \\ {\Large Dynamics} }};
			%
			%
			%
			\foreach \i [count=\j] in {00,01,02,04,07,11,14} {
				\node(prior_1_\i)[anchor=north, inner sep=0, outer sep=0] at 
				([xshift=0.00cm]post_\i.south) 
				{\includegraphics[height=3cm]
					{./imgs/supplementary/qual_robot_00/PlaySlot_act_figs/prior_1_\i.png}
				};
				\draw[black](prior_1_\i.south west) rectangle (prior_1_\i.north east);
			}
			\node(PlaySlotLabel)[anchor=east,draw=none,ultra thick,inner sep=0, outer sep=0] at 
			([xshift=-0.2cm, yshift=0.0cm]prior_1_00.west)
			{\makecell{ {\Large `Up \& Left'} \\ {Action 1} }};
			%
			%
			%
			\foreach \i [count=\j] in {00,01,02,04,07,11,14} {
				\node(prior_2_\i)[anchor=north, inner sep=0, outer sep=0] at 
				([xshift=0.00cm]prior_1_\i.south) 
				{\includegraphics[height=3cm]
					{./imgs/supplementary/qual_robot_00/PlaySlot_act_figs/prior_2_\i.png}
				};
				\draw[black](prior_2_\i.south west) rectangle (prior_2_\i.north east);
			}
			\node(PlaySlotLabel)[anchor=east,draw=none,ultra thick,inner sep=0, outer sep=0] at 
			([xshift=-0.9cm, yshift=0.0cm]prior_2_00.west)
			{\makecell{ {\Large `Right'} \\ {Action 2} }};
			%
			%
			%
			\foreach \i [count=\j] in {00,01,02,04,07,11,14} {
				\node(prior_3_\i)[anchor=north, inner sep=0, outer sep=0] at 
				([xshift=0.00cm]prior_2_\i.south) 
				{\includegraphics[height=3cm]
					{./imgs/supplementary/qual_robot_00/PlaySlot_act_figs/prior_3_\i.png}
				};
				\draw[black](prior_3_\i.south west) rectangle (prior_3_\i.north east);
			}
			\node(PlaySlotLabel)[anchor=east,draw=none,ultra thick,inner sep=0, outer sep=0] at 
			([xshift=-0.9cm, yshift=0.0cm]prior_3_00.west)
			{\makecell{ {\Large `Stay'} \\ {Action 3} }};
			%
			%
			%
			%
			\foreach \i [count=\j] in {00,01,02,04,07,11,14} {
				\node(prior_4_\i)[anchor=north, inner sep=0, outer sep=0] at 
				([xshift=0.00cm]prior_3_\i.south) 
				{\includegraphics[height=3cm]
					{./imgs/supplementary/qual_robot_00/PlaySlot_act_figs/prior_4_\i.png}
				};
				\draw[black](prior_4_\i.south west) rectangle (prior_4_\i.north east);
			}
			\node(PlaySlotLabel)[anchor=east,draw=none,ultra thick,inner sep=0, outer sep=0] at 
			([xshift=-0.2cm, yshift=0.0cm]prior_4_00.west)
			{\makecell{ {\Large `Up \& Right'} \\ {Action 4} }};
			%
			%
			%
			%
			\foreach \i [count=\j] in {00,01,02,04,07,11,14} {
				\node(prior_5_\i)[anchor=north, inner sep=0, outer sep=0] at 
				([xshift=0.00cm]prior_4_\i.south) 
				{\includegraphics[height=3cm]
					{./imgs/supplementary/qual_robot_00/PlaySlot_act_figs/prior_5_\i.png}
				};
				\draw[black](prior_5_\i.south west) rectangle (prior_5_\i.north east);
			}
			\node(PlaySlotLabel)[anchor=east,draw=none,ultra thick,inner sep=0, outer sep=0] at 
			([xshift=-0.9cm, yshift=0.0cm]prior_5_00.west)
			{\makecell{ {\Large `Left'} \\ {Action 5} }};
			%
			%
			%
			%
			\foreach \i [count=\j] in {00,01,02,04,07,11,14} {
				\node(prior_6_\i)[anchor=north, inner sep=0, outer sep=0] at 
				([xshift=0.00cm]prior_5_\i.south) 
				{\includegraphics[height=3cm]
					{./imgs/supplementary/qual_robot_00/PlaySlot_act_figs/prior_6_\i.png}
				};
				\draw[black](prior_6_\i.south west) rectangle (prior_6_\i.north east);
			}
			\node(PlaySlotLabel)[anchor=east,draw=none,ultra thick,inner sep=0, outer sep=0] at 
			([xshift=-0.9cm, yshift=0.0cm]prior_6_00.west)
			{\makecell{ {\Large `Up'} \\ {Action 6} }};
			%
			%
			
			%
			%
			\foreach \i [count=\j] in {00,01,02,04,07,11,14} {
				\node(prior_7_\i)[anchor=north, inner sep=0, outer sep=0] at 
				([xshift=0.00cm]prior_6_\i.south) 
				{\includegraphics[height=3cm]
					{./imgs/supplementary/qual_robot_00/PlaySlot_act_figs/prior_7_\i.png}
				};
				\draw[black](prior_7_\i.south west) rectangle (prior_7_\i.north east);
			}
			\node(PlaySlotLabel)[anchor=east,draw=none,ultra thick,inner sep=0, outer sep=0] at 
			([xshift=-0.9cm, yshift=0.0cm]prior_7_00.west)
			{\makecell{ {\Large `Left'} \\ {Action 7} }};
			%
			%
			%
			%
			\foreach \i [count=\j] in {00,01,02,04,07,11,14} {
				\node(prior_8_\i)[anchor=north, inner sep=0, outer sep=0] at 
				([xshift=0.00cm]prior_7_\i.south) 
				{\includegraphics[height=3cm]
					{./imgs/supplementary/qual_robot_00/PlaySlot_act_figs/prior_8_\i.png}
				};
				\draw[black](prior_8_\i.south west) rectangle (prior_8_\i.north east);
			}
			\node(PlaySlotLabel)[anchor=east,draw=none,ultra thick,inner sep=0, outer sep=0] at 
			([xshift=-0.1cm, yshift=0.0cm]prior_8_00.west)
			{\makecell{ {\Large `Up \& Right'} \\ {Action 8} }};
			%
			%
			%
			%
			%
			%
			%
			%
			\node(0)[anchor=south,draw=none,ultra thick,inner sep=0, outer sep=0] at 	([yshift=0.1cm]gt_00.north)
			{\large{$t=1$}};
			\node(0)[anchor=south,draw=none,ultra thick,inner sep=0, outer sep=0] at 	([yshift=0.1cm]gt_01.north)
			{\large{$2$}};
			\node(0)[anchor=south,draw=none,ultra thick,inner sep=0, outer sep=0] at 	([yshift=0.1cm]gt_02.north)
			{\large{$3$}};
			\node(0)[anchor=south,draw=none,ultra thick,inner sep=0, outer sep=0] at 	([yshift=0.1cm]gt_03.north)
			{\large{$4$}};
			\node(0)[anchor=south,draw=none,ultra thick,inner sep=0, outer sep=0] at 	([yshift=0.1cm]gt_4.north)
			{\large{$6$}};
			\node(0)[anchor=south,draw=none,ultra thick,inner sep=0, outer sep=0] at 	([yshift=0.1cm]gt_5.north)
			{\large{$9$}};
			\node(0)[anchor=south,draw=none,ultra thick,inner sep=0, outer sep=0] at 	([yshift=0.1cm]gt_6.north)
			{\large{$13$}};
			\node(0)[anchor=south,draw=none,ultra thick,inner sep=0, outer sep=0] at 	([yshift=0.1cm]gt_7.north)
			{\large{$16$}};
		\end{tikzpicture}
	}
	\vspace{-0.4cm}
	\caption{
		 \Method~predictions conditioned on different latent actions, including the inferred inverse dynamics, as well as each action prototype learned on \RobotDB.
		 We generate a sequence by repeatedly conditioning the prediction process on a a single action prototype.
		 The model learns action prototypes that control the robot to move consistently on a specific direction.
	}
	\label{fig: sup acts robot 00}
\end{figure}

\Figure{fig: sup acts robot 00} depicts the effect each action prototype learned by \Method{} on the
\RobotDB~dataset.
\Method{} learns consistent semantically meaningful action prototypes that control the robot arm to move
in a specific direction.
We note that some actions prototypes, e.g, action 5 and 7, perform semantically similar actions but with different velocities.

\subsubsection{\ButtonPress~Dataset}

\begin{figure}[t]

	\resizebox{1.0 \linewidth}{!}{
	\begin{tikzpicture} 
		\node(P0)[fill=none] {};
		%
		%
		\node(seed)[anchor=north west, inner sep=0, outer sep=0] at (P0) 
		{\includegraphics[height=3cm]
			{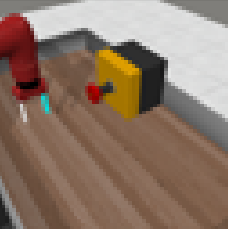}
		};
		\draw[black](seed.south west) rectangle (seed.north east);
		\node(0)[anchor=south,draw=none,ultra thick,inner sep=0, outer sep=0] at
		([xshift=-0.25cm, yshift=-1.65cm]seed.north west)
		{{\rotatebox{90}{\Large GT}}};
		\node(target_00)[anchor=west, inner sep=0, outer sep=0] at
		([xshift=0.00cm]seed.east) 
		{\includegraphics[height=3cm]
			{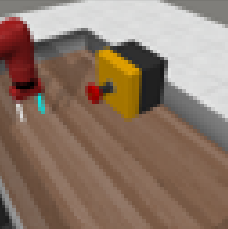}
		};
		\draw[black](target_00.south west) rectangle (target_00.north east);
		\node(target_01)[anchor=west, inner sep=0, outer sep=0] at
		([xshift=0.00cm]target_00.east) 
		{\includegraphics[height=3cm]
			{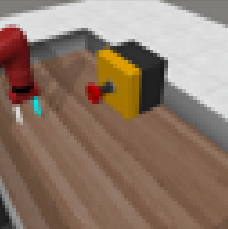}
		};
		\draw[black](target_01.south west) rectangle (target_01.north east);
		\node(target_02)[anchor=west, inner sep=0, outer sep=0] at
		([xshift=0.00cm]target_01.east) 
		{\includegraphics[height=3cm]
			{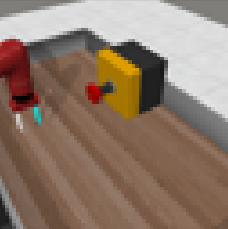}
		};
		\draw[black](target_02.south west) rectangle (target_02.north east);
		\node(target_04)[anchor=west, inner sep=0, outer sep=0] at
		([xshift=0.00cm]target_02.east)
		{\includegraphics[height=3cm]
			{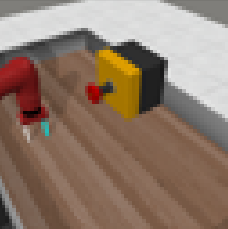}
		};
		\draw[black](target_04.south west) rectangle (target_04.north east);
		\node(target_07)[anchor=west, inner sep=0, outer sep=0] at
		([xshift=0.00cm]target_04.east)
		{\includegraphics[height=3cm]
			{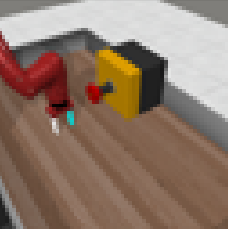}
		};
		\draw[black](target_07.south west) rectangle (target_07.north east);
		\node(target_11)[anchor=west, inner sep=0, outer sep=0] at
		([xshift=0.00cm]target_07.east)
		{\includegraphics[height=3cm]
			{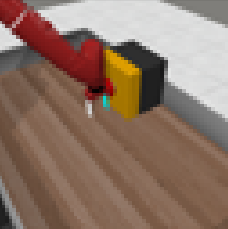}
		};
		\draw[black](target_11.south west) rectangle (target_11.north east);
		\node(target_14)[anchor=west, inner sep=0, outer sep=0] at
		([xshift=0.00cm]target_11.east)
		{\includegraphics[height=3cm]
			{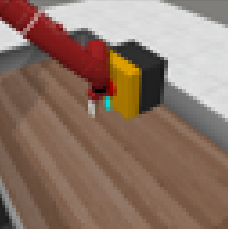}
		};
		\draw[black](target_14.south west) rectangle (target_14.north east);
		%
		%
		%
		\node(playslot_00)[anchor=north, inner sep=0, outer sep=0] at
		([xshift=0.00cm]target_00.south) 
		{\includegraphics[height=3cm]
			{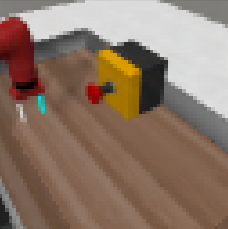}
		};
		\draw[black](playslot_00.south west) rectangle (playslot_00.north east);
		\node(playslot_01)[anchor=north, inner sep=0, outer sep=0] at
		([xshift=0.00cm]target_01.south) 
		{\includegraphics[height=3cm]
			{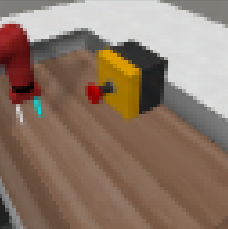}
		};
		\draw[black](playslot_01.south west) rectangle (playslot_01.north east);
		\node(playslot_02)[anchor=north, inner sep=0, outer sep=0] at
		([xshift=0.00cm]target_02.south) 
		{\includegraphics[height=3cm]
			{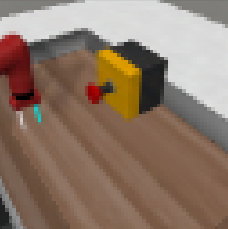}
		};
		\draw[black](playslot_02.south west) rectangle (playslot_02.north east);
		\node(playslot_04)[anchor=north, inner sep=0, outer sep=0] at
		([xshift=0.00cm]target_04.south) 
		{\includegraphics[height=3cm]
			{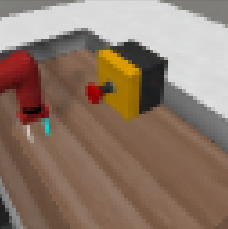}
		};
		\draw[black](playslot_04.south west) rectangle (playslot_04.north east);
		\node(playslot_07)[anchor=north, inner sep=0, outer sep=0] at
		([xshift=0.00cm]target_07.south) 
		{\includegraphics[height=3cm]
			{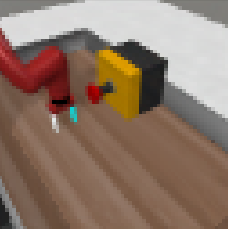}
		};
		\draw[black](playslot_07.south west) rectangle (playslot_07.north east);
		\node(playslot_11)[anchor=north, inner sep=0, outer sep=0] at
		([xshift=0.00cm]target_11.south) 
		{\includegraphics[height=3cm]
			{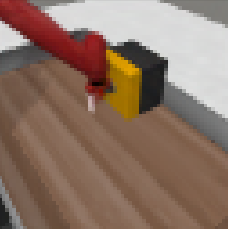}
		};
		\draw[black](playslot_11.south west) rectangle (playslot_11.north east);
		\node(playslot_14)[anchor=north, inner sep=0, outer sep=0] at
		([xshift=0.00cm]target_14.south) 
		{\includegraphics[height=3cm]
			{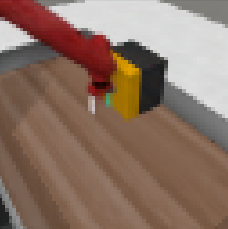}
		};
		\draw[black](playslot_14.south west) rectangle (playslot_14.north east);
		\node(PlaySlotLabel)[anchor=east,draw=none,ultra thick,inner sep=0, outer sep=0] at 	([xshift=-0.5cm, yshift=0.0cm]playslot_00.west)
		{\Large{\Method}};
		%
		%
		%
		%
		\node(svg_00)[anchor=north, inner sep=0, outer sep=0] at
		([xshift=0.00cm]playslot_00.south) 
		{\includegraphics[height=3cm]
			{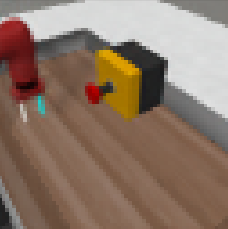}
		};
		\draw[black](svg_00.south west) rectangle (svg_00.north east);
		\node(svg_01)[anchor=north, inner sep=0, outer sep=0] at
		([xshift=0.00cm]playslot_01.south) 
		{\includegraphics[height=3cm]
			{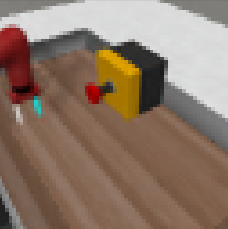}
		};
		\draw[black](svg_01.south west) rectangle (svg_01.north east);
		\node(svg_02)[anchor=north, inner sep=0, outer sep=0] at
		([xshift=0.00cm]playslot_02.south) 
		{\includegraphics[height=3cm]
			{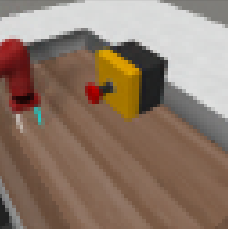}
		};
		\draw[black](svg_02.south west) rectangle (svg_02.north east);
		\node(svg_04)[anchor=north, inner sep=0, outer sep=0] at
		([xshift=0.00cm]playslot_04.south) 
		{\includegraphics[height=3cm]
			{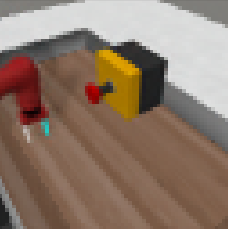}
		};
		\draw[black](svg_04.south west) rectangle (svg_04.north east);
		\node(svg_07)[anchor=north, inner sep=0, outer sep=0] at
		([xshift=0.00cm]playslot_07.south) 
		{\includegraphics[height=3cm]
			{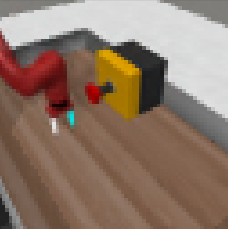}
		};
		\draw[black](svg_07.south west) rectangle (svg_07.north east);
		\node(svg_11)[anchor=north, inner sep=0, outer sep=0] at
		([xshift=0.00cm]playslot_11.south) 
		{\includegraphics[height=3cm]
			{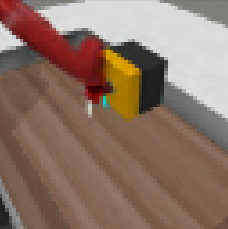}
		};
		\draw[black](svg_11.south west) rectangle (svg_11.north east);
		\node(svg_14)[anchor=north, inner sep=0, outer sep=0] at
		([xshift=0.00cm]playslot_14.south) 
		{\includegraphics[height=3cm]
			{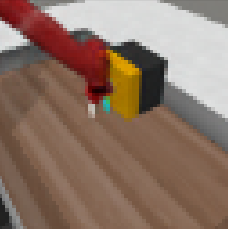}
		};
		\draw[black](svg_14.south west) rectangle (svg_14.north east);
		\node(PlaySlotLabel)[anchor=east,draw=none,ultra thick,inner sep=0, outer sep=0] at 	([xshift=-0.5cm, yshift=0.0cm]svg_00.west)
		{\Large{SVG}};
		%
		%
		%
		\node(CADDY_00)[anchor=north, inner sep=0, outer sep=0] at
		([xshift=0.00cm]svg_00.south) 
		{\includegraphics[height=3cm]
			{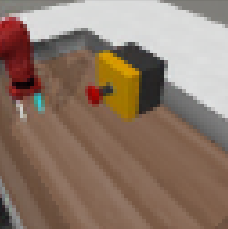}
		};
		\draw[black](CADDY_00.south west) rectangle (CADDY_00.north east);
		\node(CADDY_01)[anchor=north, inner sep=0, outer sep=0] at
		([xshift=0.00cm]svg_01.south) 
		{\includegraphics[height=3cm]
			{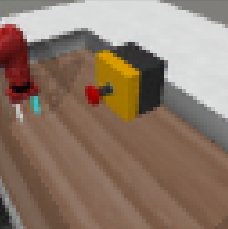}
		};
		\draw[black](CADDY_01.south west) rectangle (CADDY_01.north east);
		\node(CADDY_02)[anchor=north, inner sep=0, outer sep=0] at
		([xshift=0.00cm]svg_02.south) 
		{\includegraphics[height=3cm]
			{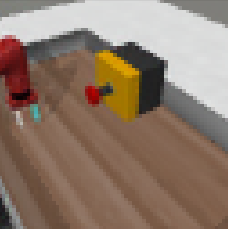}
		};
		\draw[black](CADDY_02.south west) rectangle (CADDY_02.north east);
		\node(CADDY_04)[anchor=north, inner sep=0, outer sep=0] at
		([xshift=0.00cm]svg_04.south) 
		{\includegraphics[height=3cm]
			{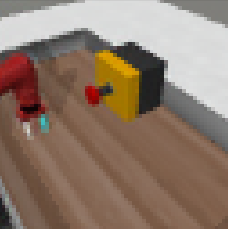}
		};
		\draw[black](CADDY_04.south west) rectangle (CADDY_04.north east);
		\node(CADDY_07)[anchor=north, inner sep=0, outer sep=0] at
		([xshift=0.00cm]svg_07.south) 
		{\includegraphics[height=3cm]
			{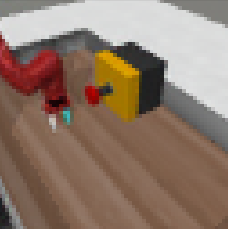}
		};
		\draw[black](CADDY_07.south west) rectangle (CADDY_07.north east);
		\node(CADDY_11)[anchor=north, inner sep=0, outer sep=0] at
		([xshift=0.00cm]svg_11.south) 
		{\includegraphics[height=3cm]
			{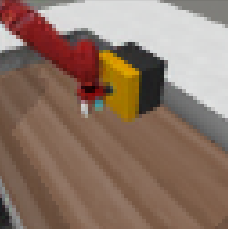}
		};
		\draw[black](CADDY_11.south west) rectangle (CADDY_11.north east);
		\node(CADDY_14)[anchor=north, inner sep=0, outer sep=0] at
		([xshift=0.00cm]svg_14.south) 
		{\includegraphics[height=3cm]
			{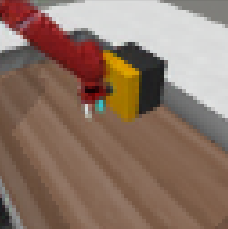}
		};
		\draw[black](CADDY_14.south west) rectangle (CADDY_14.north east);
		\node(PlaySlotLabel)[anchor=east,draw=none,ultra thick,inner sep=0, outer sep=0] at 	([xshift=-0.5cm, yshift=0.0cm]CADDY_00.west)
		{\Large{CADDY}};
		%
		%
		%
		%
		\node(0)[anchor=south,draw=none,ultra thick,inner sep=0, outer sep=0] at 	([yshift=0.1cm]seed.north)
		{\large{$t=1$}};
		\node(0)[anchor=south,draw=none,ultra thick,inner sep=0, outer sep=0] at 	([yshift=0.1cm]target_00.north)
		{\large{$2$}};
		\node(0)[anchor=south,draw=none,ultra thick,inner sep=0, outer sep=0] at 	([yshift=0.1cm]target_01.north)
		{\large{$3$}};
		\node(0)[anchor=south,draw=none,ultra thick,inner sep=0, outer sep=0] at 	([yshift=0.1cm]target_02.north)
		{\large{$4$}};
		\node(0)[anchor=south,draw=none,ultra thick,inner sep=0, outer sep=0] at 	([yshift=0.1cm]target_04.north)
		{\large{$6$}};
		\node(0)[anchor=south,draw=none,ultra thick,inner sep=0, outer sep=0] at 	([yshift=0.1cm]target_07.north)
		{\large{$9$}};
		\node(0)[anchor=south,draw=none,ultra thick,inner sep=0, outer sep=0] at 	([yshift=0.1cm]target_11.north)
		{\large{$13$}};
		\node(0)[anchor=south,draw=none,ultra thick,inner sep=0, outer sep=0] at 	([yshift=0.1cm]target_14.north)
		{\large{$16$}};
	\end{tikzpicture}
}
	
\caption{
	Qualitative comparison on \ButtonPress.
	All methods successfully reconstruct the ground truth sequence by inferring the robot's trajectory.
}
\label{fig: sup button qual}
\end{figure}

\begin{figure}[t]
	\resizebox{1.0 \linewidth}{!}{
		\begin{tikzpicture} 
			\node(P0)[fill=none] {};
			%
			%
			\node(gt_00)[anchor=north west, inner sep=0, outer sep=0] at (P0) 
			{\includegraphics[height=3cm]
				{./imgs/supplementary/qual_button/PlaySlot_ind_figs/seed_00.png}
			};
			\draw[black](gt_00.south west) rectangle (gt_00.north east);
			\node(0)[anchor=south,draw=none,ultra thick,inner sep=0, outer sep=0] at
			([xshift=-0.25cm, yshift=-1.65cm]gt_00.north west)
			{{\rotatebox{90}{\Large GT}}};
			\node(gt_01)[anchor=west, inner sep=0, outer sep=0] at
			([xshift=0.00cm]gt_00.east) 
			{\includegraphics[height=3cm]
				{./imgs/supplementary/qual_button/PlaySlot_ind_figs/target_00.png}
			};
			\draw[black](gt_01.south west) rectangle (gt_01.north east);
			\node(gt_02)[anchor=west, inner sep=0, outer sep=0] at
			([xshift=0.00cm]gt_01.east) 
			{\includegraphics[height=3cm]
				{./imgs/supplementary/qual_button/PlaySlot_ind_figs/target_01.png}
			};
			\draw[black](gt_02.south west) rectangle (gt_02.north east);
			\node(gt_03)[anchor=west, inner sep=0, outer sep=0] at
			([xshift=0.00cm]gt_02.east) 
			{\includegraphics[height=3cm]
				{./imgs/supplementary/qual_button/PlaySlot_ind_figs/target_02.png}
			};
			\draw[black](gt_03.south west) rectangle (gt_03.north east);
			\node(gt_05)[anchor=west, inner sep=0, outer sep=0] at
			([xshift=0.00cm]gt_03.east)
			{\includegraphics[height=3cm]
				{./imgs/supplementary/qual_button/PlaySlot_ind_figs/target_04.png}
			};
			\draw[black](gt_05.south west) rectangle (gt_05.north east);
			\node(gt_08)[anchor=west, inner sep=0, outer sep=0] at
			([xshift=0.00cm]gt_05.east)
			{\includegraphics[height=3cm]
				{./imgs/supplementary/qual_button/PlaySlot_ind_figs/target_07.png}
			};
			\draw[black](gt_08.south west) rectangle (gt_08.north east);
			\node(gt_10)[anchor=west, inner sep=0, outer sep=0] at
			([xshift=0.00cm]gt_08.east)
			{\includegraphics[height=3cm]
				{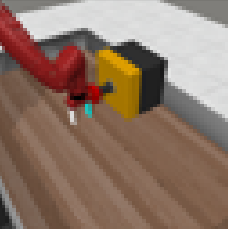}
			};
			\draw[black](gt_10.south west) rectangle (gt_10.north east);
			\node(gt_12)[anchor=west, inner sep=0, outer sep=0] at
			([xshift=0.00cm]gt_10.east)
			{\includegraphics[height=3cm]
				{./imgs/supplementary/qual_button/PlaySlot_ind_figs/target_11.png}
			};
			\draw[black](gt_12.south west) rectangle (gt_12.north east);
			\node(gt_15)[anchor=west, inner sep=0, outer sep=0] at
			([xshift=0.00cm]gt_12.east)
			{\includegraphics[height=3cm]
				{./imgs/supplementary/qual_button/PlaySlot_ind_figs/target_14.png}
			};
			\draw[black](gt_15.south west) rectangle (gt_15.north east);
			%
			%
			%
			%
			%
			\node(playslot_01)[anchor=north, inner sep=0, outer sep=0] at
			([xshift=0.00cm]gt_01.south) 
			{\includegraphics[height=3cm]
				{./imgs/supplementary/qual_button/PlaySlot_ind_figs/pred_00.png}
			};
			\draw[black](playslot_01.south west) rectangle (playslot_01.north east);
			\node(playslot_02)[anchor=north, inner sep=0, outer sep=0] at
			([xshift=0.00cm]gt_02.south) 
			{\includegraphics[height=3cm]
				{./imgs/supplementary/qual_button/PlaySlot_ind_figs/pred_01.png}
			};
			\draw[black](playslot_02.south west) rectangle (playslot_02.north east);
			\node(playslot_03)[anchor=north, inner sep=0, outer sep=0] at
			([xshift=0.00cm]gt_03.south) 
			{\includegraphics[height=3cm]
				{./imgs/supplementary/qual_button/PlaySlot_ind_figs/pred_02.png}
			};
			\draw[black](playslot_03.south west) rectangle (playslot_03.north east);
			\node(playslot_05)[anchor=north, inner sep=0, outer sep=0] at
			([xshift=0.00cm]gt_05.south) 
			{\includegraphics[height=3cm]
				{./imgs/supplementary/qual_button/PlaySlot_ind_figs/pred_04.png}
			};
			\draw[black](playslot_05.south west) rectangle (playslot_05.north east);
			\node(playslot_08)[anchor=north, inner sep=0, outer sep=0] at
			([xshift=0.00cm]gt_08.south) 
			{\includegraphics[height=3cm]
				{./imgs/supplementary/qual_button/PlaySlot_ind_figs/pred_07.png}
			};
			\draw[black](playslot_08.south west) rectangle (playslot_08.north east);
			\node(playslot_10)[anchor=north, inner sep=0, outer sep=0] at
			([xshift=0.00cm]gt_10.south) 
			{\includegraphics[height=3cm]
				{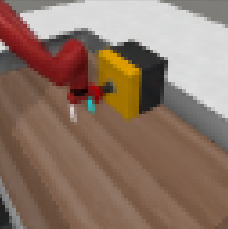}
			};
			\draw[black](playslot_10.south west) rectangle (playslot_10.north east);
			\node(playslot_12)[anchor=north, inner sep=0, outer sep=0] at
			([xshift=0.00cm]gt_12.south) 
			{\includegraphics[height=3cm]
				{./imgs/supplementary/qual_button/PlaySlot_ind_figs/pred_11.png}
			};
			\draw[black](playslot_12.south west) rectangle (playslot_12.north east);
			\node(playslot_15)[anchor=north, inner sep=0, outer sep=0] at
			([xshift=0.00cm]gt_15.south) 
			{\includegraphics[height=3cm]
				{./imgs/supplementary/qual_button/PlaySlot_ind_figs/pred_14.png}
			};
			\draw[black](playslot_15.south west) rectangle (playslot_15.north east);
			\node(PlaySlotLabel)[anchor=east,draw=none,ultra thick,inner sep=0, outer sep=0] at 	([xshift=-0.5cm, yshift=0.0cm]playslot_00.west)
			{\Large{Predictions}};
			%
			%
			%
			%
			\foreach \i [count=\j] in {01,02,03,05,08,10,12,15} {
				\node(segm_\i)[anchor=north, inner sep=0, outer sep=0] at 
				([xshift=0.00cm]playslot_\i.south) 
				{\includegraphics[height=3cm]
					{./imgs/supplementary/qual_button/PlaySlot_ind_figs/segm_\i.png}
				};
				\draw[black](segm_\i.south west) rectangle (segm_\i.north east);
			}
			\node(PlaySlotLabel)[anchor=east,draw=none,ultra thick,inner sep=0, outer sep=0] at 
			([xshift=-0.5cm, yshift=0.0cm]segm_01.west)
			{\makecell{\Large{Slot}\\ \Large{Masks}}};
			%
			%
			%
			\foreach \i [count=\j] in {01,02,03,05,08,10,12,15} {
				\node(obj_0_\j)[anchor=north, inner sep=0, outer sep=0] at 
				([xshift=0.00cm]segm_\i.south) 
				{\includegraphics[height=3cm]
					{./imgs/supplementary/qual_button/PlaySlot_ind_figs/MaskedObj_00_\i.png}
				};
				\draw[black](obj_0_\j.south west) rectangle (obj_0_\j.north east);
			}
			\node(PlaySlotLabel)[anchor=east,draw=none,ultra thick,inner sep=0, outer sep=0] at
			([xshift=-0.5cm, yshift=0.0cm]obj_0_1.west)
			{\makecell{\Large{Object 1}}};
			%
			%
			%
			\foreach \i [count=\j] in {01,02,03,05,08,10,12,15} {
				\node(obj_1_\j)[anchor=north, inner sep=0, outer sep=0] at 
				([xshift=0.00cm]obj_0_\j.south) 
				{\includegraphics[height=3cm]
					{./imgs/supplementary/qual_button/PlaySlot_ind_figs/MaskedObj_01_\i.png}
				};
				\draw[black](obj_1_\j.south west) rectangle (obj_1_\j.north east);
			}
			\node(PlaySlotLabel)[anchor=east,draw=none,ultra thick,inner sep=0, outer sep=0] at
			([xshift=-0.5cm, yshift=0.0cm]obj_1_1.west)
			{\makecell{\Large{Object 2}}};
			%
			%
			\foreach \i [count=\j] in {01,02,03,05,08,10,12,15} {
				\node(obj_2_\j)[anchor=north, inner sep=0, outer sep=0] at 
				([xshift=0.00cm]obj_1_\j.south) 
				{\includegraphics[height=3cm]
					{./imgs/supplementary/qual_button/PlaySlot_ind_figs/MaskedObj_02_\i.png}
				};
				\draw[black](obj_2_\j.south west) rectangle (obj_2_\j.north east);
			}
			\node(PlaySlotLabel)[anchor=east,draw=none,ultra thick,inner sep=0, outer sep=0] at
			([xshift=-0.5cm, yshift=0.0cm]obj_2_1.west)
			{\makecell{\Large{Object 3}}};
			%
			%
			%
			\foreach \i [count=\j] in {01,02,03,05,08,10,12,15} {
				\node(obj_4_\j)[anchor=north, inner sep=0, outer sep=0] at 
				([xshift=0.00cm]obj_2_\j.south) 
				{\includegraphics[height=3cm]
					{./imgs/supplementary/qual_button/PlaySlot_ind_figs/MaskedObj_03_\i.png}
				};
				\draw[black](obj_4_\j.south west) rectangle (obj_4_\j.north east);
			}
			\node(PlaySlotLabel)[anchor=east,draw=none,ultra thick,inner sep=0, outer sep=0] at
			([xshift=-0.5cm, yshift=0.0cm]obj_4_1.west)
			{\makecell{\Large{Object 4}}};
			%
			%
			%
			%
			\node(0)[anchor=south,draw=none,ultra thick,inner sep=0, outer sep=0] at 	([yshift=0.1cm]gt_00.north)
			{\large{$t=1$}};
			\node(0)[anchor=south,draw=none,ultra thick,inner sep=0, outer sep=0] at 	([yshift=0.1cm]gt_01.north)
			{\large{$2$}};
			\node(0)[anchor=south,draw=none,ultra thick,inner sep=0, outer sep=0] at 	([yshift=0.1cm]gt_02.north)
			{\large{$3$}};
			\node(0)[anchor=south,draw=none,ultra thick,inner sep=0, outer sep=0] at 	([yshift=0.1cm]gt_03.north)
			{\large{$4$}};
			\node(0)[anchor=south,draw=none,ultra thick,inner sep=0, outer sep=0] at 	([yshift=0.1cm]gt_05.north)
			{\large{$6$}};
			\node(0)[anchor=south,draw=none,ultra thick,inner sep=0, outer sep=0] at 	([yshift=0.1cm]gt_08.north)
			{\large{$9$}};
			\node(0)[anchor=south,draw=none,ultra thick,inner sep=0, outer sep=0] at 	([yshift=0.1cm]gt_10.north)
			{\large{$11$}};
			\node(0)[anchor=south,draw=none,ultra thick,inner sep=0, outer sep=0] at 	([yshift=0.1cm]gt_12.north)
			{\large{$13$}};
			\node(0)[anchor=south,draw=none,ultra thick,inner sep=0, outer sep=0] at 	([yshift=0.1cm]gt_15.north)
			{\large{$16$}};
		\end{tikzpicture}
	}
\vspace{-0.7cm}
\caption{
	\Method~predictions and object representations on a \ButtonPress~sequence.
	We visualize the ground truth sequence, the predicted frames, segmentation obtained by assigning a different color to each slot mask, as well as the object reconstructions for four slots.
	\Method assigns one slot to the background, one slot for box and red button, and two slots for different parts of the robot arm.
	}
\label{fig: sup button objs 00}
\end{figure}

\Figure{fig: sup button qual} shows a comparisons between \Method, CADDY and SVG on the \ButtonPress~dataset.
All methods successfully predict the ground truth sequence by inferring and modeling the robot's dynamics,

\Figure{fig: sup button objs 00} shows \Method's~predictions and object representations on a \ButtonPress~sequence.
We visualize the ground truth sequence, the predicted frames, segmentation obtained by assigning a different color to each slot mask, as well as the object reconstructions for four slots.
\Method~assigns one slot to the background, one slot for box and red button, and two slots for different parts of the robot arm.

\begin{figure}[t]
	\resizebox{1.0 \linewidth}{!}{
		\begin{tikzpicture} 
			\node(P0)[fill=none] {};
			%
			%
			\node(seed)[anchor=north west, inner sep=0, outer sep=0] at (P0) 
			{\includegraphics[height=3cm]
				{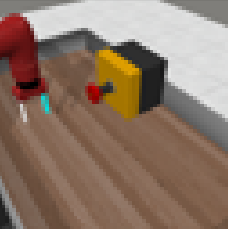}
			};
			\draw[black](gt_00.south west) rectangle (gt_00.north east);
			\node(0)[anchor=south,draw=none,ultra thick,inner sep=0, outer sep=0] at
			([xshift=-0.25cm, yshift=-1.65cm]seed.north west)
			{{\rotatebox{90}{\Large GT}}};
			\node(gt_1)[anchor=west, inner sep=0, outer sep=0] at
			([xshift=0.00cm]seed.east) 
			{\includegraphics[height=3cm]
				{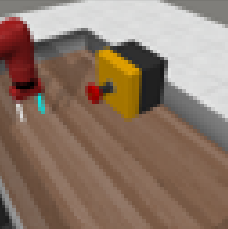}
			};
			\draw[black](gt_1.south west) rectangle (gt_1.north east);
			\node(gt_2)[anchor=west, inner sep=0, outer sep=0] at
			([xshift=0.00cm]gt_1.east) 
			{\includegraphics[height=3cm]
				{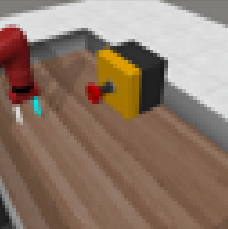}
			};
			\draw[black](gt_2.south west) rectangle (gt_2.north east);
			\node(gt_3)[anchor=west, inner sep=0, outer sep=0] at
			([xshift=0.00cm]gt_2.east) 
			{\includegraphics[height=3cm]
				{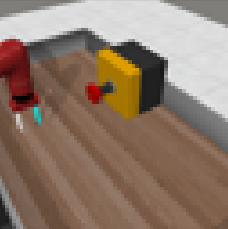}
			};
			\draw[black](gt_3.south west) rectangle (gt_3.north east);
			\node(gt_4)[anchor=west, inner sep=0, outer sep=0] at
			([xshift=0.00cm]gt_3.east)
			{\includegraphics[height=3cm]
				{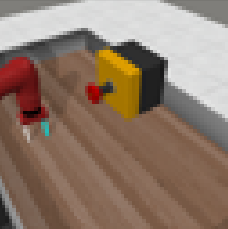}
			};
			\draw[black](gt_4.south west) rectangle (gt_4.north east);
			\node(gt_5)[anchor=west, inner sep=0, outer sep=0] at
			([xshift=0.00cm]gt_4.east)
			{\includegraphics[height=3cm]
				{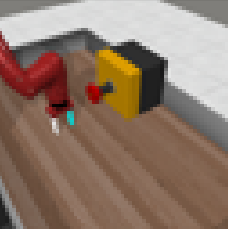}
			};
			\draw[black](gt_5.south west) rectangle (gt_5.north east);
			\node(gt_6)[anchor=west, inner sep=0, outer sep=0] at
			([xshift=0.00cm]gt_5.east)
			{\includegraphics[height=3cm]
				{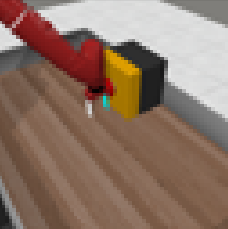}
			};
			\draw[black](gt_6.south west) rectangle (gt_6.north east);
			\node(gt_7)[anchor=west, inner sep=0, outer sep=0] at
			([xshift=0.00cm]gt_6.east)
			{\includegraphics[height=3cm]
				{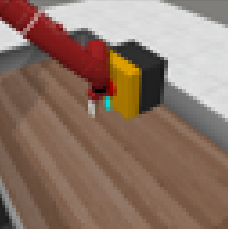}
			};
			\draw[black](gt_7.south west) rectangle (gt_7.north east);
			%
			%
			%
			\foreach \i [count=\j] in {00,01,02,04,07,11,14} {
				\node(post_\i)[anchor=north, inner sep=0, outer sep=0] at 
				([xshift=0.00cm]gt_\j.south) 
				{\includegraphics[height=3cm]
					{./imgs/supplementary/qual_button/PlaySlot_act_figs/post_\i.png}
				};
				\draw[black](post_\i.south west) rectangle (post_\i.north east);
			}
			\node(PlaySlotLabel)[anchor=east,draw=none,ultra thick,inner sep=0, outer sep=0] at 
			([xshift=-0.5cm, yshift=0.0cm]post_00.west)
			{\makecell{ {\Large Inferred} \\ {\Large Dynamics} }};
			%
			%
			%
			\foreach \i [count=\j] in {00,01,02,04,07,11,14} {
				\node(prior_1_\i)[anchor=north, inner sep=0, outer sep=0] at 
				([xshift=0.00cm]post_\i.south) 
				{\includegraphics[height=3cm]
					{./imgs/supplementary/qual_button/PlaySlot_act_figs/prior_1_\i.png}
				};
				\draw[black](prior_1_\i.south west) rectangle (prior_1_\i.north east);
			}
			\node(PlaySlotLabel)[anchor=east,draw=none,ultra thick,inner sep=0, outer sep=0] at 
			([xshift=-0.6cm, yshift=0.0cm]prior_1_00.west)
			{\makecell{ {\Large Action 1} }};
			%
			%
			%
			\foreach \i [count=\j] in {00,01,02,04,07,11,14} {
				\node(prior_2_\i)[anchor=north, inner sep=0, outer sep=0] at 
				([xshift=0.00cm]prior_1_\i.south) 
				{\includegraphics[height=3cm]
					{./imgs/supplementary/qual_button/PlaySlot_act_figs/prior_2_\i.png}
				};
				\draw[black](prior_2_\i.south west) rectangle (prior_2_\i.north east);
			}
			\node(PlaySlotLabel)[anchor=east,draw=none,ultra thick,inner sep=0, outer sep=0] at 
			([xshift=-0.6cm, yshift=0.0cm]prior_2_00.west)
			{\makecell{ {\Large Action 2} }};
			%
			%
			%
			\foreach \i [count=\j] in {00,01,02,04,07,11,14} {
				\node(prior_3_\i)[anchor=north, inner sep=0, outer sep=0] at 
				([xshift=0.00cm]prior_2_\i.south) 
				{\includegraphics[height=3cm]
					{./imgs/supplementary/qual_button/PlaySlot_act_figs/prior_3_\i.png}
				};
				\draw[black](prior_3_\i.south west) rectangle (prior_3_\i.north east);
			}
			\node(PlaySlotLabel)[anchor=east,draw=none,ultra thick,inner sep=0, outer sep=0] at 
			([xshift=-0.6cm, yshift=0.0cm]prior_3_00.west)
			{\makecell{ {\Large Action 3} }};
			%
			%
			%
			%
			\foreach \i [count=\j] in {00,01,02,04,07,11,14} {
				\node(prior_4_\i)[anchor=north, inner sep=0, outer sep=0] at 
				([xshift=0.00cm]prior_3_\i.south) 
				{\includegraphics[height=3cm]
					{./imgs/supplementary/qual_button/PlaySlot_act_figs/prior_4_\i.png}
				};
				\draw[black](prior_4_\i.south west) rectangle (prior_4_\i.north east);
			}
			\node(PlaySlotLabel)[anchor=east,draw=none,ultra thick,inner sep=0, outer sep=0] at 
			([xshift=-0.6cm, yshift=0.0cm]prior_4_00.west)
			{\makecell{ {\Large Action 4} }};
			%
			%
			%
			%
			\foreach \i [count=\j] in {00,01,02,04,07,11,14} {
				\node(prior_5_\i)[anchor=north, inner sep=0, outer sep=0] at 
				([xshift=0.00cm]prior_4_\i.south) 
				{\includegraphics[height=3cm]
					{./imgs/supplementary/qual_button/PlaySlot_act_figs/prior_5_\i.png}
				};
				\draw[black](prior_5_\i.south west) rectangle (prior_5_\i.north east);
			}
			\node(PlaySlotLabel)[anchor=east,draw=none,ultra thick,inner sep=0, outer sep=0] at 
			([xshift=-0.6cm, yshift=0.0cm]prior_5_00.west)
			{\makecell{ {\Large Action 5} }};
			%
			%
			%
			%
			\foreach \i [count=\j] in {00,01,02,04,07,11,14} {
				\node(prior_6_\i)[anchor=north, inner sep=0, outer sep=0] at 
				([xshift=0.00cm]prior_5_\i.south) 
				{\includegraphics[height=3cm]
					{./imgs/supplementary/qual_button/PlaySlot_act_figs/prior_6_\i.png}
				};
				\draw[black](prior_6_\i.south west) rectangle (prior_6_\i.north east);
			}
			\node(PlaySlotLabel)[anchor=east,draw=none,ultra thick,inner sep=0, outer sep=0] at 
			([xshift=-0.6cm, yshift=0.0cm]prior_6_00.west)
			{\makecell{ {\Large Action 6} }};
			%
			%
			
			%
			%
			\foreach \i [count=\j] in {00,01,02,04,07,11,14} {
				\node(prior_7_\i)[anchor=north, inner sep=0, outer sep=0] at 
				([xshift=0.00cm]prior_6_\i.south) 
				{\includegraphics[height=3cm]
					{./imgs/supplementary/qual_button/PlaySlot_act_figs/prior_7_\i.png}
				};
				\draw[black](prior_7_\i.south west) rectangle (prior_7_\i.north east);
			}
			\node(PlaySlotLabel)[anchor=east,draw=none,ultra thick,inner sep=0, outer sep=0] at 
			([xshift=-0.6cm, yshift=0.0cm]prior_7_00.west)
			{\makecell{ {\Large Action 7} }};
			%
			%
			%
			%
			\foreach \i [count=\j] in {00,01,02,04,07,11,14} {
				\node(prior_8_\i)[anchor=north, inner sep=0, outer sep=0] at 
				([xshift=0.00cm]prior_7_\i.south) 
				{\includegraphics[height=3cm]
					{./imgs/supplementary/qual_button/PlaySlot_act_figs/prior_8_\i.png}
				};
				\draw[black](prior_8_\i.south west) rectangle (prior_8_\i.north east);
			}
			\node(PlaySlotLabel)[anchor=east,draw=none,ultra thick,inner sep=0, outer sep=0] at 
			([xshift=-0.6cm, yshift=0.0cm]prior_8_00.west)
			{\makecell{ {\Large Action 8} }};
			%
			%
			%
			%
			%
			%
			%
			%
			\node(0)[anchor=south,draw=none,ultra thick,inner sep=0, outer sep=0] at 	([yshift=0.1cm]gt_00.north)
			{\large{$t=1$}};
			\node(0)[anchor=south,draw=none,ultra thick,inner sep=0, outer sep=0] at 	([yshift=0.1cm]gt_01.north)
			{\large{$2$}};
			\node(0)[anchor=south,draw=none,ultra thick,inner sep=0, outer sep=0] at 	([yshift=0.1cm]gt_02.north)
			{\large{$3$}};
			\node(0)[anchor=south,draw=none,ultra thick,inner sep=0, outer sep=0] at 	([yshift=0.1cm]gt_03.north)
			{\large{$4$}};
			\node(0)[anchor=south,draw=none,ultra thick,inner sep=0, outer sep=0] at 	([yshift=0.1cm]gt_4.north)
			{\large{$6$}};
			\node(0)[anchor=south,draw=none,ultra thick,inner sep=0, outer sep=0] at 	([yshift=0.1cm]gt_5.north)
			{\large{$9$}};
			\node(0)[anchor=south,draw=none,ultra thick,inner sep=0, outer sep=0] at 	([yshift=0.1cm]gt_6.north)
			{\large{$13$}};
			\node(0)[anchor=south,draw=none,ultra thick,inner sep=0, outer sep=0] at 	([yshift=0.1cm]gt_7.north)
			{\large{$16$}};
		\end{tikzpicture}
	}
	\vspace{-0.4cm}
	\caption{
		\Method~predictions conditioned on different latent actions, including the inferred inverse dynamics, as well as each action prototypes learned on \ButtonPress.
		We generate a sequence by repeatedly conditioning the prediction process on a a single action prototype.
		The model learns action prototypes that control the robot to move consistently on a specific direction.
	}
	\label{fig: sup acts button 00}
\end{figure}

\Figure{fig: sup acts button 00}  depicts the effect each action prototype learned by \Method on the \ButtonPress~dataset.

\subsubsection{GridShapes Dataset}

\begin{figure}[t]
	\resizebox{1.0 \linewidth}{!}{
		\begin{tikzpicture} 
			\node(P0)[fill=none] {};
			%
			%
			\node(seed)[anchor=north west, inner sep=0, outer sep=0] at (P0) 
			{\includegraphics[height=3cm]
				{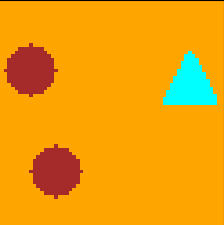}
			};
			\draw[black](gt_00.south west) rectangle (gt_00.north east);
			\node(0)[anchor=south,draw=none,ultra thick,inner sep=0, outer sep=0] at
			([xshift=-0.25cm, yshift=-1.65cm]seed.north west)
			{{\rotatebox{90}{\Large GT}}};
			\node(gt_1)[anchor=west, inner sep=0, outer sep=0] at
			([xshift=0.00cm]seed.east) 
			{\includegraphics[height=3cm]
				{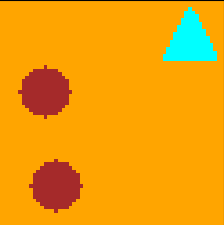}
			};
			\draw[black](gt_1.south west) rectangle (gt_1.north east);
			\node(gt_2)[anchor=west, inner sep=0, outer sep=0] at
			([xshift=0.00cm]gt_1.east) 
			{\includegraphics[height=3cm]
				{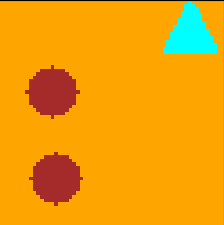}
			};
			\draw[black](gt_2.south west) rectangle (gt_2.north east);
			\node(gt_3)[anchor=west, inner sep=0, outer sep=0] at
			([xshift=0.00cm]gt_2.east) 
			{\includegraphics[height=3cm]
				{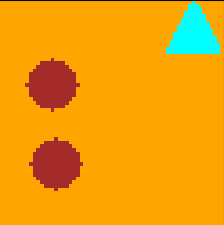}
			};
			\draw[black](gt_3.south west) rectangle (gt_3.north east);
			\node(gt_4)[anchor=west, inner sep=0, outer sep=0] at
			([xshift=0.00cm]gt_3.east)
			{\includegraphics[height=3cm]
				{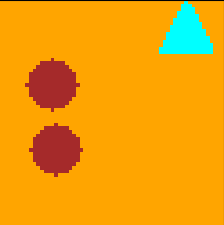}
			};
			\draw[black](gt_4.south west) rectangle (gt_4.north east);
			\node(gt_5)[anchor=west, inner sep=0, outer sep=0] at
			([xshift=0.00cm]gt_4.east)
			{\includegraphics[height=3cm]
				{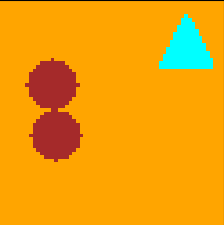}
			};
			\draw[black](gt_5.south west) rectangle (gt_5.north east);
			\node(gt_6)[anchor=west, inner sep=0, outer sep=0] at
			([xshift=0.00cm]gt_5.east)
			{\includegraphics[height=3cm]
				{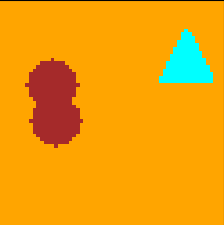}
			};
			\draw[black](gt_6.south west) rectangle (gt_6.north east);
			\node(gt_7)[anchor=west, inner sep=0, outer sep=0] at
			([xshift=0.00cm]gt_6.east)
			{\includegraphics[height=3cm]
				{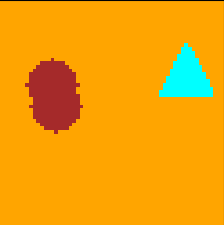}
			};
			\draw[black](gt_7.south west) rectangle (gt_7.north east);
			%
			%
			%
			\foreach \i [count=\j] in {00,01,03,05,07,09,11} {
				\node(post_\i)[anchor=north, inner sep=0, outer sep=0] at 
				([xshift=0.00cm]gt_\j.south) 
				{\includegraphics[height=3cm]
					{./imgs/supplementary/gridshapes_00/PlaySlot_ind_figs/post_\i.png}
				};
				\draw[black](post_\i.south west) rectangle (post_\i.north east);
			}
			\node(PlaySlotLabel)[anchor=east,draw=none,ultra thick,inner sep=0, outer sep=0] at 
			([xshift=-0.5cm, yshift=0.0cm]post_00.west)
			{\makecell{ {\Large Inferred} \\ {\Large Dynamics} }};
			%
			%
			%
			\foreach \i [count=\j] in {00,01,03,05,07,09,11} {
				\node(prior_1_\i)[anchor=north, inner sep=0, outer sep=0] at 
				([xshift=0.00cm]post_\i.south) 
				{\includegraphics[height=3cm]
					{./imgs/supplementary/gridshapes_00/PlaySlot_ind_figs/prior_1_\i.png}
				};
				\draw[black](prior_1_\i.south west) rectangle (prior_1_\i.north east);
			}
			\node(PlaySlotLabel)[anchor=east,draw=none,ultra thick,inner sep=0, outer sep=0] at 
			([xshift=-0.6cm, yshift=0.0cm]prior_1_00.west)
			{\makecell{ {\Large `Up'} \\ {Action 1} }};
			%
			%
			%
			\foreach \i [count=\j] in {00,01,03,05,07,09,11} {
				\node(prior_2_\i)[anchor=north, inner sep=0, outer sep=0] at 
				([xshift=0.00cm]prior_1_\i.south) 
				{\includegraphics[height=3cm]
					{./imgs/supplementary/gridshapes_00/PlaySlot_ind_figs/prior_2_\i.png}
				};
				\draw[black](prior_2_\i.south west) rectangle (prior_2_\i.north east);
			}
			\node(PlaySlotLabel)[anchor=east,draw=none,ultra thick,inner sep=0, outer sep=0] at 
			([xshift=-0.6cm, yshift=0.0cm]prior_2_00.west)
			{\makecell{ {\Large `Right'} \\ {Action 2} }};
			%
			%
			%
			\foreach \i [count=\j] in {00,01,03,05,07,09,11} {
				\node(prior_3_\i)[anchor=north, inner sep=0, outer sep=0] at 
				([xshift=0.00cm]prior_2_\i.south) 
				{\includegraphics[height=3cm]
					{./imgs/supplementary/gridshapes_00/PlaySlot_ind_figs/prior_3_\i.png}
				};
				\draw[black](prior_3_\i.south west) rectangle (prior_3_\i.north east);
			}
			\node(PlaySlotLabel)[anchor=east,draw=none,ultra thick,inner sep=0, outer sep=0] at 
			([xshift=-0.6cm, yshift=0.0cm]prior_3_00.west)
			{\makecell{ {\Large `Down'} \\ {Action 3} }};
			%
			%
			%
			%
			\foreach \i [count=\j] in {00,01,03,05,07,09,11} {
				\node(prior_4_\i)[anchor=north, inner sep=0, outer sep=0] at 
				([xshift=0.00cm]prior_3_\i.south) 
				{\includegraphics[height=3cm]
					{./imgs/supplementary/gridshapes_00/PlaySlot_ind_figs/prior_4_\i.png}
				};
				\draw[black](prior_4_\i.south west) rectangle (prior_4_\i.north east);
			}
			\node(PlaySlotLabel)[anchor=east,draw=none,ultra thick,inner sep=0, outer sep=0] at 
			([xshift=-0.6cm, yshift=0.0cm]prior_4_00.west)
			{\makecell{ {\Large `Stay'} \\ {Action 4} }};
			%
			%
			%
			%
			\foreach \i [count=\j] in {00,01,03,05,07,09,11} {
				\node(prior_5_\i)[anchor=north, inner sep=0, outer sep=0] at 
				([xshift=0.00cm]prior_4_\i.south) 
				{\includegraphics[height=3cm]
					{./imgs/supplementary/gridshapes_00/PlaySlot_ind_figs/prior_5_\i.png}
				};
				\draw[black](prior_5_\i.south west) rectangle (prior_5_\i.north east);
			}
			\node(PlaySlotLabel)[anchor=east,draw=none,ultra thick,inner sep=0, outer sep=0] at 
			([xshift=-0.6cm, yshift=0.0cm]prior_5_00.west)
			{\makecell{ {\Large `Left'} \\ {Action 5} }};
			%
			%
			%
			%
			\node(0)[anchor=south,draw=none,ultra thick,inner sep=0, outer sep=0] at 	([yshift=0.1cm]gt_00.north)
			{\large{$t=1$}};
			\node(0)[anchor=south,draw=none,ultra thick,inner sep=0, outer sep=0] at 	([yshift=0.1cm]gt_01.north)
			{\large{$2$}};
			\node(0)[anchor=south,draw=none,ultra thick,inner sep=0, outer sep=0] at 	([yshift=0.1cm]gt_02.north)
			{\large{$3$}};
			\node(0)[anchor=south,draw=none,ultra thick,inner sep=0, outer sep=0] at 	([yshift=0.1cm]gt_03.north)
			{\large{$5$}};
			\node(0)[anchor=south,draw=none,ultra thick,inner sep=0, outer sep=0] at 	([yshift=0.1cm]gt_4.north)
			{\large{$7$}};
			\node(0)[anchor=south,draw=none,ultra thick,inner sep=0, outer sep=0] at 	([yshift=0.1cm]gt_5.north)
			{\large{$9$}};
			\node(0)[anchor=south,draw=none,ultra thick,inner sep=0, outer sep=0] at 	([yshift=0.1cm]gt_6.north)
			{\large{$11$}};
			\node(0)[anchor=south,draw=none,ultra thick,inner sep=0, outer sep=0] at 	([yshift=0.1cm]gt_7.north)
			{\large{$13$}};
		\end{tikzpicture}
	}
	\vspace{-0.4cm}
	\caption{
		\Method~predictions conditioned on different latent actions, including the inferred inverse dynamics, as well as each action prototypes learned on the GridShapes dataset.
		We generate a sequence by repeatedly conditioning the prediction process on a a single action prototype.
		The model learns the five possible actions and achieves sharp predictions by forecasting the motion of each object individually.
	}
	\label{fig: sup acts gridshapes 00}
\end{figure}

\Figure{fig: sup acts gridshapes 00} depicts the effect each action prototype learned by \Method{} on a variant of GridShapes with three shapes.
\Method{} successfully captures the five possible object movements—up, down, left, right, and stay—predicting future frames by modeling the motion of each object independently.
However, we observe that \Method{} sometimes generates artifacts when shapes reach the image boundaries.
We attribute this to the training data, where objects change direction upon reaching the boundary mimicking a bouncing effect.
This leads to poor prediction performance when conditioning the model to predict outside its training distribution.

\subsubsection{Sketchy Dataset}

\begin{figure}[t]
	\resizebox{1.0 \linewidth}{!}{
		\begin{tikzpicture} 
			\node(P0)[fill=none] {};
			%
			%
			\node(gt_00)[anchor=north west, inner sep=0, outer sep=0] at (P0) 
			{\includegraphics[height=3cm]
				{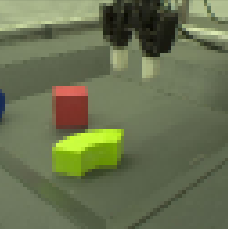}
			};
			\draw[black](gt_00.south west) rectangle (gt_00.north east);
			\node(0)[anchor=south,draw=none,ultra thick,inner sep=0, outer sep=0] at
			([xshift=-0.25cm, yshift=-1.65cm]gt_00.north west)
			{{\rotatebox{90}{\Large GT}}};
			\node(gt_01)[anchor=west, inner sep=0, outer sep=0] at
			([xshift=0.00cm]gt_00.east) 
			{\includegraphics[height=3cm]
				{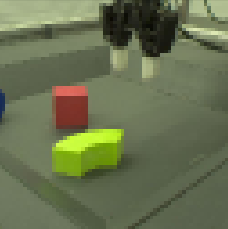}
			};
			\draw[black](gt_01.south west) rectangle (gt_01.north east);
			\node(gt_02)[anchor=west, inner sep=0, outer sep=0] at
			([xshift=0.00cm]gt_01.east) 
			{\includegraphics[height=3cm]
				{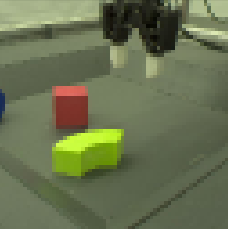}
			};
			\draw[black](gt_02.south west) rectangle (gt_02.north east);
			\node(gt_03)[anchor=west, inner sep=0, outer sep=0] at
			([xshift=0.00cm]gt_02.east) 
			{\includegraphics[height=3cm]
				{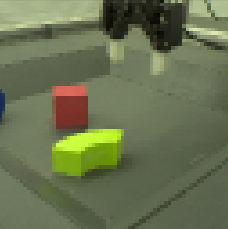}
			};
			\draw[black](gt_03.south west) rectangle (gt_03.north east);
			\node(gt_05)[anchor=west, inner sep=0, outer sep=0] at
			([xshift=0.00cm]gt_03.east)
			{\includegraphics[height=3cm]
				{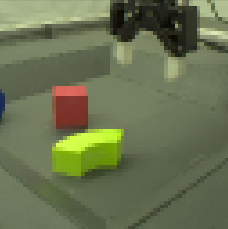}
			};
			\draw[black](gt_05.south west) rectangle (gt_05.north east);
			\node(gt_08)[anchor=west, inner sep=0, outer sep=0] at
			([xshift=0.00cm]gt_05.east)
			{\includegraphics[height=3cm]
				{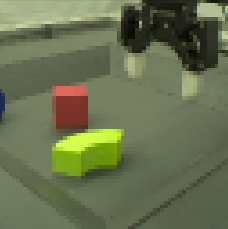}
			};
			\draw[black](gt_08.south west) rectangle (gt_08.north east);
			\node(gt_10)[anchor=west, inner sep=0, outer sep=0] at
			([xshift=0.00cm]gt_08.east)
			{\includegraphics[height=3cm]
				{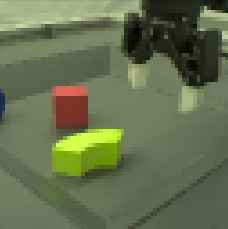}
			};
			\draw[black](gt_10.south west) rectangle (gt_10.north east);
			\node(gt_12)[anchor=west, inner sep=0, outer sep=0] at
			([xshift=0.00cm]gt_10.east)
			{\includegraphics[height=3cm]
				{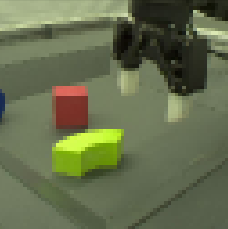}
			};
			\draw[black](gt_12.south west) rectangle (gt_12.north east);
			\node(gt_15)[anchor=west, inner sep=0, outer sep=0] at
			([xshift=0.00cm]gt_12.east)
			{\includegraphics[height=3cm]
				{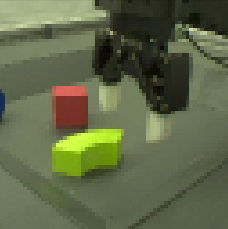}
			};
			\draw[black](gt_15.south west) rectangle (gt_15.north east);
			%
			%
			%
			%
			%
			\node(playslot_01)[anchor=north, inner sep=0, outer sep=0] at
			([xshift=0.00cm]gt_01.south) 
			{\includegraphics[height=3cm]
				{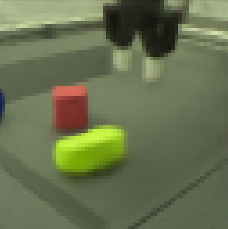}
			};
			\draw[black](playslot_01.south west) rectangle (playslot_01.north east);
			\node(playslot_02)[anchor=north, inner sep=0, outer sep=0] at
			([xshift=0.00cm]gt_02.south) 
			{\includegraphics[height=3cm]
				{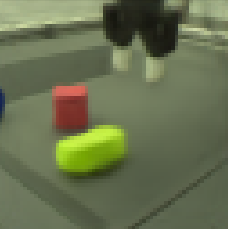}
			};
			\draw[black](playslot_02.south west) rectangle (playslot_02.north east);
			\node(playslot_03)[anchor=north, inner sep=0, outer sep=0] at
			([xshift=0.00cm]gt_03.south) 
			{\includegraphics[height=3cm]
				{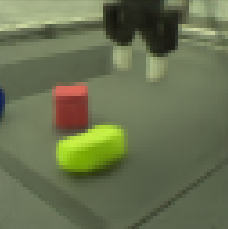}
			};
			\draw[black](playslot_03.south west) rectangle (playslot_03.north east);
			\node(playslot_05)[anchor=north, inner sep=0, outer sep=0] at
			([xshift=0.00cm]gt_05.south) 
			{\includegraphics[height=3cm]
				{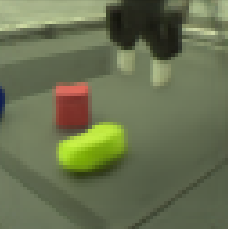}
			};
			\draw[black](playslot_05.south west) rectangle (playslot_05.north east);
			\node(playslot_08)[anchor=north, inner sep=0, outer sep=0] at
			([xshift=0.00cm]gt_08.south) 
			{\includegraphics[height=3cm]
				{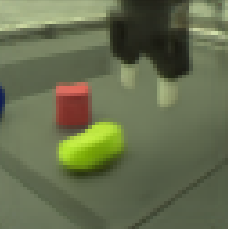}
			};
			\draw[black](playslot_08.south west) rectangle (playslot_08.north east);
			\node(playslot_10)[anchor=north, inner sep=0, outer sep=0] at
			([xshift=0.00cm]gt_10.south) 
			{\includegraphics[height=3cm]
				{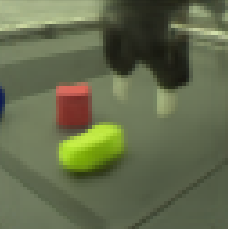}
			};
			\draw[black](playslot_10.south west) rectangle (playslot_10.north east);
			\node(playslot_12)[anchor=north, inner sep=0, outer sep=0] at
			([xshift=0.00cm]gt_12.south) 
			{\includegraphics[height=3cm]
				{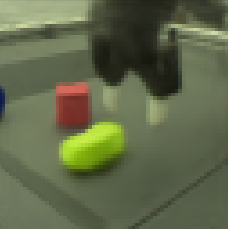}
			};
			\draw[black](playslot_12.south west) rectangle (playslot_12.north east);
			\node(playslot_15)[anchor=north, inner sep=0, outer sep=0] at
			([xshift=0.00cm]gt_15.south) 
			{\includegraphics[height=3cm]
				{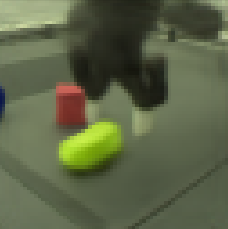}
			};
			\draw[black](playslot_15.south west) rectangle (playslot_15.north east);
			\node(PlaySlotLabel)[anchor=east,draw=none,ultra thick,inner sep=0, outer sep=0] at 	([xshift=-0.5cm, yshift=0.0cm]playslot_00.west)
			{\Large{Predictions}};
			%
			%
			%
			%
			\foreach \i [count=\j] in {01,02,03,05,08,10,12,15} {
				\node(segm_\i)[anchor=north, inner sep=0, outer sep=0] at 
				([xshift=0.00cm]playslot_\i.south) 
				{\includegraphics[height=3cm]
					{./imgs/supplementary/sketchy_00/PlaySlot_ind_figs/segm_\i.png}
				};
				\draw[black](segm_\i.south west) rectangle (segm_\i.north east);
			}
			\node(PlaySlotLabel)[anchor=east,draw=none,ultra thick,inner sep=0, outer sep=0] at 
			([xshift=-0.5cm, yshift=0.0cm]segm_01.west)
			{\makecell{\Large{Slot}\\ \Large{Masks}}};
			%
			%
			%
			\foreach \i [count=\j] in {01,02,03,05,08,10,12,15} {
				\node(obj_0_\j)[anchor=north, inner sep=0, outer sep=0] at 
				([xshift=0.00cm]segm_\i.south) 
				{\includegraphics[height=3cm]
					{./imgs/supplementary/sketchy_00/PlaySlot_ind_figs/MaskedObj_00_\i.png}
				};
				\draw[black](obj_0_\j.south west) rectangle (obj_0_\j.north east);
			}
			\node(PlaySlotLabel)[anchor=east,draw=none,ultra thick,inner sep=0, outer sep=0] at
			([xshift=-0.5cm, yshift=0.0cm]obj_0_1.west)
			{\makecell{\Large{Object 1}}};
			%
			%
			%
			\foreach \i [count=\j] in {01,02,03,05,08,10,12,15} {
				\node(obj_1_\j)[anchor=north, inner sep=0, outer sep=0] at 
				([xshift=0.00cm]obj_0_\j.south) 
				{\includegraphics[height=3cm]
					{./imgs/supplementary/sketchy_00/PlaySlot_ind_figs/MaskedObj_01_\i.png}
				};
				\draw[black](obj_1_\j.south west) rectangle (obj_1_\j.north east);
			}
			\node(PlaySlotLabel)[anchor=east,draw=none,ultra thick,inner sep=0, outer sep=0] at
			([xshift=-0.5cm, yshift=0.0cm]obj_1_1.west)
			{\makecell{\Large{Object 2}}};
			%
			%
			\foreach \i [count=\j] in {01,02,03,05,08,10,12,15} {
				\node(obj_2_\j)[anchor=north, inner sep=0, outer sep=0] at 
				([xshift=0.00cm]obj_1_\j.south) 
				{\includegraphics[height=3cm]
					{./imgs/supplementary/sketchy_00/PlaySlot_ind_figs/MaskedObj_04_\i.png}
				};
				\draw[black](obj_2_\j.south west) rectangle (obj_2_\j.north east);
			}
			\node(PlaySlotLabel)[anchor=east,draw=none,ultra thick,inner sep=0, outer sep=0] at
			([xshift=-0.5cm, yshift=0.0cm]obj_2_1.west)
			{\makecell{\Large{Object 3}}};
			%
			%
			%
			\foreach \i [count=\j] in {01,02,03,05,08,10,12,15} {
				\node(obj_4_\j)[anchor=north, inner sep=0, outer sep=0] at 
				([xshift=0.00cm]obj_2_\j.south) 
				{\includegraphics[height=3cm]
					{./imgs/supplementary/sketchy_00/PlaySlot_ind_figs/MaskedObj_05_\i.png}
				};
				\draw[black](obj_4_\j.south west) rectangle (obj_4_\j.north east);
			}
			\node(PlaySlotLabel)[anchor=east,draw=none,ultra thick,inner sep=0, outer sep=0] at
			([xshift=-0.5cm, yshift=0.0cm]obj_4_1.west)
			{\makecell{\Large{Object 4}}};
			%
			%
			%
			\foreach \i [count=\j] in {01,02,03,05,08,10,12,15} {
				\node(obj_5_\j)[anchor=north, inner sep=0, outer sep=0] at 
				([xshift=0.00cm]obj_4_\j.south) 
				{\includegraphics[height=3cm]
					{./imgs/supplementary/sketchy_00/PlaySlot_ind_figs/MaskedObj_06_\i.png}
				};
				\draw[black](obj_5_\j.south west) rectangle (obj_5_\j.north east);
			}
			\node(PlaySlotLabel)[anchor=east,draw=none,ultra thick,inner sep=0, outer sep=0] at
			([xshift=-0.5cm, yshift=0.0cm]obj_5_1.west)
			{\makecell{\Large{Object 5}}};
			%
			%
			%
			%
			%
			\foreach \i [count=\j] in {01,02,03,05,08,10,12,15} {
				\node(obj_6_\j)[anchor=north, inner sep=0, outer sep=0] at 
				([xshift=0.00cm]obj_5_\j.south) 
				{\includegraphics[height=3cm]
					{./imgs/supplementary/sketchy_00/PlaySlot_ind_figs/MaskedObj_07_\i.png}
				};
				\draw[black](obj_6_\j.south west) rectangle (obj_6_\j.north east);
			}
			\node(PlaySlotLabel)[anchor=east,draw=none,ultra thick,inner sep=0, outer sep=0] at
			([xshift=-0.5cm, yshift=0.0cm]obj_6_1.west)
			{\makecell{\Large{Object 6}}};
			%
			%
			%
			%
			%
			%
			\node(0)[anchor=south,draw=none,ultra thick,inner sep=0, outer sep=0] at 	([yshift=0.1cm]gt_00.north)
			{\large{$t=1$}};
			\node(0)[anchor=south,draw=none,ultra thick,inner sep=0, outer sep=0] at 	([yshift=0.1cm]gt_01.north)
			{\large{$2$}};
			\node(0)[anchor=south,draw=none,ultra thick,inner sep=0, outer sep=0] at 	([yshift=0.1cm]gt_02.north)
			{\large{$3$}};
			\node(0)[anchor=south,draw=none,ultra thick,inner sep=0, outer sep=0] at 	([yshift=0.1cm]gt_03.north)
			{\large{$4$}};
			\node(0)[anchor=south,draw=none,ultra thick,inner sep=0, outer sep=0] at 	([yshift=0.1cm]gt_05.north)
			{\large{$6$}};
			\node(0)[anchor=south,draw=none,ultra thick,inner sep=0, outer sep=0] at 	([yshift=0.1cm]gt_08.north)
			{\large{$9$}};
			\node(0)[anchor=south,draw=none,ultra thick,inner sep=0, outer sep=0] at 	([yshift=0.1cm]gt_10.north)
			{\large{$11$}};
			\node(0)[anchor=south,draw=none,ultra thick,inner sep=0, outer sep=0] at 	([yshift=0.1cm]gt_12.north)
			{\large{$13$}};
			\node(0)[anchor=south,draw=none,ultra thick,inner sep=0, outer sep=0] at 	([yshift=0.1cm]gt_15.north)
			{\large{$16$}};
		\end{tikzpicture}
	}
\vspace{-0.7cm}
\caption{
	\Method~predictions and object representations on a Sketchy sequence.
	We visualize the ground truth sequence, the predicted frames, segmentation obtained by assigning a different color to each slot mask, as well as the object reconstructions for four slots.
	\Method~assigns two slots to the workspace and background, two slots for each part of the robot gripper, and two slots for different objects present in the scene.
	}
\label{fig: sup sketchy objs 00}
\end{figure}

\Figure{fig: sup sketchy objs 00} shows \Method's~predictions and object representations on a Sketchy sequence.
We visualize the ground truth sequence, the predicted frames, segmentation obtained by assigning a different color to each slot mask, as well as the object reconstructions for four slots.
\Method~assigns two slots to the workspace and background, two slots for each part of the robot gripper, and two slots for different objects present in the scene.

\end{document}